\documentclass[11pt,a4paper]{article}
\usepackage{a4wide}
\usepackage{additional_def}

\newcommand{\gradf}{{\rm grad} f}

\newcommand{\changeBM}[1]{#1}
\newcommand{\changeHK}[1]{#1}
\newcommand{\changeHS}[1]{#1}

\newcommand{\changeHSS}[1]{#1}

\title{Riemannian stochastic variance reduced gradient\\ on Grassmann manifold}

\author{Hiroyuki Kasai\thanks{Graduate School of Informatics and Engineering, The University of Electro-Communications, Tokyo, Japan ({\tt kasai@is.uec.ac.jp}).} \and Hiroyuki Sato\thanks{Department of Information and Computer Technology, Tokyo University of Science, Tokyo, Japan ({\tt hsato@rs.tus.ac.jp}).} \and Bamdev Mishra\thanks{Core ML, Amazon.com, Bangalore, India ({\tt bamdevm@amazon.com}.)}}

\begin{document}

\maketitle

\begin{abstract}
Stochastic variance reduction algorithms have recently become popular for minimizing the average of a large, but finite, number of loss functions. \changeBM{In this paper, we propose a novel Riemannian extension of the Euclidean stochastic variance reduced gradient algorithm (R-SVRG) to a compact manifold search space.} To this end, we show the developments on the Grassmann manifold. The key challenges of averaging, addition, and subtraction of multiple gradients are addressed with notions like logarithm mapping and parallel translation of vectors on the Grassmann manifold. We \changeBM{present a} global convergence analysis \changeBM{of the proposed algorithm} with a decay step-size and \changeBM{a} local convergence rate analysis under a fixed step-size with under some natural assumptions. The proposed algorithm is applied on a number of problems on the Grassmann manifold like principal components analysis, low-rank matrix completion, and the Karcher mean computation. In all these cases, the proposed algorithm outperforms the \changeBM{standard} Riemannian stochastic gradient descent algorithm. 
\end{abstract}

\section{Introduction}
\label{Sec:intro}

A general loss minimization problem is defined as $\min_{w} f(w)$, where $f(w):= \frac{1}{N} \sum_{n=1}^N f_n(w)$, $w$ is the model variable, $N$ is the number of samples\changeBM{,} and $f_n(w)$ is the loss incurred on $n$-th sample. The {\it full gradient decent} (GD) algorithm requires evaluations of $N$ derivatives, i.e., $\sum_{n=1}^N \nabla f_n(w)$, per iteration, \changeBM{which is computationally heavy} when $N$ is very large. A popular alternative is to \changeBM{use only one derivative} $\nabla f_n(w)$ per iteration for $n$-th sample, \changeBM{which is the basis of} the {\it stochastic gradient descent} (SGD) algorithm. When a relatively large step-size is used in SGD, the train loss decreases fast \changeBM{in} the beginning, \changeBM{but results in} big fluctuations around the solution. On the other hand, when a small step-size is used, \changeBM{SGD requires a large number of iterations to converge}. \changeBM{To circumvent this issue, SGD starts with a relatively large step-size and decreases it gradually with iterations.}

\changeBM{Recently, {\it variance reduction} techniques have been proposed to} accelerate the convergence of SGD \cite{Johnson_NIPS_2013_s, Mairal_SIAMJOPT_2015,Roux_NIPS_2012_s,Shalev_arXiv_2012_s,Shalev_JMLR_2013_s,Defazio_NIPS_2014_s,Zhang_SIAMJO_2014_s}. Stochastic variance reduced gradient (SVRG) is a popular algorithm that enjoys superior convergence properties \cite{Johnson_NIPS_2013_s}. \changeBM{For smooth and strongly convex functions,} SVRG \changeBM{has convergence rates similar} to those of stochastic dual coordinate ascent  \cite{Shalev_JMLR_2013_s} and stochastic average gradient (SAG) algorithms \cite{Roux_NIPS_2012_s}. Garber and Hazan \cite{Garber_arXiv_2015_s} analyze the convergence rate for SVRG when $f$ is a convex function that is a sum of non-convex (but smooth) terms and apply this result \changeBM{to} the principal component analysis (PCA) problem. Shalev-Shwartz  \cite{Shalev_arXiv_2015_s} also proposes similar results. Allen-Zhu and Yuan \cite{Allen-Zhu-arXiv_2015_s} further study the same case with better convergence rates. Shamir \cite{Shamir_arXiv_2015_s} studies specifically the convergence properties of the variance reduction PCA algorithm. Very recently, Allen-Zhu and Hazan \cite{Allen-Zhu-arXiv_2016_s} propose a variance reduction method for faster non-convex optimization. \changeBM{However, it should be noted that all these cases assume that search space is Euclidean.}

\changeBM{In this paper, we deal with problems where \changeBM{the} variables have a manifold structure}. They include, for example, the low-rank matrix completion problem \cite{Mishra_ICDC_2014_s}, the Karcher mean computation problem, and the PCA problem. In all these problems, optimization on {\it Riemannian manifolds} has shown state-of-the-art performance. The Riemannian framework exploits the geometry of the constrained matrix search space to design efficient optimization algorithms \cite{Absil_OptAlgMatManifold_2008}. Specifically, the problem $\min_{w \in \mathcal{M}} f(w)$, where $\mathcal{M}$ is a Riemannian manifold, is solved as an \emph{unconstrained optimization problem} defined over the Riemannian manifold search space. Bonnabel \cite{Bonnabel_IEEETAC_2013_s} proposes a {\it Riemannian stochastic gradient} algorithm (R-SGD) that extends SGD from the Euclidean space to Riemannian manifolds.

\changeBM{Building upon the work of Bonnabel \cite{Bonnabel_IEEETAC_2013_s}, we propose a novel (and to the best of our knowledge, the first) extension of the stochastic variance reduction gradient algorithm in the Euclidean space to the Riemannian manifold search space (R-SVRG).} This extension is not trivial and requires particular consideration in dealing with averaging, addition and subtraction of multiple gradients at different points on the manifold $\mathcal{M}$. To this end, this paper specifically focuses on the {\it Grassmann manifold} ${\rm Gr}(r,d)$, \changeBM{which} is the set of $r$-dimensional linear subspaces \changeBM{in} $\mathbb{R}^d$. Nonetheless, the proposed algorithm and the analysis presented in this paper can be generalized to other compact Riemannian manifolds. 

The paper is organized as follows. Section \ref{Sec:GrassmannAndProblems} discusses the Grassmann manifold and three popular optimization problems, where the Grassmann manifold plays an essential role. The detailed description of R-SVRG are given in Section \ref{Sec:R-SVRG}. Section \ref{Sec:Analysis} presents the global convergence analysis and the local convergence rate analysis of R-SVRG. In Section \ref{Sec:NumericalComparison}, numerical comparisons with R-SGD on the three problems suggest superior performance of R-SVRG. 
The concrete proofs of the main theorems and the related lemmas, and additional numerical experiments are shown in Sections \changeHK{{\bf A}},  \changeHK{{\bf B}}, and  \changeHK{{\bf C}}, respectively, of the supplementary file. 
Our proposed R-SVRG is implemented in the Matlab toolbox Manopt \cite{Boumal_Manopt_2014_s}. The Matlab codes for the proposed algorithms are available at \url{https://bamdevmishra.com/codes/RSVRG/}.

\section{Grassmann manifold and problems on Grassmann manifold}
\label{Sec:GrassmannAndProblems}
This section briefly introduces the Grassmann manifold and motivates three problems on the Grassmann manifold. 

{\bf Grassmann manifold.} \changeBM{An element} on the Grassmann manifold is \changeHS{represented by a} $d \times r$ orthogonal matrix \mat{U} with \changeBM{orthonormal columns}, i.e., $\mat{U}^T\mat{U}=\mat{I}$. Two \changeBM{orthogonal} matrices \changeBM{represent the same element on} the Grassmann manifold if they are related by right multiplication of a $r\times r$ orthogonal matrix $\mat{O} \in \mathcal{O}(r)$. Equivalently, an element of the Grassmann manifold is identified with a set of $d \times r$ orthogonal matrices $[\mat{U}]: =\{\mat{U}\mat{O}_r :\mat{O} \in \mathcal{O}(r)\}$. \changeBM{In other words,} ${\rm Gr}(r,d) :={\rm St}(r,d)/ \mathcal{O}(r)$, where ${\rm St}(r,d)$ is the {\it Stiefel manifold} that is the set of matrices of size $d \times r$ with orthonormal columns. The Grassmann manifold has the structure of a Riemannian quotient manifold \cite[Section~3.4]{Absil_OptAlgMatManifold_2008}.

{\it Geodesics} on manifolds generalize the concept of straight lines in the Euclidean space. For every vector in the tangent space  $\xi \in T_w \mathcal{M}$ at $w \in \mathcal{M}$, there exists an interval $I$ about 0 and a unique geodesic $\gamma_e(t,w,\xi):I \rightarrow \mathcal{M}$ such that $\gamma_e(0)=w$ and $\dot{\gamma_e}(0) = \xi$. The mapping ${\rm Exp}_{w} :T_{w}\mathcal{M} \rightarrow \mathcal{M}: \xi \mapsto {\rm Exp}_{w} \xi=\gamma_e(1,w,\xi)$ is called the {\it exponential mapping} at $w$. If $\mathcal{M}$ is a complete manifold, exponential mapping is defined for all vectors $\xi \in T_{w}\mathcal{M}$. The exponential mapping for the Grassmann manifold from $\mat{U}(0) := \mat{U} \in {\rm Gr}(r,d)$ in the direction of $\xi \in T_{\scriptsize \mat{U}(0)}$ is given in closed form as \cite[Section 5.4]{Absil_OptAlgMatManifold_2008}
\begin{eqnarray}
\label{Eq:exponential_map}
\mat{U} (t) & = & [\mat{U}(0)  \mat{V}\ \  \mat{W}] 
	\left[
    		\begin{array}{c}
      		\cos t \Sigma  \\
      		\sin t \Sigma \\
    		\end{array}
	\right]
    \mat{V}^T,
\end{eqnarray}	
where $\xi=\mat{W} \Sigma \mat{V}^T$ is the rank-$r$ singular value decomposition of $\xi$. The $\cos(\cdot)$ and $\sin(\cdot)$ operations are only on the diagonal entries. 

{\it Parallel translation} transports a vector field along the geodesic curve $\gamma$ that satisfies $P_{\gamma}^{a \leftarrow a} = \gamma (a)$ and $\frac{D}{dt}(P_{\gamma}^{t \leftarrow a} \xi(a))=0$ \cite[Section 5.4]{Absil_OptAlgMatManifold_2008}, where $P_{\gamma}^{b \leftarrow a}$ is the parallel translation operator sending $\xi(a)$ to $\xi(b)$. The parallel translation of $\zeta \in T_{\scriptsize \mat{U}(0)}$ on the Grassmann manifold along $\gamma(t)$ with $\xi$ is given in closed form by
\begin{eqnarray}
\label{Eq:parallel_translation}
\zeta(t) & = & \left( [\mat{U}(0) \mat{V}\ \  \mat{W}] 
	\left[
    		\begin{array}{c}
      		-\sin t \Sigma \\
      		\cos t \Sigma \\
    		\end{array}
	\right]
    \mat{W}^T + (\mat{I} - \mat{W}\mat{W}^T)
    \right) \zeta.
\end{eqnarray}	

\changeBM{Given two points $w$ and $z$ on $\mathcal{M}$, the \emph{logarithm mapping} or simply  {\it log mapping} maps $z$ to a vector $\xi \in T_{w}\mathcal{M}$ on the tangent space at $w$. Specifically, it is defined by ${\rm Log}_{w} : \mathcal{M} \rightarrow T_{w} \mathcal{M} :{\rm Exp}_{w} \xi \mapsto {\rm Log}_{w}({\rm Exp}_{w} \xi) = \xi$. It should be noted that it satisfies ${\rm dist}(w,z)=\| {\rm Log}_{w}(z)\|_{w}$, where ${\rm dist}: \mathcal{M} \times \mathcal{M} \rightarrow \mathbb{R}$ is the \emph{shortest} distance between $w$ and $z$. The logarithm map of $\mat{U}(t)$ at $\mat{U}(0)$ on the Grassmann manifold is given by}
\begin{eqnarray}
\label{Eq:logarithm_map}
\xi  &=& \log_{\scriptsize \mat{U}(0)}(\mat{U}(t)) \ = \ \mat{W} \arctan(\Sigma) \mat{V}^T,
\end{eqnarray}	
where $\mat{W}\Sigma \mat{V}^T$ \changeBM{is the rank-$r$ singular value decomposition of} $(\mat{U}(t) - \mat{U}(0) \mat{U}(0)^T  \mat{U}(t))\allowbreak(\mat{U}(0)^T \mat{U}(t))^{-1}$.

{\bf Problems on Grassmann manifold.} \changeBM{In this paper, we focus on three popular problems on the Grassmann manifold, which are the PCA, low-rank matrix completion, and the Karcher mean computation problems. In all these problems, full gradient methods, e.g., the steepest descent algorithm, become prohibitively computationally expensive when $N$ is very large, and the stochastic gradient approach is one promising way to achieve scalability.}

Given \changeBM{an} orthonormal matrix projector $\mat{U} \in {\rm St}(r,d)$, the PCA problem is to minimize the sum of squared residual errors between projected data points and the original data as
\begin{eqnarray}
\label{Eq:PCA}
{\displaystyle \min_{{\scriptsize \mat{U} \in {\rm St}(r,d)}}} &{\displaystyle \frac{1}{N}  \sum_{n=1}^N \| \vec{x}_n -  \mat{U}\mat{U}^T \vec{x}_n \|_2^2},
\end{eqnarray}
where $\vec{x}_n$ is a data vector of size $d\times 1$. The problem (\ref{Eq:PCA}) is equivalent to maximizing $\frac{1}{N} \sum_{n=1}^N \vec{x}_n^T\mat{U}\mat{U}^T\vec{x}_n$.
Here, the critical points in the space ${\rm St}(r,d)$ are not isolated because the cost function remains unchanged under the group action $\mat{U} \mapsto \mat{UO}$ for all orthogonal matrices $\mat{O}$ of size $r \times r$. Subsequently, the problem (\ref{Eq:PCA}) is an optimization problem on the Grassmann manifold ${\rm Gr}(r,d)$.

The Karcher mean is introduced as a notion of \emph{mean} on Riemannian manifolds by Karcher \cite{Karcher_1977_s}. It generalizes the notion of an  ``average''  on the manifold. Given $N$ points on the Grassmann manifold with matrix representations $\mat{Q}_1,\ldots,\mat{Q}_N$, the Karcher mean is defined as the solution to the problem
\begin{eqnarray}
\label{Eq:KarcherMean}
{\displaystyle \min_{{\scriptsize \mat{U} \in {\rm St}(r, d)}}} &{\displaystyle \frac{1}{2N} \sum_{n=1}^N ({\rm dist}(\mat{U}, \mat{Q}_n))^2},
\end{eqnarray}
\changeBM{where ${\rm dist}$ is the geodesic distance between the elements on the Grassmann manifold. The gradient of this loss function is $ \frac{1}{N} \sum_{n=1}^{N}  -{\rm Log}_{\scriptsize \mat{U}} (\mat{Q}_n)$, where ${\rm Log}$ is the log map defined in (\ref{Eq:logarithm_map}).} The Karcher mean on the Grassmann manifold ${\rm Gr}(r,d)$ is frequently used for computer vision problems such as visual object categorization and pose categorization \cite{Jayasumana_IEEETPAMI_2015_s}. Since recursive calculations of the Karcher mean are needed with each new arriving visual image, the stochastic gradient algorithm becomes \changeBM{an appealing choice} for large datasets.

\changeBM{The matrix completion problem \changeHK{is to complete} an incomplete matrix $\mat{X}$, say of size $d \times N$, from a small number of entries{. For this purpose, it assumes} a low-rank model for the matrix. If $\Omega$ is the set of the indices for which we know the entries in $\mat{X}$, the rank-$r$ matrix completion problem amounts to solving the problem
\begin{equation}\label{Eq:MC_batch}
\begin{array}{ll}
{\displaystyle \min_{{\scriptsize \mat{U}} \in \mathbb{R}^{d \times r}, {\scriptsize \mat{A}} \in \mathbb{R}^{r \times N}}} & \|\mathcal{P}_{\Omega}(\mat{UA}) - \mathcal{P}_{\Omega}(\mat X) \|_F^2,
\end{array}
\end{equation}
where the operator $\mathcal{P}_{\Omega}(\mat{X}_{ij})=\mat{X}_{ij}$ if $(i,j) \in \Omega$ and $\mathcal{P}_{\Omega}(\mat{X}_{ij})=0$ otherwise is called the orthogonal sampling operator.
Partitioning $\mat{X} = [\vec{x}_1, \vec{x}_2, \ldots, \vec{x}_n] $, the problem (\ref{Eq:MC_batch}) is equivalent to the problem
\begin{eqnarray}
\label{Eq:MC}
{\displaystyle \min_{{\scriptsize \mat{U}} \in \mathbb{R}^{d \times r}, \vec{a}_n \in \mathbb{R}^{r}}} 
& 
{\displaystyle \frac{1}{N} \sum_{n=1}^N \|  \mathcal{P}_{\Omega_n}(\mat{U} \vec{a}_n) - \mathcal{P}_{\Omega_n}(\vec{x}_n) \|_2^2,
}
\end{eqnarray}
where $\vec{x}_n \in \mathbb{R}^d$ and the operator $\mathcal{P}_{\Omega_n}$ the sampling operator for the $n$-th column. Given \mat{U}, $\vec{a}_n$ in (\ref{Eq:MC}) admits a closed-form solution. Consequently, the problem (\ref{Eq:MC}) only depends on the column space of $\mat{U}$ and is on the Grassmann manifold \cite{Balzano_arXiv_2010_s}.}

\section{Riemannian stochastic variance reduced gradient on Grassmann \changeHS{manifold}}
\label{Sec:R-SVRG}

After a brief explanation of the variance reduced gradient variants in the Euclidean space, the  Riemannian stochastic variance reduced gradient on the Grassmann manifold is proposed.  

{\bf Variance reduced gradient variants in the Euclidean space.} The SGD update in the Euclidean space is $w_{t+1}  =  w_{t}-\eta v_t$, where $v_t$ is a randomly selected vector that is called as the {\it stochastic gradient} \changeBM{and $\eta $ is the step-size}. SGD assumes an {\it unbiased estimator} of the full gradient as $\mathbb{E}_n[\nabla f_n(w_t)] = \nabla f (w_t)$. Many \changeBM{recent} variants of the variance reduced gradient of SGD attempt to reduce its variance $\mathbb{E}[\| v_t - \nabla f(w_t)\|^2]$ \changeBM{as} $t$ increases to achieve \changeBM{better} convergence \cite{Johnson_NIPS_2013_s, Mairal_SIAMJOPT_2015,Roux_NIPS_2012_s,Shalev_arXiv_2012_s,Shalev_JMLR_2013_s,Defazio_NIPS_2014_s,Zhang_SIAMJO_2014_s}. \changeBM{SVRG, proposed in \cite{Johnson_NIPS_2013_s},} introduces an explicit variance reduction strategy with double loops where $s$-th outer loop, called $s$-th {\it epoch}, has $m_s$ inner iterations. SVRG first keeps $\tilde{w}=w_{m_s}^{s-1}$ or $\tilde{w}=w_{t}^{s-1}$ for randomly chosen $t\in\{1,\ldots,m_{s-1}\}$ at the end of $(s\!\!-\!\!1)$-th epoch, and also sets the initial value of $s$-th epoch as $w_0^s=\tilde{w}$. It then computes a full gradient $\nabla f(\tilde{w})$. Subsequently, denoting the selected random index $i \in \{1, \ldots, N\}$ by $i_t^s$, SVRG randomly picks $i_t^s$-th sample for each $t \geq 1$ at $s \geq 1$ and computes the {\it modified stochastic gradient} $v_t^s$ as
\begin{eqnarray}
\label{Eq:E-SVRG}
v_t^s & = & \nabla f_{i_t^s} (w_{t-1}^{s}) -  \nabla f_{i_t^s} (\tilde{w}^{s-1})  + \nabla f(\tilde{w}^{s-1}). 
\end{eqnarray}
It should be noted that SVRG can be regarded as one special case of S2GD (Semi-stochastic gradient descent), which differs in the number of inner loop iterations chosen \cite{Konecny_arXiv_2013}.

{\bf Proposed Riemannian extension of SVRG on Grassmann \changeHS{manifold} (R-SVRG).}
We propose a Riemannian extension of SVRG, i.e., R-SVRG. Here, we denote the Riemannian stochastic gradient for $i_t^s$-th sample as $\gradf_{i_t^s}(\tilde{\mat{U}}^{s-1})$ and the {\it modified Riemannian stochastic gradient} as $\xi_t^s$ instead of $v_t^s$ \changeBM{to show differences with the Euclidean case}.

\changeBM{The way R-SVRG reduces} the variance is analogous to the SVRG algorithm in the Euclidean case. More specifically, R-SVRG keeps a $\tilde{\mat{U}}^{s-1} \in \mathcal{M}= {\rm Gr}(r,d)$ after $m_{s-1}$ stochastic update steps of $(s\!\!-\!\!1)$-th epoch, and computes the full Riemannian gradient $\gradf (\tilde{\mat{U}}^{s-1})=\frac{1}{N} \sum_{i=1}^{N} \gradf_i(\tilde{\mat{U}}^{s-1})$ only for this stored $\tilde{\mat{U}}^{s-1}$. The algorithm also computes the Riemannian stochastic gradient $\gradf_{i_t^s}(\tilde{\mat{U}}^{s-1})$ that corresponds to this $i_t^s$-th sample. Then, picking $i_t^s$-th sample for each $t$-th inner iteration of $s$-th epoch at $\mat{U}_{t-1}^{s}$, we calculate $\xi_t^s$ \changeBM{in} the same \changeBM{way} as $v_t^s$ in (\ref{Eq:E-SVRG}), i.e., by modifying the stochastic gradient $\gradf_{i_t^s}(\mat{U}_{t-1}^{s})$ using both $\gradf (\tilde{\mat{U}}^{s-1})$ and $\gradf_{i_t^s}(\tilde{\mat{U}}^{s-1})$. \changeBM{Translating the right-hand side of (\ref{Eq:E-SVRG}) to the manifold $\mathcal{M}$} involves the sum of \changeBM{$\gradf_{i_t^s}(\mat{U}_{t-1}^{s})$, $\gradf_{i_t^s}(\tilde{\mat{U}}^{s-1})$, and $\gradf(\tilde{\mat{U}}^{s-1})$, which belong to two separate tangent spaces $T_{\scriptsize \mat{U}_{t-1}^{s}}\mathcal{M}$ and  $T_{\scriptsize \tilde{\mat{U}}^{s-1}}\mathcal{M}$.} \changeBM{This operation requires particular attention on a manifold and parallel translation provides an adequate and flexible solution to handle multiple elements on two separated tangent spaces.} More concretely, $\gradf_{i_t^s}(\tilde{\mat{U}}^{s-1})$ and $\gradf(\tilde{\mat{U}}^{s-1})$ are firstly parallel-transported to $T_{\scriptsize \mat{U}_{t-1}^{s}}\mathcal{M}$ at the current point $\mat{U}_{t-1}^{s}$, then they are ready to be added to $\gradf_{i_t^s}(\mat{U}_{t-1}^{s})$ on $T_{\scriptsize \mat{U}_{t-1}^{s}}\mathcal{M}$. Consequently, the modified Riemannian stochastic gradient $\xi_t^s$ at $t$-th inner iteration of $s$-th epoch is set as
\begin{eqnarray}
\label{Eq:R-SVRG-Grad-paralleltrans}
\xi_t^s & = & \gradf_{i_t^s}(\mat{U}_{t-1}^{s}) -  P_{\gamma}^{\scriptsize \mat{U}_{t-1}^{s} \leftarrow \tilde{\mat{U}}^{s-1}} \left(\gradf_{i_t^s}(\tilde{\mat{U}}^{s-1}) \right) + P_{\gamma}^{\scriptsize \mat{U}_{t-1}^{s} \leftarrow \tilde{\mat{U}}^{s-1}} \left(\gradf(\tilde{\mat{U}}^{s-1})\right), 
\end{eqnarray}
where $P_{\gamma}^{\scriptsize \mat{U}_{t-1}^{s} \leftarrow \tilde{\mat{U}}^{s-1}}(\cdot)$ represents a parallel-translation operator from $\tilde{\mat{U}}^{s-1}$ to $\mat{U}_{t-1}^{s}$ on the Grassmann manifold defined in (\ref{Eq:parallel_translation}). Furthermore, for this parallel translation, we need to calculate the tangent vector from $\tilde{\mat{U}}^{s-1}$ to $\mat{U}_{t-1}^{s}$. This is given by \changeBM{the} logarithm mapping defined in (\ref{Eq:logarithm_map}). Consequently, the final update rule of R-SVRG is defined as $\mat{U}_{t}^{s} =  {\rm Exp}_{\scriptsize \mat{U}_{t-1}^{s}}( - \eta \xi_t^s)$. It should be noted that the modified direction $\xi_t^s$ is also a Riemannian stochastic gradient of $f$ at $\mat{U}_{t-1}^{s}$.

Conditioned on $\mat{U}_{t-1}^s$, we take the expectation with respect to $i_t^s$ and obtain
\begin{eqnarray*}
\mathbb{E}_{i_t^s}[\xi_t^s] & = & 
\mathbb{E}_{i_t^s}[\gradf_{i_t^s}(\mat{U}_{t-1}^{s})] - 
P_{\gamma}^{\scriptsize \mat{U}_{t-1}^{s} \leftarrow \tilde{\mat{U}}^{s-1}} 
\left(
\mathbb{E}_{i_t^s}[\gradf_{i_t^s}(\tilde{\mat{U}}^{s-1}) ] -
\gradf(\tilde{\mat{U}}^{s-1})
\right) \nonumber\\
& = & \gradf(\mat{U}_{t-1}^{s}) -
P_{\gamma}^{\scriptsize \mat{U}_{t-1}^{s} \leftarrow \tilde{\mat{U}}^{s-1}} 
\left(
\gradf(\tilde{\mat{U}}^{s-1})  -
\gradf(\tilde{\mat{U}}^{s-1})
\right)  \nonumber\\
& = & \gradf(\mat{U}_{t-1}^{s}).
\end{eqnarray*}

The theoretical analysis of convergence of the Euclidean SVRG algorithm assumes that the beginning vector $\mat{U}_0^s$ of $s$-th epoch is set to be the average \changeHK{or randomly selected value} of the $(s\!\!-\!\!1)$-th epoch \cite[Figure~1]{Johnson_NIPS_2013_s}. \changeHK{On the other hand}, the set of the last vector in the $(s\!\!-\!\!1)$-th epoch, i.e., $\mat{U}_{m_{s-1}}^{s-1}$ shows the superior performances on the Euclidean SVRG algorithm. Therefore, for our local convergence rate analysis in {\bf Theorem \ref{Thm:LocalConvergence}}, this paper also uses, as {\bf option I}, the mean value of $\tilde{\mat{U}}^{s} = g_{m_s}(\mat{U}_1^s, \ldots \mat{U}_{m_s}^s)$ as $\tilde{\mat{U}}^s$, where $g_n(\mat{U}_1, \ldots, \mat{U}_n)$ is the Karcher mean on the Grassmann manifold. This option can also simply choose $\tilde{\mat{U}}^{s}=\mat{U}_t^s$ for $t \in \{1, \ldots, m_s\}$ at random. In addition, as {\bf option II}, we can also use the last vector in the $(s\!\!-\!\!1)$-th epoch, i.e., $\tilde{\mat{U}}^{s}=\mat{U}_{m_s}^s$
%
%
The overall algorithm with a fixed step-size is summarized in Algorithm \ref{Alg:R-SVRG}. 

\begin{algorithm}
\caption{Algorithm for R-SVRG with a fixed step-size.}
\label{Alg:R-SVRG}
\begin{algorithmic}[1]
\REQUIRE{Update frequency $m_s>0$ and step-size $\eta>0$.}
\STATE{Initialize $\tilde{\mat{U}}^0$.}
\FOR{$s=1,2, \ldots$} 
\STATE{Calculate the Riemannian full gradient $\gradf(\tilde{\mat{U}}^{s-1})$}.
\STATE{Store $\mat{U}_0^s = \tilde{\mat{U}}^{s-1}$.}
	\FOR{$t=1,2, \ldots, m_s$} 
	\STATE{Choose $i_t^s \in \{1, \ldots, N\}$ uniformly at random.}
	\STATE{Calculate the tangent vector $\zeta$ from $\tilde{\mat{U}}^{s-1}$ to $\mat{U}_{t-1}^{s}$ by logarithm mapping in (\ref{Eq:logarithm_map}).}
	\STATE{Calculate the modified Riemannian stochastic gradient $\xi_t^s$ in (\ref{Eq:R-SVRG-Grad-paralleltrans}) by parallel-translating $\gradf(\tilde{\mat{U}}^{s-1})$ and $\gradf_{i_t^s}(\tilde{\mat{U}}^{s-1})$ \changeHS{along} $\zeta$ in (\ref{Eq:parallel_translation}) as\\
	$\xi_t^s = \gradf_{i_t^s}(\mat{U}_{t-1}^{s}) -  P_{\gamma}^{{\scriptsize \mat{U}}_{t-1}^{s} \leftarrow \tilde{{\scriptsize \mat{U}}}^{s-1}} \left(\gradf_{i_t^s}(\tilde{\mat{U}}^{s-1})  -  \gradf(\tilde{\mat{U}}^{s-1})\right)$.}
	\STATE{Update $\mat{U}_t^s$ from $\mat{U}_{t-1}^s$ as $\mat{U}_t^s = {\rm Exp}_{\scriptsize \mat{U}_{t-1}^s}\left(- \eta \xi_t^s \right)$ with the exponential mapping (\ref{Eq:exponential_map}).}	
	\ENDFOR
	\STATE{{\bf option I}: $\tilde{\mat{U}}^{s}=g_{m_s}(\mat{U}_1^s,\ldots,\mat{U}_{m_s}^s)$ (or $\tilde{\mat{U}}^{s}=\mat{U}_t^s$ for randomly chosen $t \in \{1, \ldots, m_s\}$).}	
	\STATE{{\bf option II}: $\tilde{\mat{U}}^{s}=\mat{U}^s_{m_s}$.}
\ENDFOR
\end{algorithmic}
\end{algorithm}

Additionally, the variants of the variance reduced SGD need full gradient calculation every epoch at the beginning. This poses a bigger overhead than the ordinal SGD algorithm at the beginning of the process, and eventually, this causes {\it cold-start} property on them. \changeBM{To avoid this, \cite{Konecny_arXiv_2013} in the Euclidean space proposes to use standard SGD updating only for first epoch.} This paper also adopts this simple modification of R-SVRG, denoted as R-SVRG+. We do not analyze this extension and leave this as an open problem. 

As mentioned earlier, each iteration of R-SVRG has double loops to reduce the variance of the modified stochastic gradient $\xi_t^s$. $s$-th epoch, i.e., outer loop, requires $N+2m_s$ gradient evaluations, where $N$ is for the full gradient $\gradf(\tilde{\mat{U}}^{s-1})$ at the beginning of each $s$-th epoch and $2m_s$ is for inner iterations since each inner step needs two gradient evaluations, i.e., $\gradf_{i_t^s}(\mat{U}_{t-1}^{s})$ and $\gradf_{i_t^s}(\tilde{\mat{U}}^{s-1})$. However, if $\gradf_{i_t^s}(\tilde{\mat{U}}^{s-1})$ for each sample are stored at the beginning of $s$-th epoch like SAG, the evaluations for each inner loop result in $m_s$. Finally, $s$-th epoch requires $N+m_s$ evaluations. It is natural to choose $m_s$ to be the same order of $N$, but slightly larger (for example $m_s = 5N$ for non-convex problems is suggested in \cite{Johnson_NIPS_2013_s}).

\section{Main result: convergence analysis}
\label{Sec:Analysis}

In this section, we provide the results of our convergence analysis.
The actual proofs of all the theorems and lemmas are given in the supplementary material.

We first introduce a global convergence result under a {\it decay step-size} below. 
\begin{Thm} 
Consider {\bf Algorithm 1} on a connected Riemannian manifold $\mathcal{M}$ \changeHK{of which} injectivity radius \changeHK{is} uniformly bounded from below by $I > 0$. \changeHK{Suppose} \changeBM{that} the sequence of step-sizes $(\eta_t^s)_{m_s \geq t \geq 1, s \geq 1}$ satisfies the condition that $\sum (\eta_t^s)^2< \infty$ and $\sum \eta_t^s=+\infty$. Suppose there exists a compact set $K$ such that $w_t^s \in K$ for all $t \geq 0$. We also suppose that the gradient is bounded on 
$K$, i.e., there exists $A > 0$ such that for all $w \in K$ and $n \in \{1,2,\ldots,n\}$, \changeHK{and} we have $\| \gradf (w) \| \leq A/3$ \changeHS{and $\| \gradf_{n}(w)\| \le A/3$}. Then $f(w_t^s)$ converges a.s. and $\gradf (w_t^s) \rightarrow 0$ a.s. \changeHK{.}
\end{Thm} 
\begin{proof}
Note that $\xi_t^s \le A$ from the triangle inequality.
The proof is done by bounding above the expectation of $f(w_{t+1}^s) - f(w_t^s)$ and $\| \gradf (w_{t+1}^s)\|^2 - \| \gradf (w_{t}^s)\|^2$.
See {\bf Theorem B.2} for details of the proof.
\end{proof}

Then, we show a local convergence rate analysis. For this purpose, we first show a lemma \changeBM{that upper bounds the variance of $\xi_t^s$}. Subsequently, the local convergence rate theorem for R-SVRG in {\bf Algorithm 1} is given. It should be also noted that the lemma and theorem in this section hold for any compact manifold. In addition, this analysis holds under a {\it fixed step-size} setup. Here, we assume throughout the following analysis that the functions $f_n$ are $\beta$-{\it Lipschitz continuously differentiable} (See {\bf Assumption 1} in Section \changeHK{{\bf B}}). 
\begin{Lem}
\label{Lem:UpperBoundVariance}
Let $\mathbb{E}_{i_t^s}[\cdot]$ be the expectation with respect to \changeHK{the distribution of} the random choice of $i_t^s$.
When each $\gradf_{n}$ is $\beta$-Lipschitz continuously differentiable, the upper bound of the variance of $\xi_t^s$ is given by
\begin{eqnarray*}
	\mathbb{E}_{i_t^s}[\| \xi_t^s \|^2] &\leq &
\beta^2 (14({\rm dist}(w_{t-1}^s,w^*))^2 + 8{\rm dist}(\tilde{w}^{s-1},w^*))^2  ).
\end{eqnarray*}
\end{Lem}
\begin{proof}
The proof is analogous to that of SVRG algorithm in the Euclidean \changeBM{space}. However, the distance evaluations of points should be done appropriately on the corresponding same tangent space using parallel translation. The actual proof is in  {\bf Lemma C.3} of the supplementary material file. 
\end{proof}

Lemma \ref{Lem:UpperBoundVariance} implies that the variance of $\xi_t^s$ converges to zero when both $\mat{U}_t^s$ and $\tilde{\mat{U}}^{s-1}$ converge to $\mat{U}^*$. Finally, we provide the main theorem of this paper for the local convergence rate of R-SVRG.
\begin{Thm}
\label{Thm:LocalConvergence}
Let $\mathcal{M}$ be the Grassmann manifold and $\mat{U}^* \in \mathcal{M}$ be a non-degenerate local minimizer of $f$ (i.e., ${\rm grad} f(\mat{U}^*)=0$ and \changeHS{the Hessian ${\rm Hess}f(\mat{U}^*)$ of $f$ at $\mat{U}^*$} is positive definite). \changeHS{Assume that there exists a convex neighborhood $\mathcal{U}$ of $\mat{U}^* \in {\mathcal{M}}$ and a positive real number $\sigma$ such that the smallest eigenvalue of the Hessian of $f$ at each $\mat{U} \in \mathcal{U}$ is not less than $\sigma$}.
When each $\gradf_{n}$ is $\beta$-Lipschitz continuously differentiable \changeHS{and $\eta > 0$ is sufficiently small such that $0 < \eta(\sigma - 14 \eta \beta^2) < 1$}, 
it then follows that for any sequence $\{\tilde{\mat{U}}^s\}$ generated by the algorithm converging to $\mat{U}^*$, there exists $K>0$ such that for all $s>K$,
\begin{eqnarray*}
\mathbb{E}[({\rm dist}(\tilde{\mat{U}}^s,\mat{U}^*))^2]& \leq & 
\frac{4(1+ 8 m \eta^2  \beta^2)}{ \eta m (\sigma - 14 \eta \beta^2)} 
\mathbb{E}[({\rm dist}(\tilde{\mat{U}}^{s-1},\mat{U}^*))^2].
\end{eqnarray*}
\end{Thm}
\begin{proof}
The proof starts with bounding above the expectation of the distance between $\mat{U}_{t}^s$ and $\mat{U}^{*}$ with respect to the random choice of $i_t^s$, where the curvature of the Grassmann manifold and {\bf Lemma 6} in \cite{Zhang_JMLR_2016_s}, which corresponds to the law of cosines in the Euclidean space, are fully used. See {\bf Theorem C.5} for the complete proof.
\end{proof}

\section{Numerical comparisons}
\label{Sec:NumericalComparison}

This section compares the performance of R-SVRG(+) with the Riemannian extension of SGD, i.e., R-SGD, where the Riemannian stochastic gradient algorithm is $\gradf_{i_t^s}(\mat{U}_{t-1}^{s})$ instead of $\xi_t^s$ in (\ref{Eq:R-SVRG-Grad-paralleltrans}). We also compare with R-SD, which is the Riemannian steepest descent algorithm with the backtracking line search \cite[Chapters~4]{Absil_OptAlgMatManifold_2008}. \changeBM{We consider both {\it fixed} step-size as well as {\it decay} step-size sequences}. The decay step-size sequence uses the decay $\eta_k = \eta_0(1+ \eta_0 \lambda \lfloor k/m_s \rfloor)^{-1}$ where $k$ is the number of iterations used. We select ten choices of $\eta_0$, and consider three $\lambda = \{10^{-1}, 10^{-2}, 10^{-3}\}$. \changeBM{In addition, since the global convergence needs a decay step-size condition and the local convergence rate analysis holds for a fixed step-size (Section \ref{Sec:Analysis}),} we consider a {\it hybrid} step-size sequence that follows the decay step-size at less than $s_{TH}$ epoch, and subsequently switches to a fixed step-size. All experiments use $s_{TH}=5$ in this experiment. $m_s=5N$ is also fixed by following \cite{Johnson_NIPS_2013_s}, and  batch-size is fixed to 10. In all the figures, the $x$-axis is the computational cost measured by the number of gradient computations divided by $N$. Algorithms are initialized randomly and are stopped when either the stochastic gradient norm is below $10^{-8}$ or the number of iterations exceeds $100$. Additional numerical experiments are shown in Section \changeHK{{\bf C}} of the supplementary material file. It should be noted that all results except R-SD are the best-tuned results. All simulations are performed in Matlab on a 2.6 GHz Intel Core i7 PC with 16 GB RAM. 

\begin{figure}[h]
\begin{center}
	\begin{minipage}[t]{0.32\textwidth}
	\begin{center}
		\includegraphics[width=\textwidth]{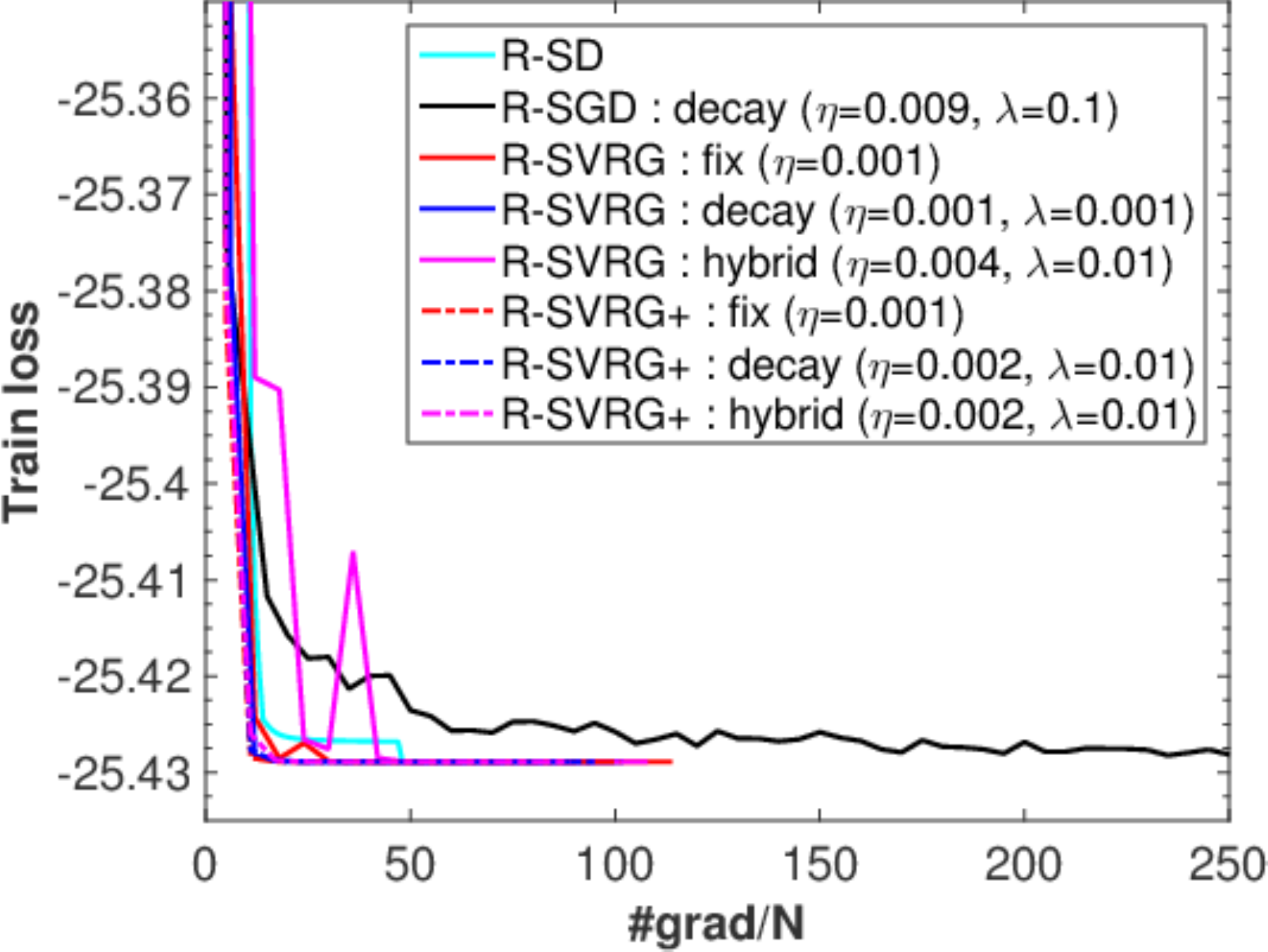}\\
		
		{\small (a) Train loss (enlarged).}
		
	\end{center} 
	\end{minipage}
	\hspace*{-0.1cm}
	\begin{minipage}[t]{0.32\textwidth}
	\begin{center}
		\includegraphics[width=\textwidth]{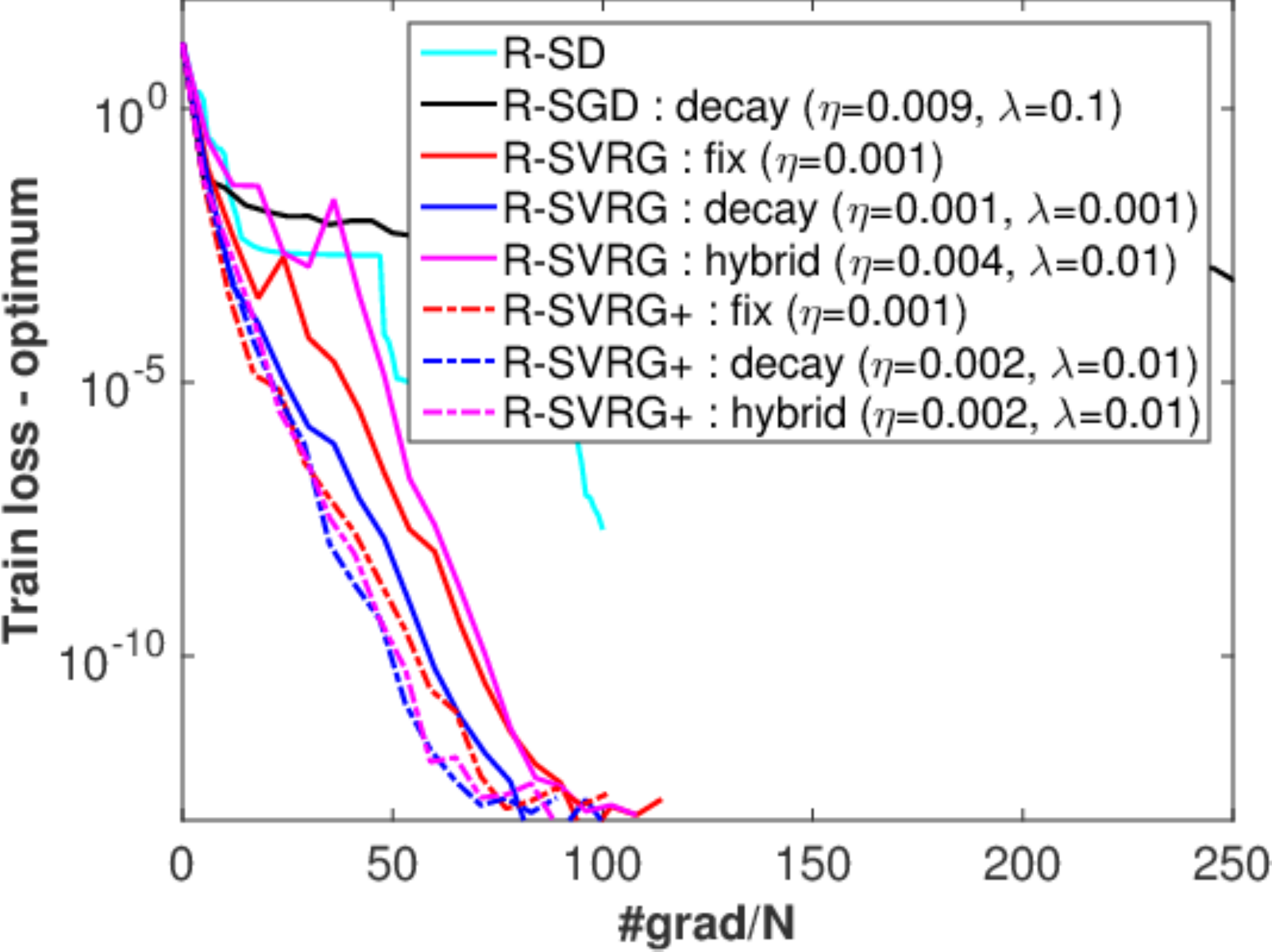}\\
		
		{\small (b) Optimality gap. }
		
	\end{center} 
	\end{minipage}
	\hspace*{-0.1cm}
	\begin{minipage}[t]{0.32\textwidth}
	\begin{center}
		\includegraphics[width=\textwidth]{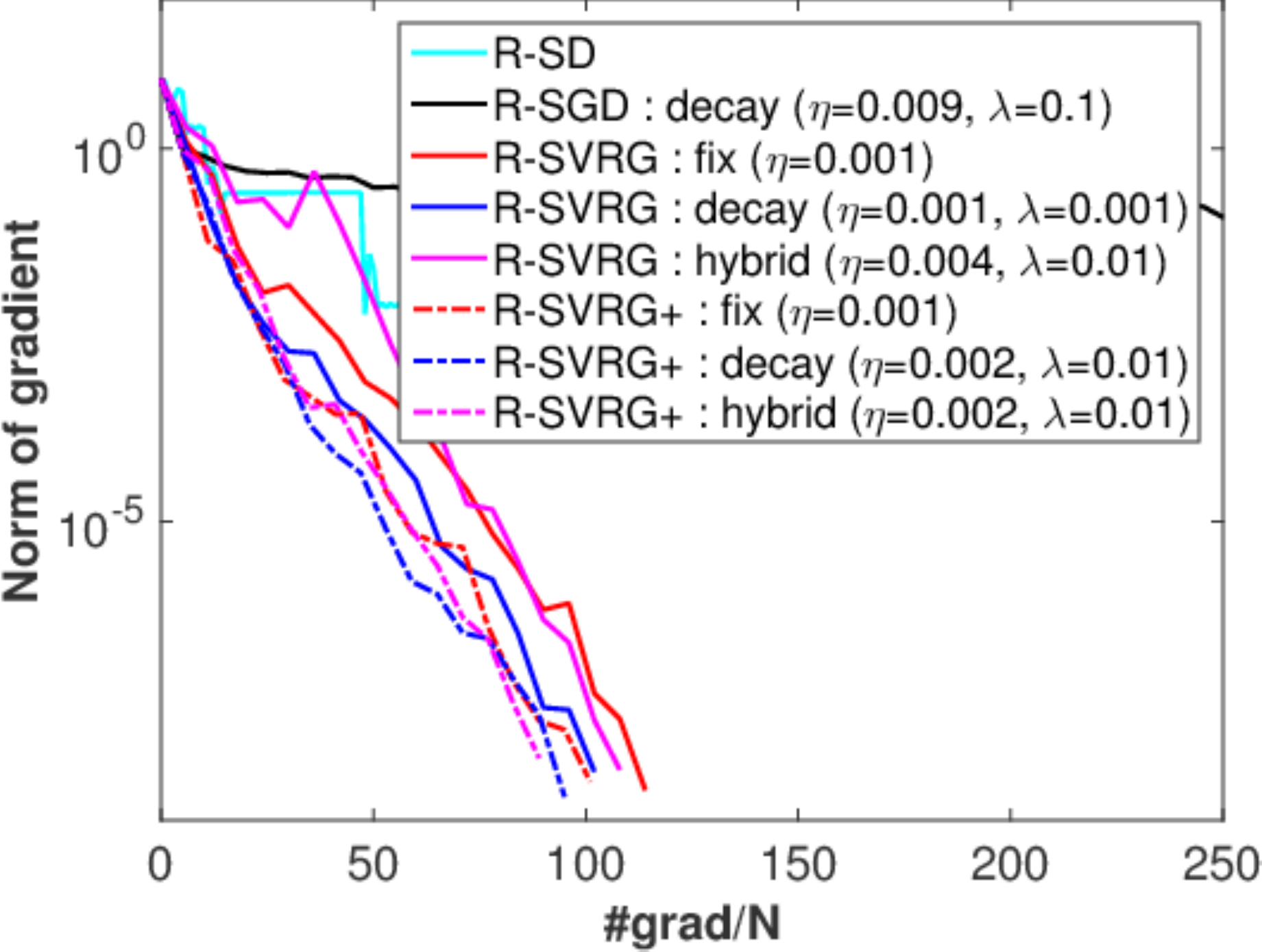}\\
		
		{\small (c) Norm of gradient.}
		
	\end{center} 
	\end{minipage}
\caption{Performance evaluations on PCA problem. }
\label{fig:PCA_results}
\end{center}		
\end{figure}

{\bf PCA problem (\ref{Eq:PCA}).} We first consider the PCA problem. \changeBM{Figures} \ref{fig:PCA_results}(a)-(c) show the results of the train loss, {\it optimality gap}, and the norm of gradient, respectively, where $N=10000$, $d=20$, and $r=5$. $\eta_0$ is $\{10^{-3},2\times10^{-3}, \ldots, 10^{-2}\}$. The optimality gap evaluates the performance against the minimum loss, which is obtained by the Matlab function {\tt pca}. Figure \ref{fig:PCA_results}(a) shows the enlarged results of the train loss, where all algorithms of R-SVRG(+) yield better convergence properties. Among the step-size sequences of R-SVRG(+), the hybrid sequence shows the best performance among all. Between R-SVRG and R-SVRG+, the latter shows superior performance for all step-size \changeBM{sequences}. For the optimality gap plots in Figure \ref{fig:PCA_results}(b), the results follow similar trends as those of train loss plots. In Figure \ref{fig:PCA_results}(c), while the gradient norm of SGD stays at higher values, those of R-SVRG and R-SVRG+ converge to lower values in all cases.

\begin{figure}[htbp]
\begin{center}
	\begin{minipage}[t]{0.32\textwidth}
	\begin{center}
		\includegraphics[width=\textwidth]{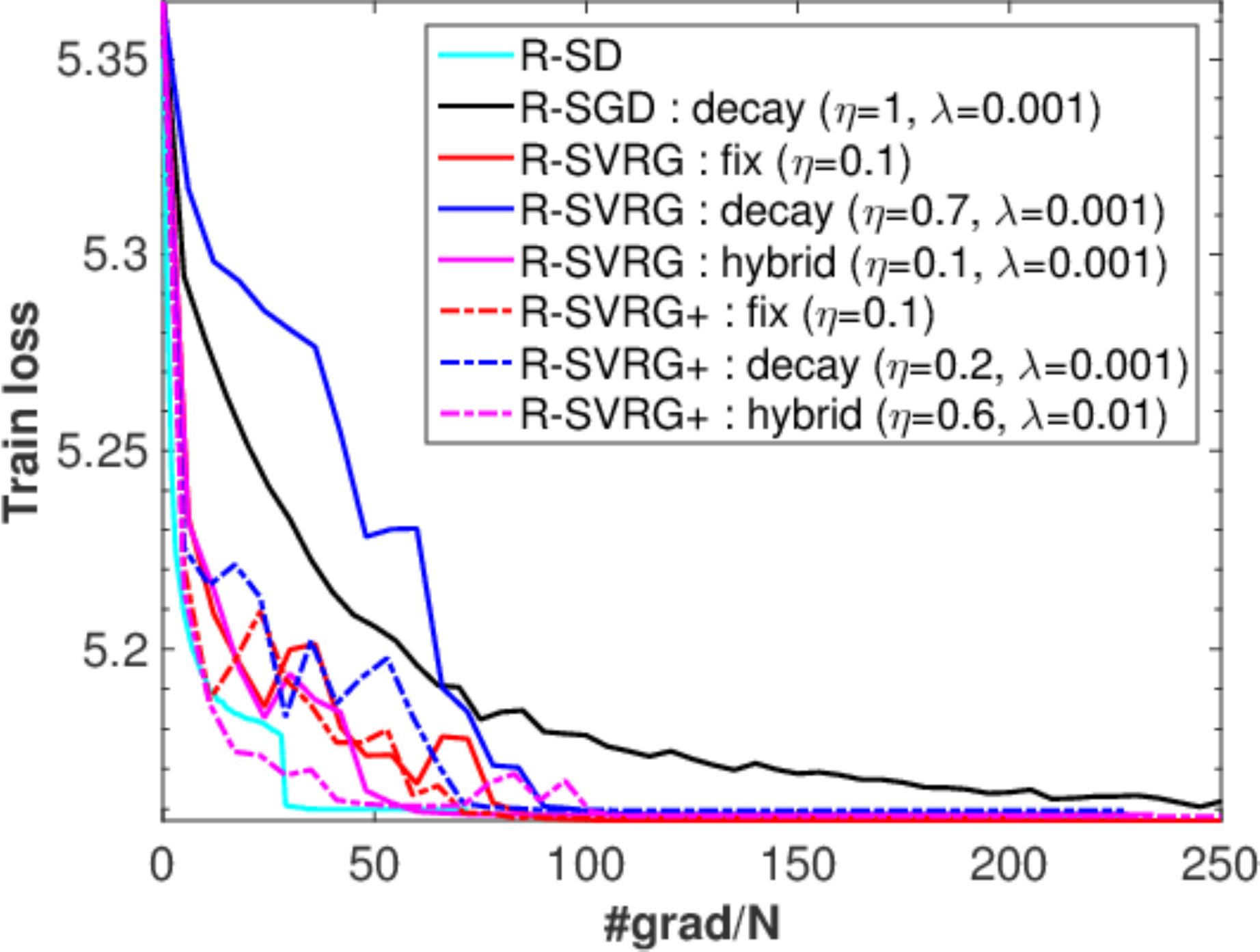}\\
		
		{\small (a) Train loss.}
		
	\end{center} 
	\end{minipage}
	\hspace*{-0.1cm}
	\begin{minipage}[t]{0.32\textwidth}
	\begin{center}
		\includegraphics[width=\textwidth]{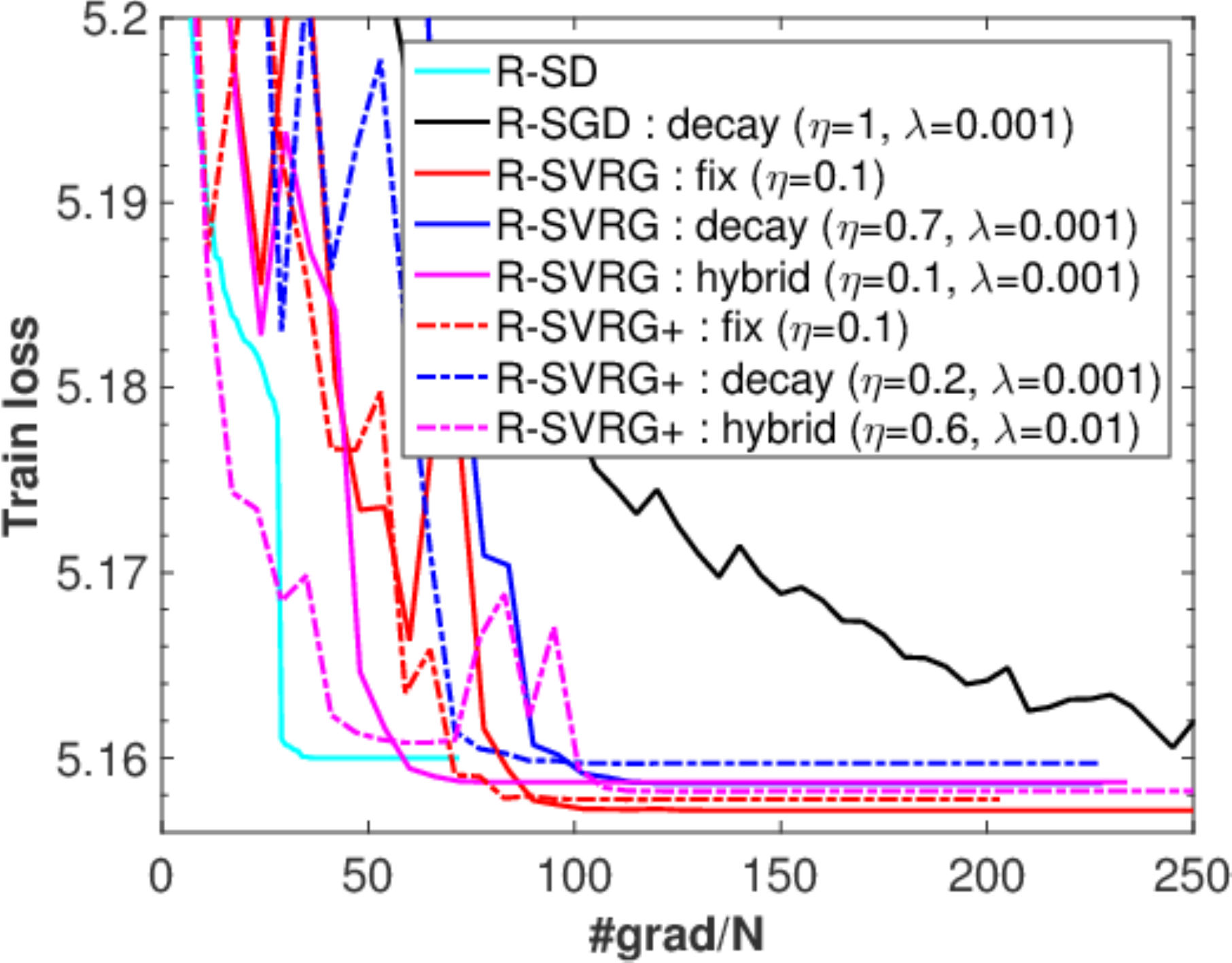}\\
		
		{\small (b) Train loss (enlarged). }
		
	\end{center} 
	\end{minipage}
	\hspace*{-0.1cm}
	\begin{minipage}[t]{0.32\textwidth}
	\begin{center}
		\includegraphics[width=\textwidth]{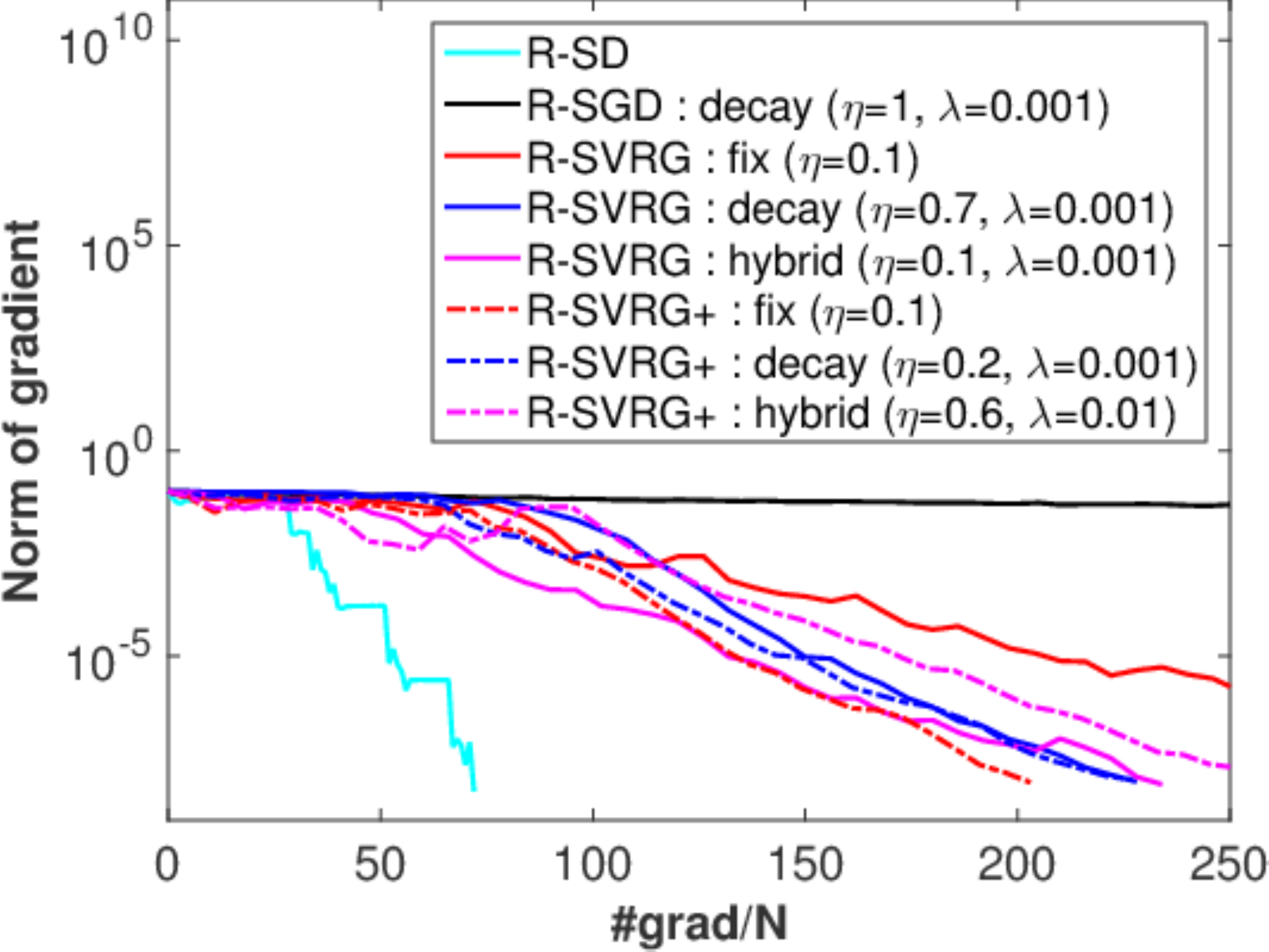}\\
		
		{\small (c) Norm of gradient.}
		
	\end{center} 
	\end{minipage}
\caption{Performance evaluations on Karcher mean problem.}
\label{fig:KarcherMean_results}
\end{center}

\begin{center}
	\begin{minipage}[t]{0.32\textwidth}
	\begin{center}
		\includegraphics[width=\textwidth]{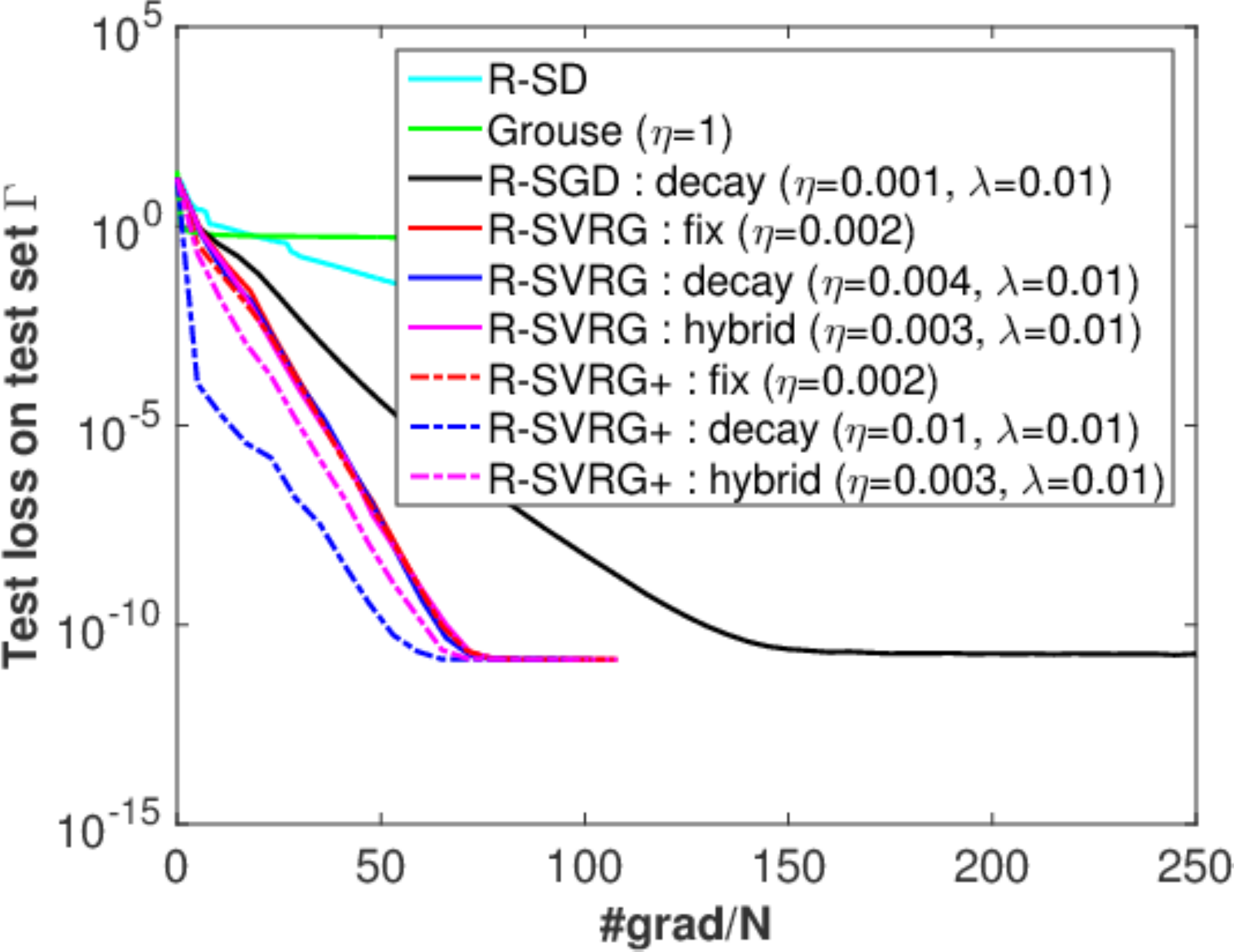}\\
		
		{\small (a) Test loss (synthetic).}
		
	\end{center} 
	\end{minipage}
	\hspace*{-0.1cm}
	\begin{minipage}[t]{0.32\textwidth}
	\begin{center}
		\includegraphics[width=\textwidth]{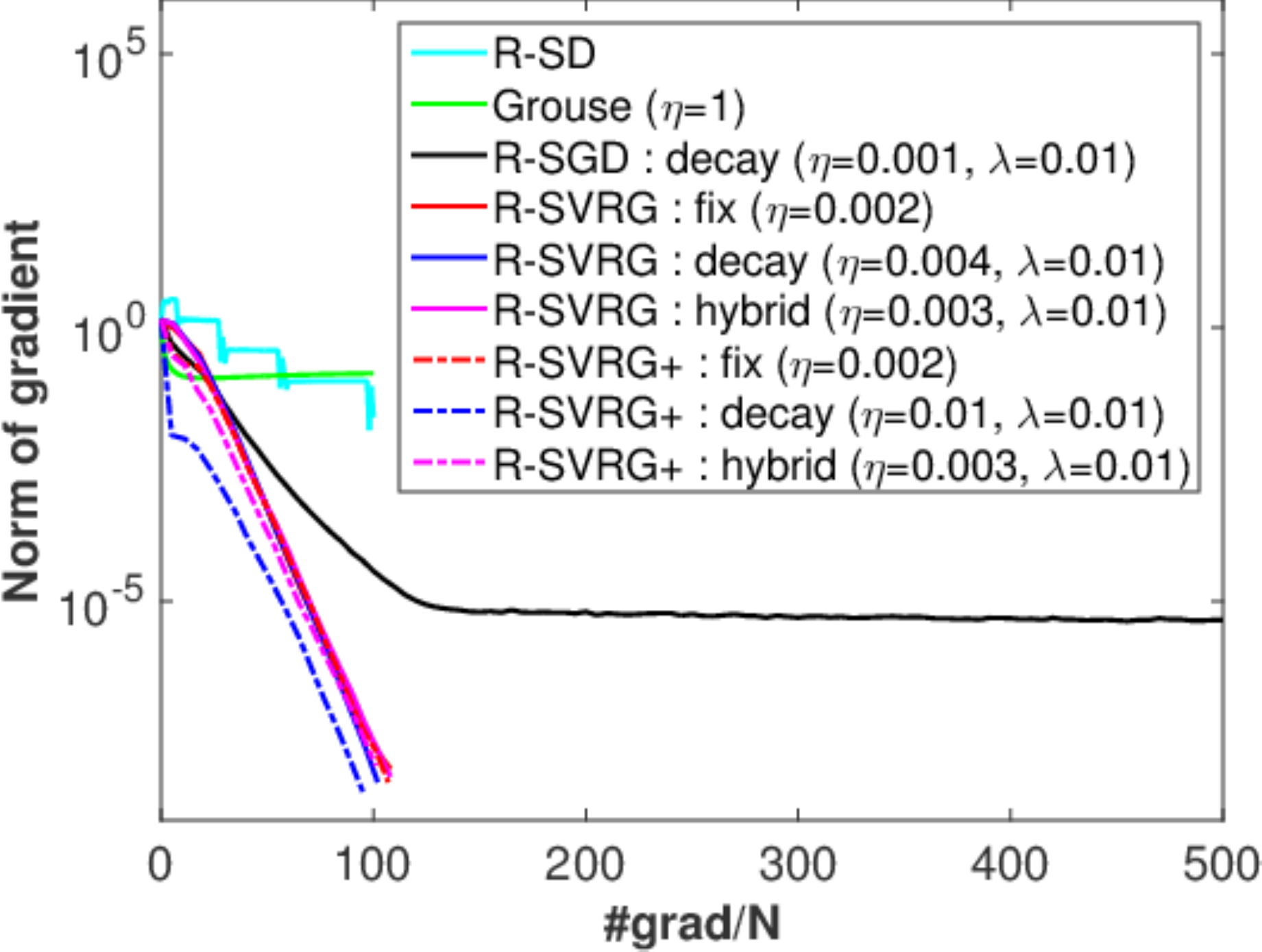}\\
		
		{\small (b) Norm of gradient (synthetic). }
		
	\end{center} 
	\end{minipage}
	\hspace*{-0.1cm}
	\begin{minipage}[t]{0.32\textwidth}
	\begin{center}
		\includegraphics[width=\textwidth]{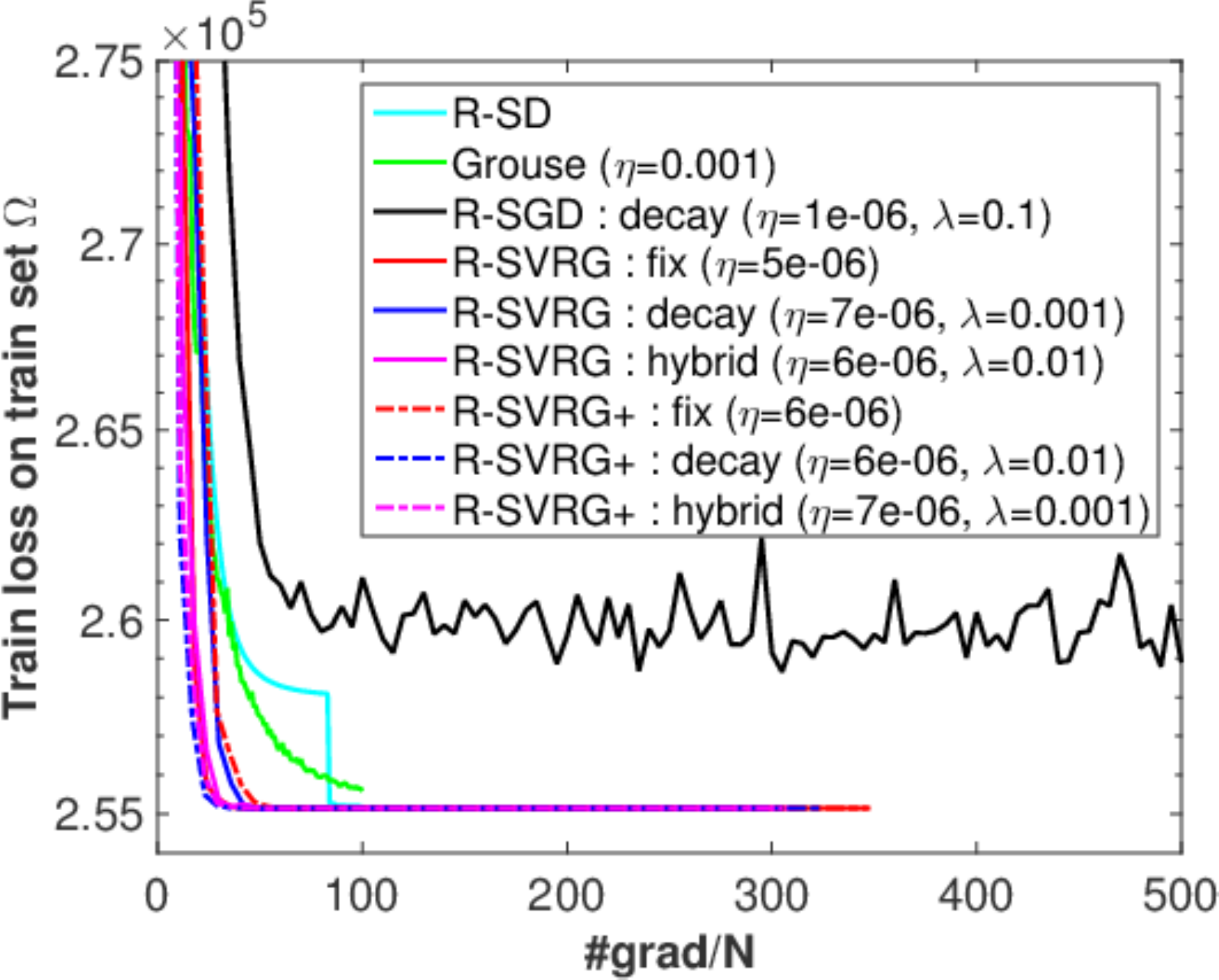}\\
		
		{\small (c) Train loss (Jester).}
		
	\end{center} 
	\end{minipage}
	\vspace*{0.4cm}
	
	\hspace*{-0.1cm}
	\begin{minipage}[t]{0.32\textwidth}
	\begin{center}
		\includegraphics[width=\textwidth]{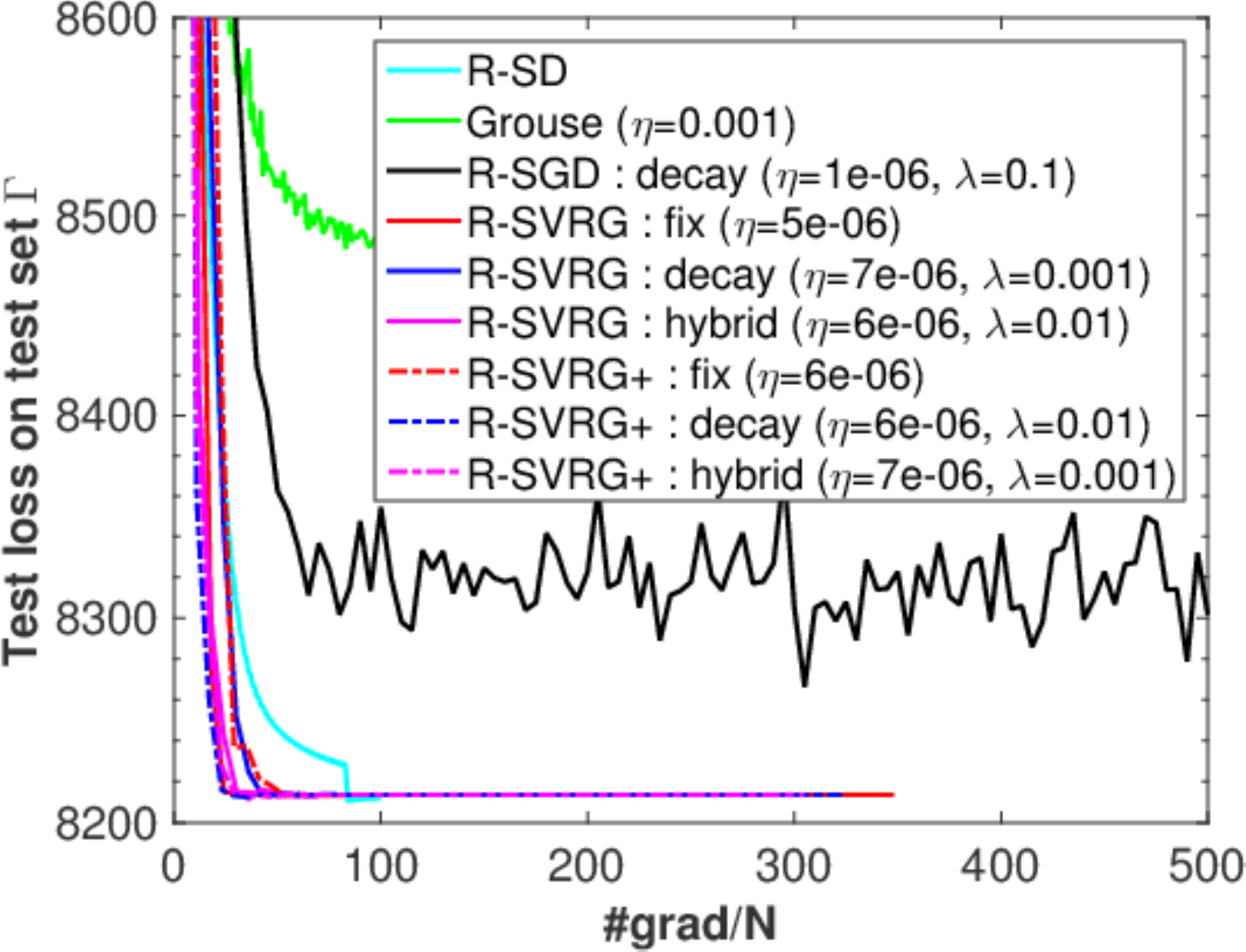}\\
		
		{\small (d) Test loss (Jester).}
		
	\end{center} 
	\end{minipage}
	\begin{minipage}[t]{0.32\textwidth}
	\begin{center}
		\includegraphics[width=\textwidth]{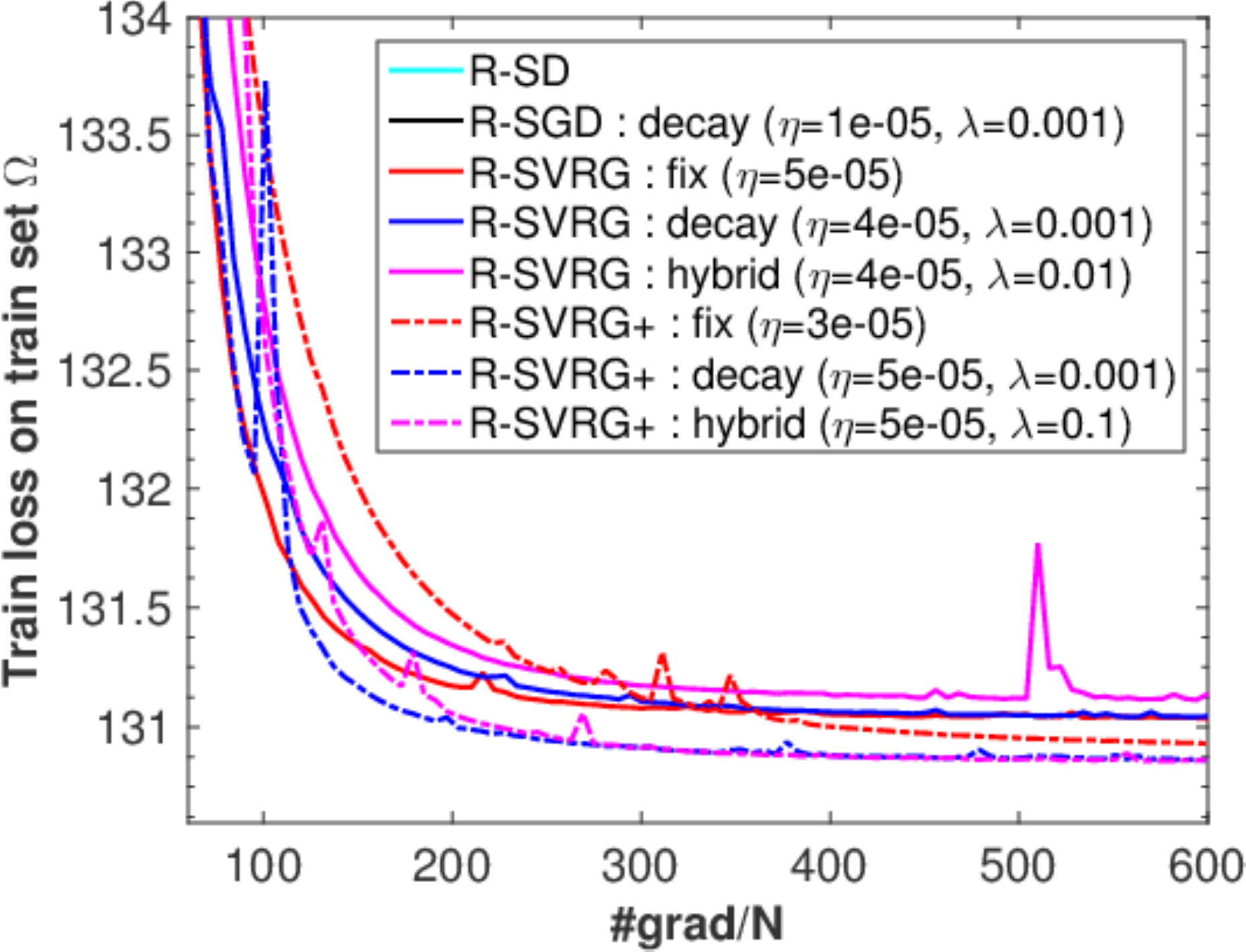}\\
		{\small (e) Train loss (MovieLens-1M).}
	\end{center} 
	\end{minipage}
	\hspace*{-0.1cm}
	\begin{minipage}[t]{0.32\textwidth}
	\begin{center}
		\includegraphics[width=\textwidth]{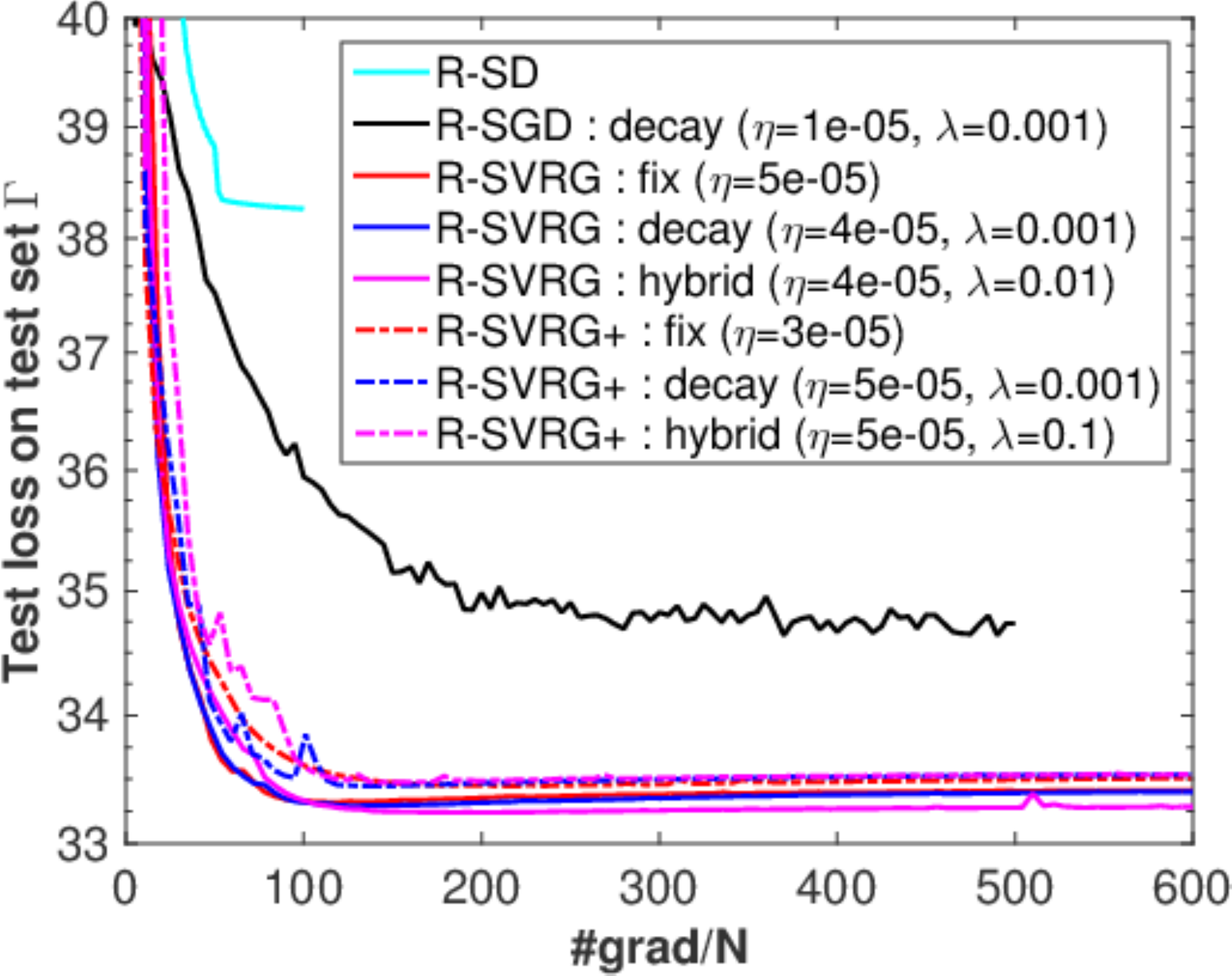}\\
		{\small (f) Test loss (MovieLens-1M).}
	\end{center} 
	\end{minipage}	
\caption{Performance evaluations on low-rank matrix completion problem.}
\label{fig:MC_results}
\end{center}	
\end{figure}

{\bf Karcher mean problem (\ref{Eq:KarcherMean}).} We compute the Karcher mean of $N$ number of $r$-dimensional subspaces in $\mathbb{R}^d$. Figures \ref{fig:KarcherMean_results}(a)-(c) show the results of the train loss, the enlarged train loss, and the norm of gradient, respectively, where $N=1000$, $d=300$, and $r=5$. The ten choices of $\eta_0$ are $\{0.1,0.2, \ldots, 1.0\}$. R-SVRG(+) outperforms R-SGD, and the final loss of R-SVRG(+) is less than that of R-SD. It should be noted that R-SVRG+ with the fixed and decay step-sizes decreases faster in the beginning, but eventually, R-SVRG converges to lower losses.

{\bf Matrix completion problem (\ref{Eq:MC}).} The proposed algorithms are also compared with Grouse \cite{Balzano_arXiv_2010_s}, a state-of-the-art stochastic descent algorithm on the Grassmann manifold. We first consider a synthetic dataset with $N=5000$, $d=500$ with rank $r=5$. \changeHK{Each experiment is} initialized randomly as suggested in \cite{Kressner_BIT_2014_s}. The ten choices of $\eta_0$ are $\{10^{-3},2\times10^{-3}, \ldots, 10^{-2}\}$ for R-SGD and R-SVRG(+) and $\{0.1,0.2, \ldots, 1.0\}$ for Grouse. This instance considers the loss on a test set $\Gamma$, which is different from the training set $\Omega$. We also \changeHK{consider the lower condition number (CN) of the matrix, where the CN represents} the ratio of the largest to the lowest singular value a matrix. 
This instance uses CN=$5$. The over-sampling ratio (OS) is $5$, where the OS \changeHK{expresses} the \changeHK{known} number of entries. An OS of $5$ implies that $5(N+d-r)r$ \changeHK{samples are} randomly and uniformly \changeHK{sampled}  out of the total $Nd$ entries \changeHK{as known} entries. \changeBM{Figures} \ref{fig:MC_results}(a) and (b) show the results of loss on test set $\Gamma$ and the norm of gradient, respectively. The results show the superior performance of our proposed algorithms.

Next, we consider the Jester dataset 1 \cite{Goldberg_IR_2001_s} \changeHK{which consists} of ratings of $100$ jokes \changeHK{evaluated by} $24983$ users. Each rating is a real number \changeHK{ranging from} $-10$ \changeHK{to} $10$. We randomly extract two ratings per user as the training set $\Omega$ and test set $\Gamma$. The algorithms are run by fixing the rank to $r=5$ with random initialization. $\eta_0$ is chosen from $\{10^{-6},2\times10^{-6}, \ldots, 10^{-5}\}$ for SGD and SVRG(+) and $\{10^{-3},2\times10^{-3}, \ldots, 10^{-2}\}$ for Grouse. Figures \ref{fig:MC_results}(c) and (d) show the superior performance of R-SVRG(+) on both the train and test sets.

As a final test, we compare the algorithms on the MovieLens-1M dataset, which is downloaded from {\tt http://grouplens.org/datasets/movielens/}. The dataset has a million ratings corresponding to $6040$ users and $3952$ movies. $\eta_0$ is chosen from $\{10^{-5},2\times10^{-5}, \ldots, 10^{-4}\}$. Figures \ref{fig:MC_results}(e) and (f) show the results on the train and test set of all the algorithms except Grouse, which faces issues with convergence on this datatset. R-SVRG(+) shows much faster convergence speed than others, and R-SVRG is better than R-SVRG+ in terms of the final test loss for all step-size algorithms.

\section{Conclusion}
\changeBM{We have proposed a Riemannian stochastic variance reduced gradient algorithm (R-SVRG). The proposed algorithm stems from the variance reduced gradient algorithm in the Euclidean \changeBM{space, but is now extended to Riemannian manifolds}. The central difficulty of averaging, addition, and subtraction of multiple gradients on a Riemannian manifold is handled with classical notion of parallel transport. We proved that R-SVRG generates globally convergent sequences with a decay step-size condition and is locally linearly convergent with a fixed step-size under some natural assumptions. We have shown the developments on the Grassmann manifold. Numerical comparisons on three popular problems on the Grassmann manifold suggested the superior performance of R-SVRG on various different benchmarks.}

\small
\bibliographystyle{unsrt}
\bibliography{stochastic_online_learning,matrix_tensor_completion}

\clearpage
\appendix

\renewcommand\thefigure{A.\arabic{figure}}  
\setcounter{figure}{0} 

\renewcommand\thetable{A.\arabic{table}}  
\setcounter{table}{0} 

\renewcommand{\theequation}{A.\arabic{equation}}
\setcounter{equation}{0}

\hrule height 1mm depth 0mm width 140mm
\vspace{0.2cm}

\begin{center}
{\bf\LARGE Supplementary material}
\end{center}
\vspace{0.3cm}

\hrule height 0.2mm depth 0mm width 140mm

\section{Global convergence analysis}
\label{Sec:ConvergenceAnalysis}

We assume \changeHK{that the sequence of step-sizes $(\eta_t^s)_{t \geq 1, s \geq 1}$ satisfies}
\begin{eqnarray}
\label{Eq:StepsizeCondition}
\sum (\eta_t^s)^2 \ <\  \infty {\ \ \ \rm and\ \ \ } \sum \eta_t^s \ =\  +\infty.
\end{eqnarray}

We also note the following proposition.
\begin{Prop}[\cite{Fisk_1965_s}]
\label{Prop:FiskPro}
Let $(X_n)_{n \in \mathbb{N}}$ be a non-negative stochastic process \changeHK{that has} bounded positive variations, i.e., $\sum_0^{\infty} \mathbb{E}([\mathbb{E}(X_{n+1} - X_n) | \mathcal{F}_n]^{+}) < \infty$. Then\changeHK{, we call such a process as} a quasi-martingale, \changeHK{where} 
\begin{eqnarray*}
\sum_{n=0}^{\infty} |\mathbb{E}[ X_{n+1} - X_n| \mathcal{F}_n] | < \infty \ \ \ a.s.\ , {\rm and\ } X_n\ {\rm converges\ } a.s. \changeHK{.}
\end{eqnarray*}
\end{Prop}

Now, we \changeHK{prove that the proposed algorithm converges a.s.} under some assumptions when the \changeHK{iteration sequences} are guaranteed to \changeHK{stay} in a compact set. \changeHK{It should be noted that if} $\mathcal{M}$ is compact, especially if $\mathcal{M}$ is the Grassmann manifold, \changeHK{this assumption is satisfied.}
\begin{Thm}
\label{Thm:GlobalConvergence}
Consider {\bf Algorithm 1} on a connected Riemannian manifold $\mathcal{M}$ \changeHK{of which} injectivity radius \changeHK{is} uniformly bounded from below by $I > 0$. \changeHK{Suppose} that the sequence of step-sizes $(\eta_t^s)_{m_s \geq t \geq 1, s \geq 1}$ satisfies the condition (\ref{Eq:StepsizeCondition}). \changeHK{Then, supposing that} there exists a compact set $K$, \changeHK{we assume} $w_t^s \in K$ for all $t \geq 0$. \changeHK{Furthermore, we assume} that the gradient \changeHK{$\gradf (w)$} is bounded on $K$, i.e., there exists $A > 0$ such that for all $w \in K$ and $n \in \{1,2,\ldots,n\}$, \changeHK{and} we have $\| \gradf (w) \| \leq A/3$ \changeHS{and $\| \gradf_{n}(w)\| \le A/3$}. Then $f(w_t^s)$ converges a.s. and $\gradf (w_t^s) \rightarrow 0$ a.s.
\end{Thm} 

\begin{proof}
\changeHK{This proof} is similar to the \changeHK{one} of the standard Riemannian SGD (see \cite{Bonnabel_IEEETAC_2013_s}).
Since $K$ is compact, all continuous functions on $K$ \changeHK{are} bounded. \changeHK{Furthermore}, \changeHK{because of} $\eta_t^s \rightarrow 0$, there exists $t_0$ such that $\eta_t^s A < I$ \changeHK{for $t \geq t_0$}. \changeHK{Now, we assume that} $t \geq t_0$. From \changeHS{the triangle inequality that $\|\xi_{t+1}^s\|\le A$, and hence} there exists a geodesic ${\rm Exp}(-\alpha \eta_t^s \xi_{t+1}^s)_{0 \leq \alpha \leq 1}$ linking $w_t^s$ and $w_{t+1}^s$ as ${\rm dist}(w_t^s, w_{t+1}^s) < I $, $\xi_t^s$ is defined and bounded as
\begin{eqnarray*}
\xi_t^s & = & \gradf_{i_t^s}(w_{t-1}^{s}) -  P_{\gamma}^{\scriptsize w_{t-1}^{s} \leftarrow \tilde{w}^{s-1}} \left(\gradf_{i_t^s}(\tilde{w}^{s-1}) \right) + P_{\gamma}^{\scriptsize w_{t-1}^{s} \leftarrow \tilde{w}^{s-1}} \left(\gradf(\tilde{w}^{s-1})\right) \\
& \le & A/3 + A/3 + A/3 = A.
\end{eqnarray*}

$f({\rm Exp}(-\eta_t^s \xi_{t+1}^s))=f(w_{t+1}^s)$ and thus the Taylor formula implies that 
\begin{eqnarray*}
f(w_{t+1}^s) - f(w_{t}^s) & \leq & - \eta_t^s 
\langle \xi_{t+1}^s, \gradf (w_t^s) 
\rangle + (\eta_t^s)^2 
\| \xi_{t+1}^s \|^2 k_1,
\end{eqnarray*}
where $k_1$ is an upper bound of the largest eigenvalues of the Riemannian Hessian of $f$.
\changeHK{We denote \changeHK{as} $\mathcal{F}_t^s$ an} increasing sequence of $\sigma$-algebras \changeHK{that consists of} the variables   \changeHK{until} just before time $t$ \changeHK{, i.e.,}
\begin{eqnarray*}
\mathcal{F}_t^s=\{i_1^1, \ldots, i_{m_1}^1, \ldots, i_1^{s-1}, \ldots, i_{m_{s-1}}^{s-1}, i_1^{s}, \ldots, i_{t-1}^{s}\}.
\end{eqnarray*}
Since $w_t^s$ is computed from $i_1^1\ldots, i_{t}^s$, it is measurable in $\mathcal{F}_{t+1}^s$. As $i_{t+1}^s$ is independent from $\mathcal{F}_{t+1}^s$ we have 
\begin{eqnarray*}
& & \hspace*{-1cm}\ \ \  \mathbb{E}[\langle \xi_{t+1}^s, \gradf (w_t^s)\rangle | \mathcal{F}_{t+1}^s] \\
& = & \mathbb{E}_{i_{t+1^s}}[\langle \xi_{t+1}^s, \gradf (w_t^s)\rangle] \\
& = & \mathbb{E}[\langle \gradf_{i_{t+1}^s}(w_{t}^{s}), \gradf(w_t^s) \rangle | \mathcal{F}_{t+1}^s] \\
&& - P_{\gamma}^{w_{t}^{s} \leftarrow \tilde{w}^{s-1}}(
\mathbb{E}[\langle \gradf_{i_{t+1}^s}(\tilde{w}^{s-1}), \gradf(w_t^s) \rangle | \mathcal{F}_{t+1}^s]  
-\mathbb{E}[\langle  \gradf(\tilde{w}^{s-1}), \gradf(w_t^s) \rangle | \mathcal{F}_{t+1}^s]
)\\
& = & \mathbb{E}_{i_{t+1}^s}[\langle \gradf_{i_{t+1}^s}(w_{t}^{s}), \gradf(w_t^s) \rangle] \\
&& -  P_{\gamma}^{w_{t}^{s} \leftarrow \tilde{w}^{s-1}} (
\mathbb{E}_{i_{t+1}^s}
[\langle \gradf_{i_{t+1}^s}(\tilde{w}^{s-1}), \gradf(w_{t+1}^s) \rangle]  
-\langle  \gradf(\tilde{w}^{s-1}), \gradf(w_{t+1}^s) \rangle 
)\\
& = & \mathbb{E}_{i_{t+1}^s}[\langle \gradf_{i_{t+1}^s}(w_{t}^{s}), \gradf(w_t^s) \rangle] \\
&& -  P_{\gamma}^{w_{t}^{s} \leftarrow \tilde{w}^{s-1}} (
\langle \gradf(\tilde{w}^{s-1}), \gradf(w_t^s) \rangle
-\langle  \gradf(\tilde{w}^{s-1}), \gradf(w_t^s) \rangle 
)\\
& = & \mathbb{E}_{i_{t+1}^s}[\langle \gradf_{i_{t+1}^s}(w_{t}^{s}), \gradf(w_t^s) \rangle] \\
& = & \| \gradf (w_t^s)\|^2,
\end{eqnarray*}
which yields that
\begin{eqnarray}
\label{Eq:ExpectationDiff}
\mathbb{E}[f(w_{t+1}^s) - f(w_{t}^s) | \mathcal{F}_{t+1}^s] & \leq & -\eta_t^s \| \gradf (w_t^s)\|^2 + (\eta_t^s)^2 A^2 k_1,
\end{eqnarray}
as $\| \xi_{t+1}^s \| \leq A$. As $f(w_{t}^s) \geq 0$, this proves $f(w_{t}^s) +\sum_t^{\infty} (\eta_t^s)^2 A^2 k_1$ is a nonnegative supermartingale. \changeHK{Therefore,} $f(w_{t}^s)$ converges a.s.\changeHK{.} \changeHK{In addition,} summing the inequalities \changeHK{yeilds}
\begin{eqnarray}
\label{Eq:ExpectationDiff2}
\sum_{t \geq t_0} \eta_t^s \| \gradf (w_t^s) \|^2 & \leq & - \sum_{t \geq t_0} \mathbb{E}[f(w_{t+1}^s) - f(w_{t}^s) | \mathcal{F}_t^s] + \sum_{t \geq t_0}  (\eta_t^s)^2 A^2 k_1.
\end{eqnarray}

\changeHK{Now} we \changeHK{show that} the right\changeHK{-hand side} term is bounded \changeHK{to prove} that the left\changeHK{-hand side} term converges. 

%
%


\changeHK{We see} that $f(w_t^s)$ satisfies the assumption of {\bf Proposition \ref{Prop:FiskPro}} \changeHK{from summation of} (\ref{Eq:ExpectationDiff}) over $t$. \changeHK{Therefore, it can be confirmed that}   $f(w_t^s)$ is a quasi-martingale \changeHK{that implies} $\sum_{t \geq t_0} \eta_t^s \| \gradf (w_t^s) \|^2$ converges a.s. \changeHK{from the} inequality (\ref{Eq:ExpectationDiff2}) where the \changeHK{first term in its right-hand side} can be bounded by its absolute value which \changeHK{stems from} the proposition. 
\changeHK{Here, although} $\eta_t^s \rightarrow 0$, this \changeHK{is not equivalent to that} $\| \gradf (w_t^s) \|$ converges a.s.\changeHK{. Then,} it can only converge to 0 a.s. if $\| \gradf (w_t^s) \|$ is \changeHK{guaranteed} to converge a.s..

\changeHK{Therefore, to prove that $\| \gradf (w_t^s) \|$ converges a.s., we} consider \changeHK{a} process $p_t^s = \|\gradf (w_t^s) \|^2$ \changeHK{which is clearly nonnegative}. \changeHK{From the assumption, we can bound} the second derivative \changeHK{as} $\| \gradf \|^2$ by $k_2$ along the geodesic \changeHK{from} $w_t^s$ \changeHK{towards} $w_{t+1}^s$, \changeHK{then we obtain from} a Taylor expansion \changeHSS{the relation}
\begin{eqnarray*}
p_{t+1}^s - p_t^s \leq -2 \eta_t^s \langle \gradf (w_t^s), (\nabla_{w_t^s}^2 f)  \xi_{t+1}^s \rangle + (\eta_t^s)^2 \|  \xi_{t+1}^s \|^2 k_2\changeHK{.}
\end{eqnarray*}
\changeHK{Furthermore, we bound} the Hessian of $f$ \changeHK{in} the compact set from below by $-k_3$.  \changeHK{Then, we obtain}
\begin{eqnarray*}
\mathbb{E}(p_{t+1}^s - p_t^s | \mathcal{F}_{t+1}^s) \leq 2 \eta_t^s \|  \gradf (w_t^s) \|^2 k_3 + (\eta_t^s)^2 A^2 k_2.
\end{eqnarray*}
\changeHK{Consequently, the guaranteeing that} the sum of the right term is finite \changeHK{represents equivalently that} $p_t^s$ is a quasi-martingale. \changeHK{Therefore, $p_t$ converges a.s.} towards a value\changeHSS{. This should be 0 as mentioned above}. \changeHK{This completes the proof.}
\end{proof}

\section{Local convergence rate analysis}
\label{AppenSec:LocalConvergenceAnalysis}

We state local convergence \changeHK{rate} properties of the algorithm of R-SVRG: local convergence to local minimizers and its convergence rate.

We fist assume throughout the following analysis that the functions $f_n$ are $\beta$-Lipschitz continuously differentiable below.
\begin{assumption}
\label{Appen_assump:Lipschitz}
We assume that a Riemannian manifold $(\mathcal{M}, g)$ has a positive injectivity radius. A real-valued functions $f_n : \mathcal{M} \rightarrow \mathbb{R}$ are {\it (locally) $\beta$-Lipschitz continuously differentiable} such that it is differentiable and there exists $\beta$ such that, for all $w$, $z$ in $\mathcal{M}$ with $dist(w,z) < i(\mathcal{M})$. In this case, it holds that \cite[Section 7.4.1]{Absil_OptAlgMatManifold_2008}
\begin{eqnarray}
	\label{Appen_Eq:LipschitzContinuous}
	\| P_{\alpha}^{0 \leftarrow 1} {\rm grad} f(z) - {\rm grad} f(w) \| & \leq & \beta {\rm dist}(z,w),
\end{eqnarray}
where $\alpha$ is the unique \changeHK{shortest} geodesic with $\alpha(0) = w$ and $\alpha(1)=z$, and $i(\mathcal{M})$ is the injectivity radius which \changeHK{represents} a lower bound on the size of the normal neighborhoods. $P_{\alpha}^{0 \leftarrow 1}(\cdot)$ is a transportation operator from $z$ to $w$.
\end{assumption}

Then, we derive the following lemma from the mean-value theorem.
\begin{Lem} 
\label{Lem:mean-value}
Let $f$ be a cost function on a Riemannian manifold $(\mathcal{M},g)$ and let $w^*$ be a critical point of $f$, i.e., $\gradf(w^*)=0$. Assume that there exists a convex neighborhood \changeHS{$\mathcal{U}$ of $w^* \in {\mathcal{M}}$} and a positive real number $\sigma$ such that the smallest eigenvalue of the Hessian of $f$ at each $w \in \mathcal{U}$ is not less than $\sigma$. Then, 
\begin{eqnarray*}
\label{Eq:mean-value}
f(z) \ge f(w) + \langle {\rm Exp}_{w}^{-1}( z), \gradf(w)\rangle_{w} + \frac{\sigma}{2} \|{\rm Exp}_{w}^{-1}(z)\|_{w}^2, \qquad w, z \in \mathcal{U}
\end{eqnarray*}
\end{Lem} 
\begin{proof}
Let $\xi={\rm Exp}_{w}^{-1}(z)$ for $w, z \in \mathcal{U}$.
From our assumption on $f$ and the mean value theorem, we have, \changeHS{for $\lambda \in \mathbb R$ sufficiently close to $1$,}
\begin{eqnarray*}
f({\rm Exp}_{w} \lambda \xi) 	&=& f(w) + \lambda \langle \gradf(w), \xi\rangle_{w} + \lambda^2 \int_0^1 (1-t)\langle {\rm Hess} f \left({\rm Exp}_{w} t\lambda\xi\right)[\xi], \xi\rangle_{w} dt \\
	&\ge& f(w) + \lambda \langle \gradf(w), \xi\rangle_{w} + \lambda^2 \sigma \|\xi\|_{w}^2 \int_0^1 (1-t)dt\\
	&=& f(w) + \lambda \langle \gradf(w), \xi\rangle_{w} + \frac{\sigma}{2}\lambda^2\|\xi\|_{w}^2.
\end{eqnarray*}
It follows that
\begin{eqnarray*}
	f(z) = f({\rm Exp}_{w}(\xi)) \ge f(w) + \langle \gradf(w), \xi\rangle_{w} + \frac{\sigma}{2} \| \xi \|_{w}^2.
\end{eqnarray*}
This completes the proof.
\end{proof}

Second, we show a property of the Karcher mean on a general Riemannian manifold.
\begin{Lem}
\label{AppenLem:KarcherMeanDistance}
Let $w_1,\dots,w_m$ be points on a Riemannian manifold $\mathcal{M}$ and let $w$ be the Karcher mean of the $m$ points.
For an arbitrary point $p$ on $\mathcal{M}$, we have
\begin{eqnarray*}
({\rm dist}(p,w))^2\le \frac{4}{m}\sum_{i=1}^m({\rm dist}(p,w_i))^2.
\end{eqnarray*}
\end{Lem}
\begin{proof}
From the triangle inequality and $(a + b)^2 \leq 2 a^2 + 2 b^2$ for real numbers $a, b$, we have for $i=1,2,\ldots,m$
\begin{equation*}
({\rm dist}(p,w))^2 \le \left({\rm dist}(p,w_i)+{\rm dist}(w_i,w)\right)^2 \le 2({\rm dist}(p,w_i))^2 + 2({\rm dist}(w_i,w))^2.
\end{equation*}
Since $w$ is the Karcher mean of $w_1,w_2,\ldots,w_m$, it holds that
\begin{equation*}
\sum_{i=1}^{m} ({\rm dist}(w,w_i))^2 \le \sum_{i=1}^m ({\rm dist}(p,w_i))^2.
\end{equation*}
It then follows that
\begin{equation*}
m  {\rm dist}(p,w)^2 \le 2\sum_{i=1}^m 
({\rm dist}(p,w_i))^2 + 2\sum_{i=1}^m ({\rm dist}(w_i,w))^2 \le 4 \sum_{i=1}^m ({\rm dist}(p,w_i))^2.
\end{equation*}
This completes the proof.
\end{proof}
We now derive the upper bound of the variance of $\xi_t^s$ as follows.

\begin{Lem} 
\label{AppenLem:UpperBoundVariance}
Let $\mathbb{E}_{i_t^s}[\cdot]$ be the expectation with respect to \changeHK{the distribution of} the random choice of $i_t^s$.
When each $\gradf_{n}$ is $\beta$-Lipschitz continuously differentiable, the upper bound of the variance of $\xi_t^s$ is given by
\begin{eqnarray}
\label{Append_Eq:UpperBoundVariance}
	\mathbb{E}_{i_t^s}[\| \xi_t^s \|^2] &\leq &
\beta^2 (14({\rm dist}(w_{t-1}^s,w^*))^2 + 8{\rm dist}(\tilde{w}^{s-1},w^*))^2  ).
\end{eqnarray}
\end{Lem}

\begin{proof}
\changeHK{The variance of $\xi_t^s$ in terms of the distance of $w_t^s$ and $\tilde{w}^{s-1}$ from $w^*$ is upper bounded as}

\begin{eqnarray*}
& & \hspace*{-1cm}\mathbb{E}_{i_t^s}[\| \xi_t^s \|^2]   \nonumber\\
&=&\mathbb{E}_{i_t^s}\left[\| \bigl( \gradf_{i_t^s}(w_{t-1}^{s}) - P_{\gamma}^{w_{t-1}^{s} \leftarrow w^{*}}(\gradf_{i_t^s}(w^{*})) \bigr)  \right. \nonumber\\
&& \left. + \bigl( P_{\gamma}^{w_{t-1}^{s} \leftarrow w^{*}}(\gradf_{i_t^s}(w^{*})) -  P_{\gamma}^{w_{t-1}^s \leftarrow \tilde{w}_{s-1}} \left(\gradf_{i_t^s}(\tilde{w}^{s-1}) \right) + P_{\gamma}^{w_{t-1}^s \leftarrow \tilde{w}_{s-1}} \left(\gradf(\tilde{w}^{s-1}) \right)\bigr) \|^2 \right] \nonumber\\
& \leq & 2\mathbb{E}_{i_t^s}\left[ \| \gradf_{i_t^s}(w_{t-1}^{s}) - P_{\gamma}^{w_{t-1}^{s} \leftarrow w^{*}}(\gradf_{i_t^s}(w^{*}))  \|^2 \right]  \nonumber\\
&& + 2\mathbb{E}_{i_t^s}\left[ \| P_{\gamma}^{w_{t-1}^s \leftarrow \tilde{w}_{s-1}} \left(\gradf_{i_t^s}(\tilde{w}^{s-1}) \right) - P_{\gamma}^{w_{t-1}^{s} \leftarrow w^{*}}\left(\gradf_{i_t^s}(w^{*})\right) - P_{\gamma}^{w_{t-1}^s \leftarrow \tilde{w}_{s-1}} \left(\gradf(\tilde{w}^{s-1}) \right) \|^2 \right]  \nonumber\\
& = & 2\mathbb{E}_{i_t^s}\left[ \| \gradf_{i_t^s}(w_{t-1}^{s}) - P_{\gamma}^{w_{t-1}^{s} \leftarrow w^{*}}(\gradf_{i_t^s}(w^{*}))  \|^2 \right]  \nonumber\\
&&+ 2\mathbb{E}_{i_t^s}\left[ \| P_{\gamma}^{w_{t-1}^s \leftarrow \tilde{w}_{s-1}} \left(\gradf_{i_t^s}(\tilde{w}^{s-1}) \right) - P_{\gamma}^{w_{t-1}^{s} \leftarrow w^{*}}(\gradf_{i_t^s}(w^{*})) \|^2 \right]  \nonumber\\
&& - 4 \left\langle P_{\gamma}^{w_{t-1}^s \leftarrow \tilde{w}_{s-1}} \left(\gradf(\tilde{w}^{s-1}) \right), P_{\gamma}^{w_{t-1}^s \leftarrow \tilde{w}_{s-1}} \left(\gradf(\tilde{w}^{s-1}) \right) - P_{\gamma}^{w_{t-1}^{s} \leftarrow w^{*}}(\gradf(w^{*}))\right\rangle  \nonumber\\
&&+ 2 \| P_{\gamma}^{w_{t-1}^s \leftarrow \tilde{w}_{s-1}} \left(\gradf(\tilde{w}^{s-1}) \right)\|^2 \nonumber\\
& = & 2\mathbb{E}_{i_t^s}\left[ \| \gradf_{i_t^s}(w_{t-1}^{s}) - P_{\gamma}^{w_{t-1}^{s} \leftarrow w^{*}}(\gradf_{i_t^s}(w^{*}))  \|^2 \right]  \nonumber\\
&&+ 2\mathbb{E}_{i_t^s}\left[ \| P_{\gamma}^{w_{t-1}^s \leftarrow \tilde{w}_{s-1}} \left(\gradf_{i_t^s}(\tilde{w}^{s-1}) \right) - P_{\gamma}^{w_{t-1}^{s} \leftarrow w^{*}}(\gradf_{i_t^s}(w^{*})) \|^2 \right]  \nonumber\\
&& - 2 \| P_{\gamma}^{w_{t-1}^s \leftarrow \tilde{w}_{s-1}} \left(\gradf(\tilde{w}^{s-1}) \right)\|^2  \nonumber\\
& \leq & 2\mathbb{E}_{i_t^s}\left[ \| \gradf_{i_t^s}(w_{t-1}^{s}) - P_{\gamma}^{w_{t-1}^{s} \leftarrow w^{*}}(\gradf_{i_t^s}(w^{*}))  \|^2 \right]  \nonumber\\
&&+ 2\mathbb{E}_{i_t^s}\left[ \| P_{\gamma}^{w_{t-1}^s \leftarrow \tilde{w}_{s-1}} \left(\gradf_{i_t^s}(\tilde{w}^{s-1}) \right) - P_{\gamma}^{w_{t-1}^{s} \leftarrow w^{*}}(\gradf_{i_t^s}(w^{*})) \|^2 \right]  \nonumber\\
& \le & 
2\mathbb{E}_{i_t^s}\left[ \| \gradf_{i_t^s}(w_{t-1}^{s}) - P_{\gamma}^{w_{t-1}^{s} \leftarrow w^{*}}(\gradf_{i_t^s}(w^{*}))  \|^2 \right]
 \nonumber\\
&&
+ 2\mathbb{E}_{i_t^s}\left[ \| P_{\gamma}^{w_{t-1}^s \leftarrow \tilde{w}_{s-1}} \left(\gradf_{i_t^s}(\tilde{w}^{s-1}) \right) - \gradf_{i_t^s}(w_{t-1}^{s}) + \gradf_{i_t^s}(w_{t-1}^{s}) - P_{\gamma}^{w_{t-1}^{s} \leftarrow w^{*}}(\gradf_{i_t^s}(w^{*})) \|^2 \right] \nonumber
\\
& \le & 
2\mathbb{E}_{i_t^s}\left[ \| \gradf_{i_t^s}(w_{t-1}^{s}) - P_{\gamma}^{w_{t-1}^{s} \leftarrow w^{*}}(\gradf_{i_t^s}(w^{*}))  \|^2 \right]
  \nonumber\\
&&
+ 4\mathbb{E}_{i_t^s}\left[ \| P_{\gamma}^{w_{t-1}^s \leftarrow \tilde{w}_{s-1}} \left(\gradf_{i_t^s}(\tilde{w}^{s-1}) \right) - \gradf_{i_t^s}(w_{t-1}^{s})\|^2 \right]
\nonumber\\
&&
 + 4\mathbb{E}_{i_t^s}\left[ \| \gradf_{i_t^s}(w_{t-1}^{s}) - P_{\gamma}^{w_{t-1}^{s} \leftarrow w^{*}}(\gradf_{i_t^s} (w^{*})) \|^2 \right] \nonumber
\\
& \overset{(\ref{Appen_Eq:LipschitzContinuous})}{\leq} & 
\beta^2 (6({\rm dist}(w_{t-1}^s,w^*))^2 + 4({\rm dist}(\tilde{w}^{s-1},w_{t-1}^s))^2  )
 \nonumber\\
& \le & 
\beta^2 (6({\rm dist}(w_{t-1}^s,w^*))^2 + 4({\rm dist}(\tilde{w}^{s-1},w^*)+{\rm dist}(w^*,w_{t-1}^s))^2  )
 \nonumber\\
& \le & 
\beta^2 (6({\rm dist}(w_{t-1}^s,w^*))^2 + 8({\rm dist}(\tilde{w}^{s-1},w^*))^2+8({\rm dist}(w^*,w_{t-1}^s))^2  )
 \nonumber\\
& = & 
\beta^2 (14({\rm dist}(w_{t-1}^s,w^*))^2 + 8({\rm dist}(\tilde{w}^{s-1},w^*))^2  ),
\end{eqnarray*}
where the first, fourth and seventh inequalities follow from $(a + b)^2 \leq 2 a^2 + 2 b^2$ for real numbers $a, b$, and the sixth inequality uses the triangle inequality. The third equality comes from 
$\mathbb{E}_{i_t^s}[\gradf_{i_t^s}(\tilde{w}^{s-1})]=\gradf(\tilde{w}^{s-1})$, and the fourth equality from $\gradf(w^*)=0$.
\end{proof}

Now we introduce {\bf Lemma 6} in \cite{Zhang_JMLR_2016_s} to evaluate the distance between $x_{t}^s$ and  $x^{*}$ using the smoothness of our objective function.
\begin{Lem}[{\bf Lemma 6} in \cite{Zhang_JMLR_2016_s}]
\label{LemZhang}
If $a$, $b$, $c$ are the sides (i.e., side lengths) of a geodesic triangle in an Alexandrov space with curvature lower bounded by $\kappa$, and $A$ is the angle between sides $b$ and $c$, then
\begin{equation*}
a^2 \le \frac{\sqrt{|\kappa|}c}{\tanh(\sqrt{|\kappa|}c)}b^2 + c^2 - 2bc\cos(A).
\end{equation*}
\end{Lem}
Note that all the theorems and lemmas above hold for the Grassmann manifold. In the last theorem, we consider the Grassmann manifold specifically.

\begin{Thm}
\label{Append_Thm:LocalConvergence}
Let $\mathcal{M}$ be the Grassmann manifold and $\mat{U}^* \in \mathcal{M}$ be a non-degenerate local minimizer of $f$ (i.e., ${\rm grad} f(\mat{U}^*)=0$ and \changeHS{the Hessian ${\rm Hess}f(\mat{U}^*)$ of $f$ at $\mat{U}^*$} is positive definite) \changeHS{and suppose that the assumption in Lemma \ref{Lem:mean-value} holds}.
When each $\gradf_{n}$ is $\beta$-Lipschitz continuously differentiable \changeHS{and $\eta > 0$ is sufficiently small such that $0 < \eta(\sigma - 14 \eta \beta^2) < 1$}, 
it then follows that for any sequence $\{\tilde{\mat{U}}^s\}$ generated by the algorithm converging to $\mat{U}^*$, there exists $K>0$ such that for all $s>K$,
\begin{eqnarray*}
\mathbb{E}[({\rm dist}(\tilde{\mat{U}}^s,\mat{U}^*))^2]& \leq & 
\frac{4(1+ 8 m \eta^2  \beta^2)}{ \eta m (\sigma - 14 \eta \beta^2)} 
\mathbb{E}[({\rm dist}(\tilde{\mat{U}}^{s-1},\mat{U}^*))^2].
\end{eqnarray*}
\end{Thm}

\begin{proof}
The Grassmann manifold is geodesically complete \cite{Absil_OptAlgMatManifold_2008} and the sectional curvature of the Grassmann manifold is bounded below by 0 \cite{tron2013riemannian_s}.
Every complete Riemannian manifold whose sectional curvature is bounded below is an Alexandrov space \cite{shiohama1993introduction_s}.
Therefore, the Grassmann manifold satisfies the assumptions in {\bf Lemma \ref{LemZhang}} with $\kappa = 0$.
Then, \changeHS{conditioned on $\mat{U}_{t-1}^s$,} the expectation of the distance between $\mat{U}_{t}^s$ and $\mat{U}^{*}$ with respect to the random choice of $i_t^s$ is evaluated as
\begin{eqnarray*}
&& \mathbb{E}_{i_t^s}\left[({\rm dist}(\mat{U}_{t}^s, \mat{U}^{*}))^2\right] \\
& \le & \mathbb{E}_{i_t^s}\left[({\rm dist}(\mat{U}_{t-1}^s, \mat{U}_t^s))^2+({\rm dist}(\mat{U}_{t-1}^s, \mat{U}^*))^2 - 2 \langle{\rm Exp}_{\scriptsize\mat{U}_{t-1}^s}^{-1}(\mat{U}_{t}^s), {\rm Exp}_{\scriptsize\mat{U}_{t-1}^s}^{-1}(\mat{U}^*)\rangle_{\scriptsize\mat{U}_{t-1}^s}\right].
\end{eqnarray*}
It follows that
\begin{eqnarray*}
&& \mathbb{E}_{i_t^s}\left[({\rm dist}(\mat{U}_{t}^s, \mat{U}^{*}))^2 - ({\rm dist}(\mat{U}_{t-1}^s, \mat{U}^*))^2\right]\\
& \le & \mathbb{E}_{i_t^s}[({\rm dist}(\mat{U}_{t-1}^s, \mat{U}_{t}^s))^2 - 2 \langle -\eta \xi_t^s, {\rm Exp}_{\scriptsize\mat{U}_{t-1}^s}^{-1}(\mat{U}^*)\rangle_{\scriptsize\mat{U}_{t-1}^s}] \\
& = & \mathbb{E}_{i_t^s}[({\rm dist}(\mat{U}_{t-1}^s, \mat{U}_{t}^s))^2 + 2\eta\langle\gradf(\mat{U}_{t-1}^s), {\rm Exp}_{\scriptsize\mat{U}_{t-1}^s}^{-1}(\mat{U}^*)\rangle_{\scriptsize\mat{U}_{t-1}^s}],
\end{eqnarray*}
where the last equality follows
\begin{eqnarray*}
\mathbb{E}_{i_t^s}[\xi_t^s] & = & 
\mathbb{E}_{i_t^s}[\gradf_{i_t^s}(\mat{U}_{t-1}^{s})] - 
P_{\gamma}^{\scriptsize \mat{U}_{t-1}^{s} \leftarrow \tilde{\mat{U}}^{s-1}} 
\left(
\mathbb{E}_{i_t^s}[\gradf_{i_t^s}(\tilde{\mat{U}}^{s-1}) ] -
\gradf(\tilde{\mat{U}}^{s-1})
\right) \nonumber\\
& = & \gradf(\mat{U}_{t-1}^{s}) -
P_{\gamma}^{\scriptsize \mat{U}_{t-1}^{s} \leftarrow \tilde{\mat{U}}^{s-1}} 
\left(
\gradf(\tilde{\mat{U}}^{s-1})  -
\gradf(\tilde{\mat{U}}^{s-1})
\right)  \nonumber\\
& = & \gradf(\mat{U}_{t-1}^{s}).
\end{eqnarray*}

{\bf Lemma \ref{Lem:mean-value}} together with the relation $f(\mat{U}^*) \le f(\mat{U}_{t-1}^s)$ yields that
\begin{eqnarray*}
\langle \gradf(\mat{U}_{t-1}^s), {\rm Exp}_{\scriptsize\mat{U}_{t-1}^s}^{-1}(\mat{U}^*)\rangle_{\scriptsize\mat{U}_{t-1}^s} \le -\frac{\sigma}{2} \|{\rm Exp}_{\scriptsize\mat{U}_{t-1}^s}^{-1}(\mat{U}^*)\|_{\scriptsize\mat{U}_{t-1}^s}^2 = -\frac{\sigma}{2} ({\rm dist}(\mat{U}_{t-1}^s, \mat{U}^*))^2,
\end{eqnarray*}
with the assumption that $K$ is sufficient large.
We thus obtain by {\bf Lemma \ref{AppenLem:UpperBoundVariance}}
\begin{eqnarray*}
&& \mathbb{E}\left[({\rm dist}(\mat{U}_{t}^s, \mat{U}^{*}))^2 - ({\rm dist}(\mat{U}_{t-1}^s, \mat{U}^*))^2\right]\\
& \le & \mathbb{E}[\|\eta\xi_t^s\|^2  - \sigma\eta({\rm dist}(\mat{U}_{t-1}^s, \mat{U}^*))^2]\\
& \overset{(\ref{Append_Eq:UpperBoundVariance})}{\le} & \eta^2\beta^2\mathbb{E}[14({\rm dist}(\mat{U}_{t-1}^s, \mat{U}^*))^2 + 8({\rm dist}(\tilde{\mat{U}}^{s-1}, \mat{U}^*))^2 - \sigma\eta({\rm dist}(\mat{U}_{t-1}^s, \mat{U}^*))^2]\\
& = & \eta(14\eta\beta^2-\sigma)\mathbb{E}[({\rm dist}(\mat{U}_{t-1}^s, \mat{U}^*))^2 + 8\eta^2\beta^2({\rm dist}(\tilde{\mat{U}}^{s-1}, \mat{U}^*))^2].
\end{eqnarray*}
It follows that
\begin{eqnarray*}
&& \mathbb{E}_{i_t^s}\left[({\rm dist}(\mat{U}_{t}^s, \mat{U}^{*}))^2 - ({\rm dist}(\mat{U}_{t-1}^s, \mat{U}^*))^2\right]\\
& \le & \eta(14\eta\beta^2-\sigma)\mathbb{E}_{i_{t}^s}\left[({\rm dist}(\mat{U}_{t-1}^s, \mat{U}^*))^2 + 8\eta^2\beta^2({\rm dist}(\tilde{\mat{U}}^{s-1}, \mat{U}^*))^2\right].
\end{eqnarray*}
Summing over $t=1, \ldots, m$ of the inner loop on $s$-th epoch, we have 
\begin{equation*}
\begin{split}
&\mathbb{E}[({\rm dist}(\mat{U}_{m}^s, \mat{U}^{*}))^2 - ({\rm dist}(\mat{U}_{0}^s,\mat{U}^{*}))^2] \\
&\hspace*{1cm}\leq \eta (14\eta \beta^2 - \sigma) 
\sum_{t=1}^m \mathbb{E}[({\rm dist}(\mat{U}_{t-1}^s,\mat{U}^*))^2]
 + 8 m \eta^2  \beta^2 ({\rm dist}(\tilde{\mat{U}}^{s-1},\mat{U}^*))^2.
\end{split}
\end{equation*}
Rearranging and using $\mat{U}_0^s = \tilde{\mat{U}}^{s-1}$, we obtain 
\begin{eqnarray*}
&& \eta (\sigma - 14 \eta \beta^2) \sum_{t=1}^{m} \mathbb{E}[({\rm dist}(\mat{U}_{t}^s,\mat{U}^*))^2] \\
& = & \changeHS{\eta (\sigma - 14 \eta \beta^2) \mathbb{E}\left[\sum_{t=0}^{m-1} ({\rm dist}(\mat{U}_{t}^s,\mat{U}^*))^2 + ({\rm dist}(\mat{U}_{m}^s,\mat{U}^*))^2 - ({\rm dist}(\mat{U}_{0}^s,\mat{U}^*))^2 \right]}\\
& \leq & \mathbb{E}\left[\changeHS{({\rm dist}(\mat{U}_{0}^s,\mat{U}^*))^2 - ({\rm dist}(\mat{U}_{m}^s,\mat{U}^*))^2 + 8 m \eta^2 \beta^2 ({\rm dist}(\mat{U}_{0}^s,\mat{U}^*))^2  \right. \\
& & \left. -\eta(\sigma - 14 \eta \beta^2) ( ({\rm dist}(\mat{U}_{0}^s,\mat{U}^*))^2 - ({\rm dist}(\mat{U}_{m}^s,\mat{U}^*))^2)}\right]\\
& \leq & \changeHS{(1 - \eta (\sigma - 14 \eta \beta^2) + 8 m \eta^2 \beta^2)\mathbb{E}[({\rm dist}(\mat{U}_{m}^s,\mat{U}^*))^2]}\\
& \leq & (1+ 8 m \eta^2  \beta^2) \mathbb{E}[({\rm dist}(\tilde{\mat{U}}^{s-1},\mat{U}^*))^2].
\end{eqnarray*}
Using $\tilde{\mat{U}}^{s} = g_m(\mat{U}_1^s, \ldots,\mat{U}_m^s)$ and {\bf Lemma \ref{AppenLem:KarcherMeanDistance}}, we obtain
\begin{eqnarray*}
\mathbb{E}[({\rm dist}(\tilde{\mat{U}}^s,\mat{U}^*))^2]& \leq & 
\frac{4(1+ 8 m \eta^2  \beta^2)}{\eta m(\sigma - 14\eta \beta^2)} 
\mathbb{E}[({\rm dist}(\tilde{\mat{U}}^{s-1},\mat{U}^*))^2].
\end{eqnarray*}
\end{proof}
\changeHS{In the above theorem, we note that, from the definitions of $\beta$ and $\sigma$, $\beta$ can be chosen arbitrarily large and $\sigma$ arbitrarily small.
Therefore, $\eta = \sigma / 28\beta^2$, for example, satisfies $0 < \eta(\sigma - 14 \eta \beta^2) < 1 $ for sufficiently large $\beta$ and small $\sigma$.}


%
%
\section{Additional numerical comparison}
\label{AppenSec:AdditionalNumericalComparison}

In addition to the representative numerical comparisons in the paper, we show additional numerical experiments.

\begin{figure}[htbp]
\begin{center}
	\begin{minipage}[t]{0.32\textwidth}
	\begin{center}
		\includegraphics[width=\textwidth]{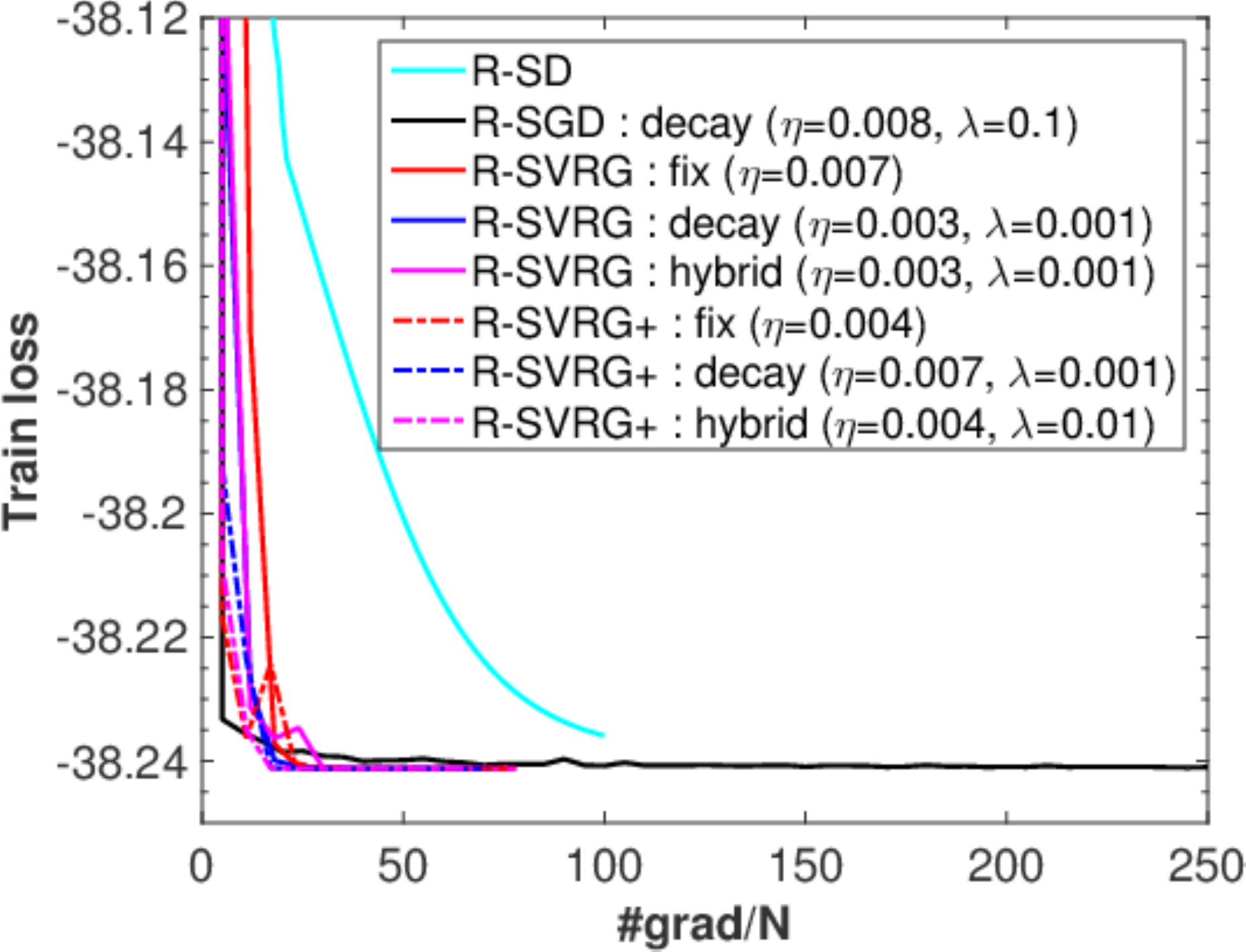}\\
		
		{\small (a-1) Train loss (enlarged).}
		
	\end{center} 
	\end{minipage}
	\hspace*{-0.1cm}
	\begin{minipage}[t]{0.32\textwidth}
	\begin{center}
		\includegraphics[width=\textwidth]{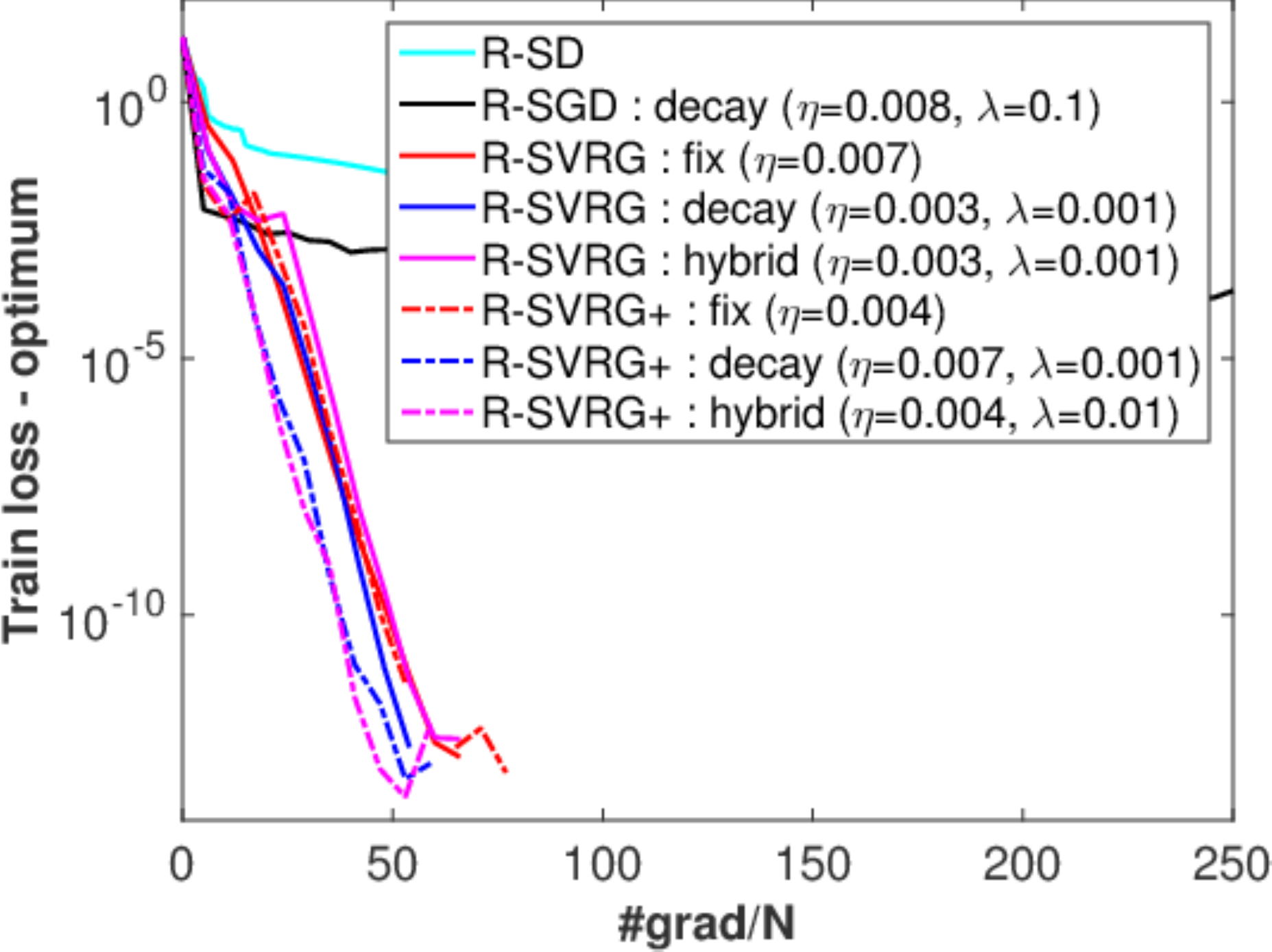}\\
		
		{\small (a-2) Optimality gap. }
		
	\end{center} 
	\end{minipage}
	\hspace*{-0.1cm}
	\begin{minipage}[t]{0.32\textwidth}
	\begin{center}
		\includegraphics[width=\textwidth]{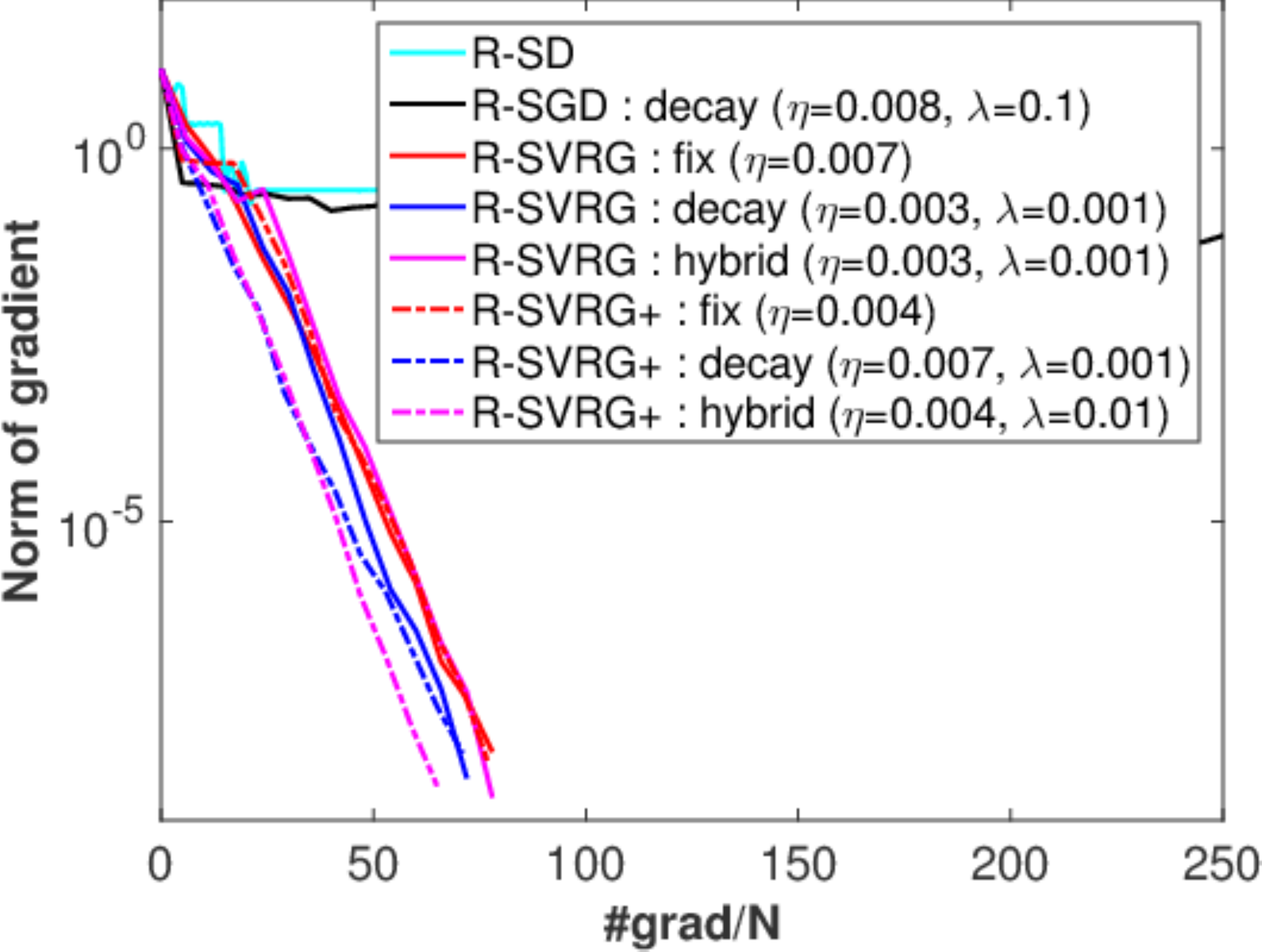}\\
		
		{\small (a-3) Norm of gradient.}
		
	\end{center} 
	\end{minipage}
\vspace*{0.4cm}
	
(a) $N=10000, d=20, r=10$.
\vspace*{0.5cm}

	\begin{minipage}[t]{0.32\textwidth}
	\begin{center}
		\includegraphics[width=\textwidth]{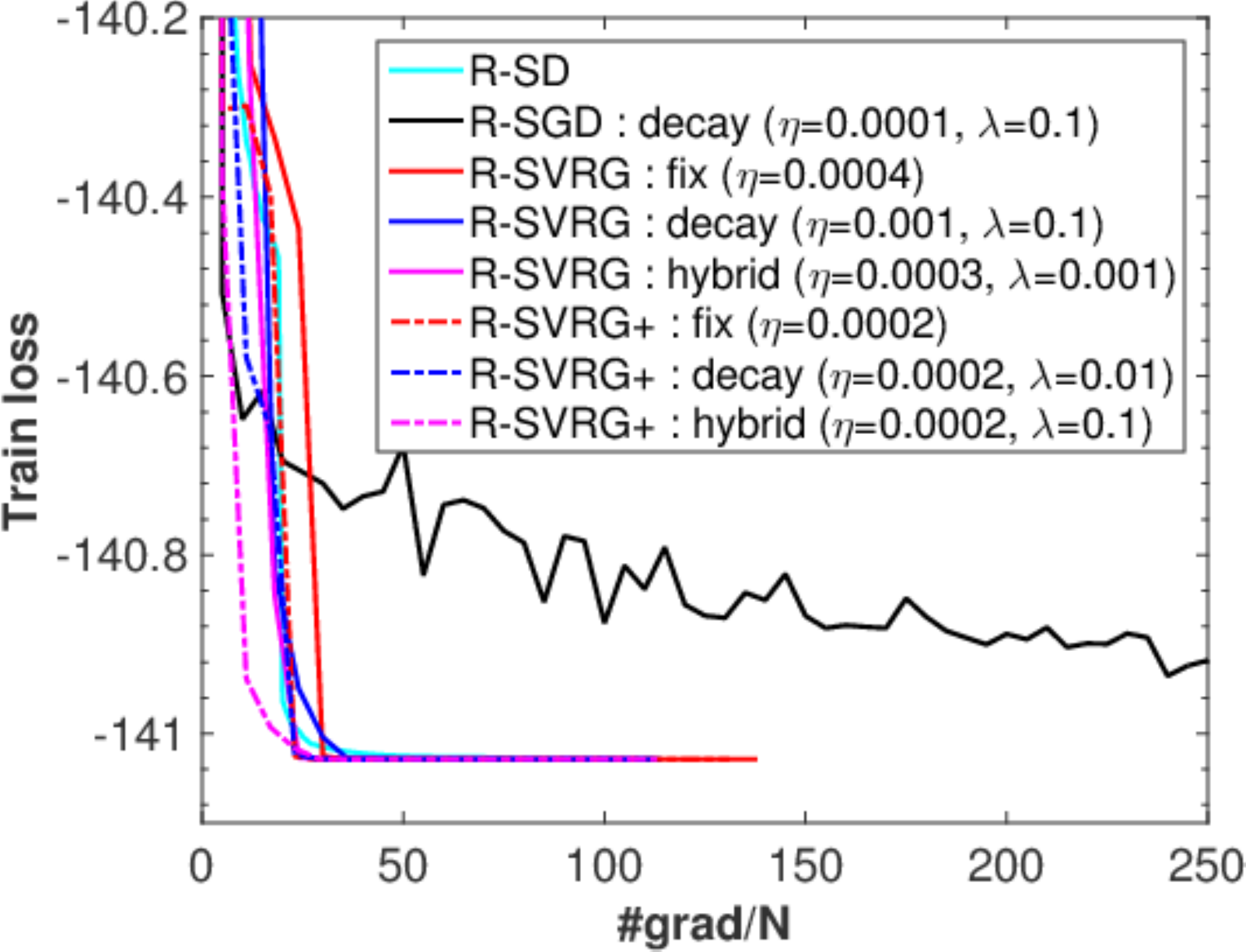}\\
		
		{\small (b-1) Train loss (enlarged).}
		
	\end{center} 
	\end{minipage}
	\hspace*{-0.1cm}
	\begin{minipage}[t]{0.32\textwidth}
	\begin{center}
		\includegraphics[width=\textwidth]{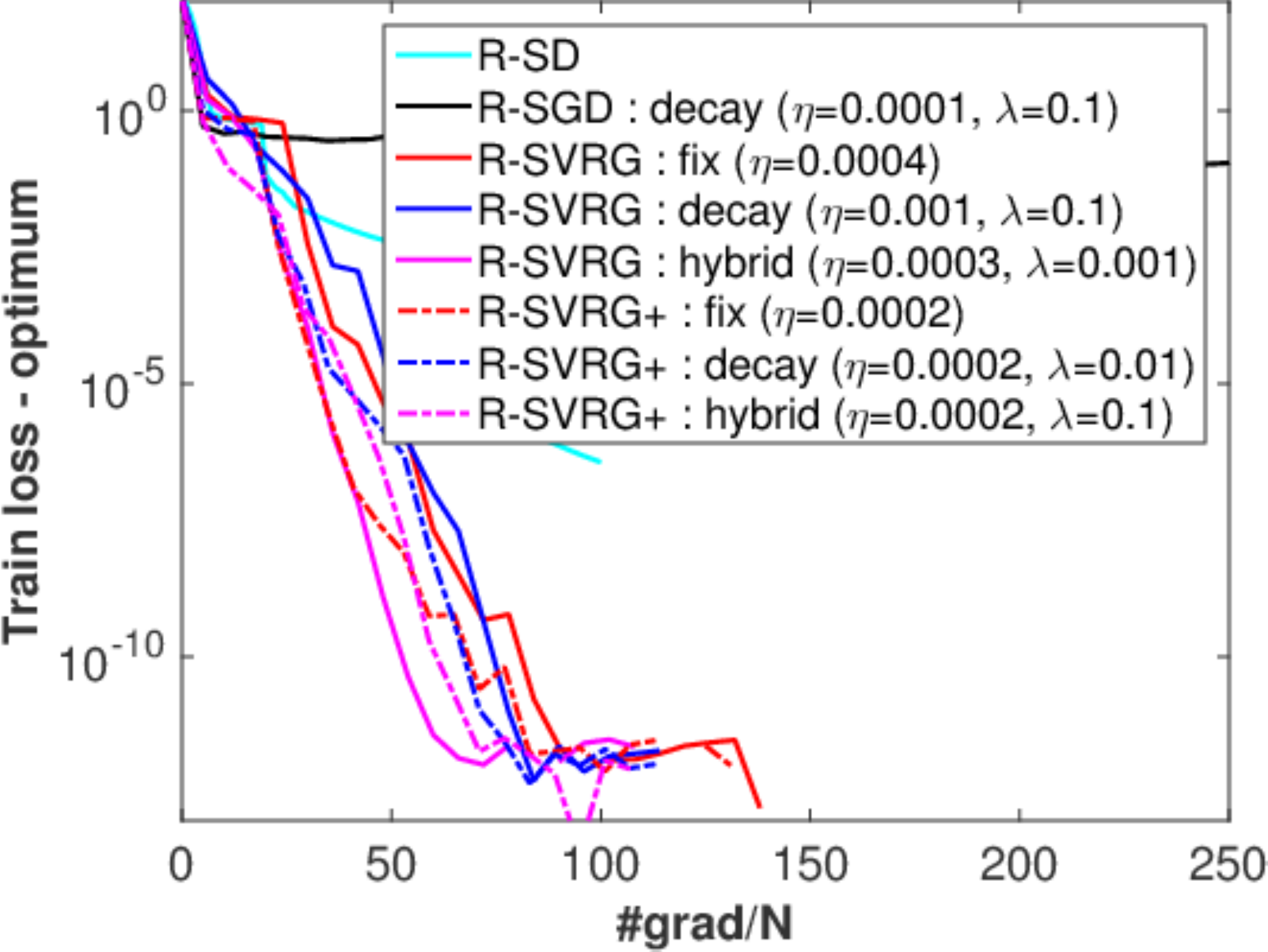}\\
		
		{\small (b-2) Optimality gap. }
		
	\end{center} 
	\end{minipage}
	\hspace*{-0.1cm}
	\begin{minipage}[t]{0.32\textwidth}
	\begin{center}
		\includegraphics[width=\textwidth]{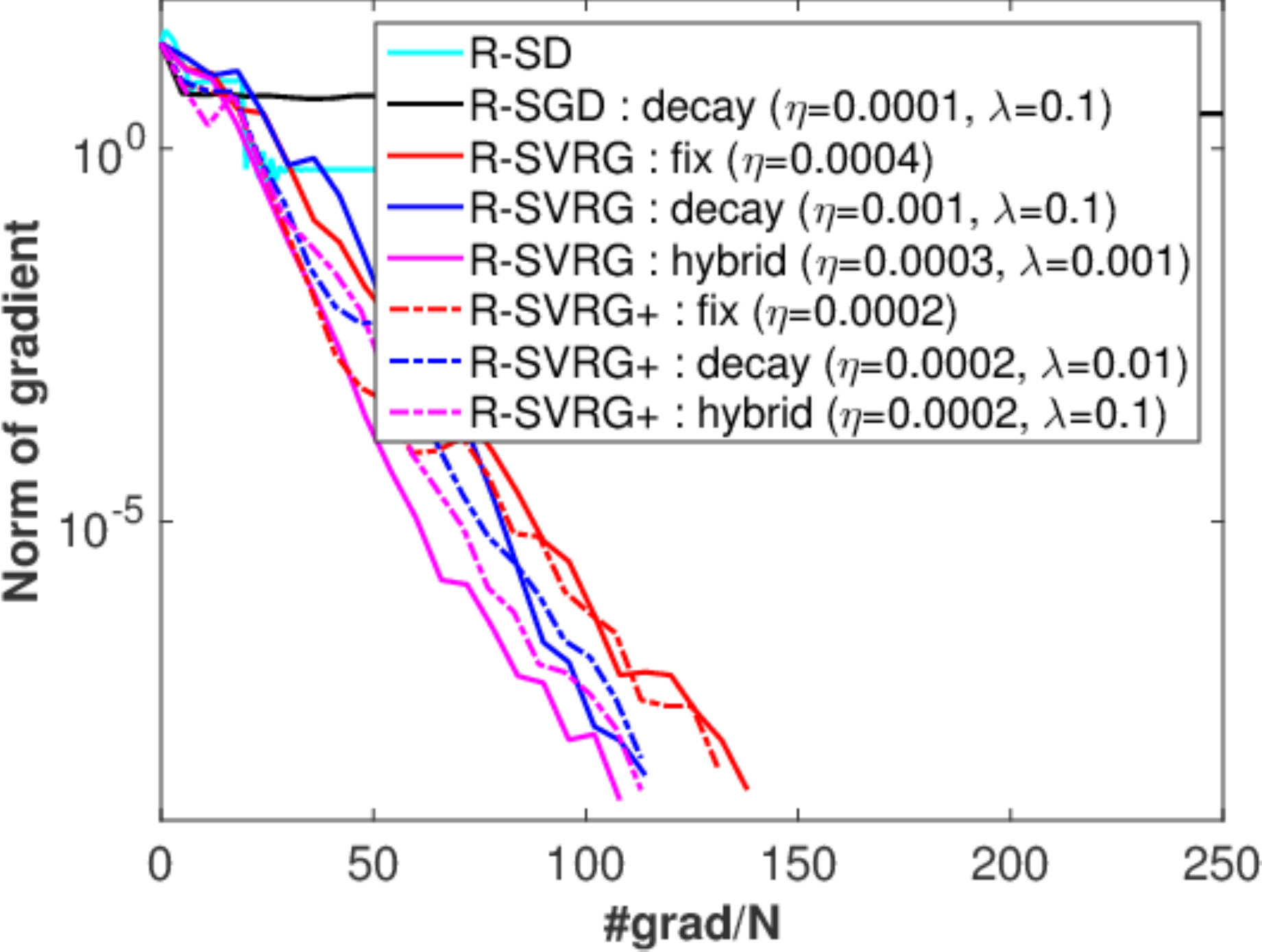}\\
		
		{\small (b-3) Norm of gradient.}
		
	\end{center} 
	\end{minipage}
\vspace*{0.5cm}
	
(b) $N=10000, d=100, r=5$.
\vspace*{0.5cm}

	\begin{minipage}[t]{0.32\textwidth}
	\begin{center}
		\includegraphics[width=\textwidth]{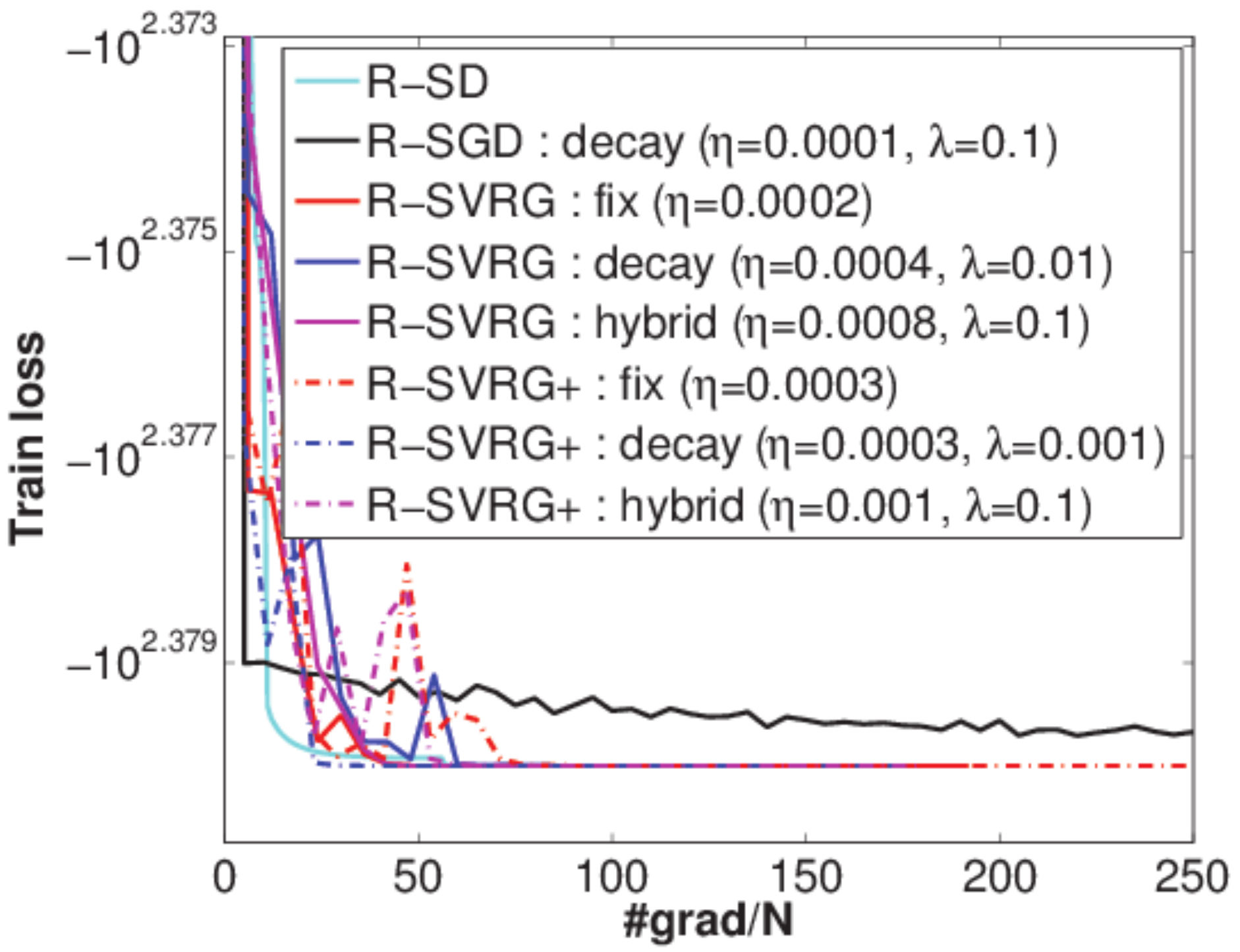}\\
		
		{\small (c-1) Train loss (enlarged).}
		
	\end{center} 
	\end{minipage}
	\hspace*{-0.1cm}
	\begin{minipage}[t]{0.32\textwidth}
	\begin{center}
		\includegraphics[width=\textwidth]{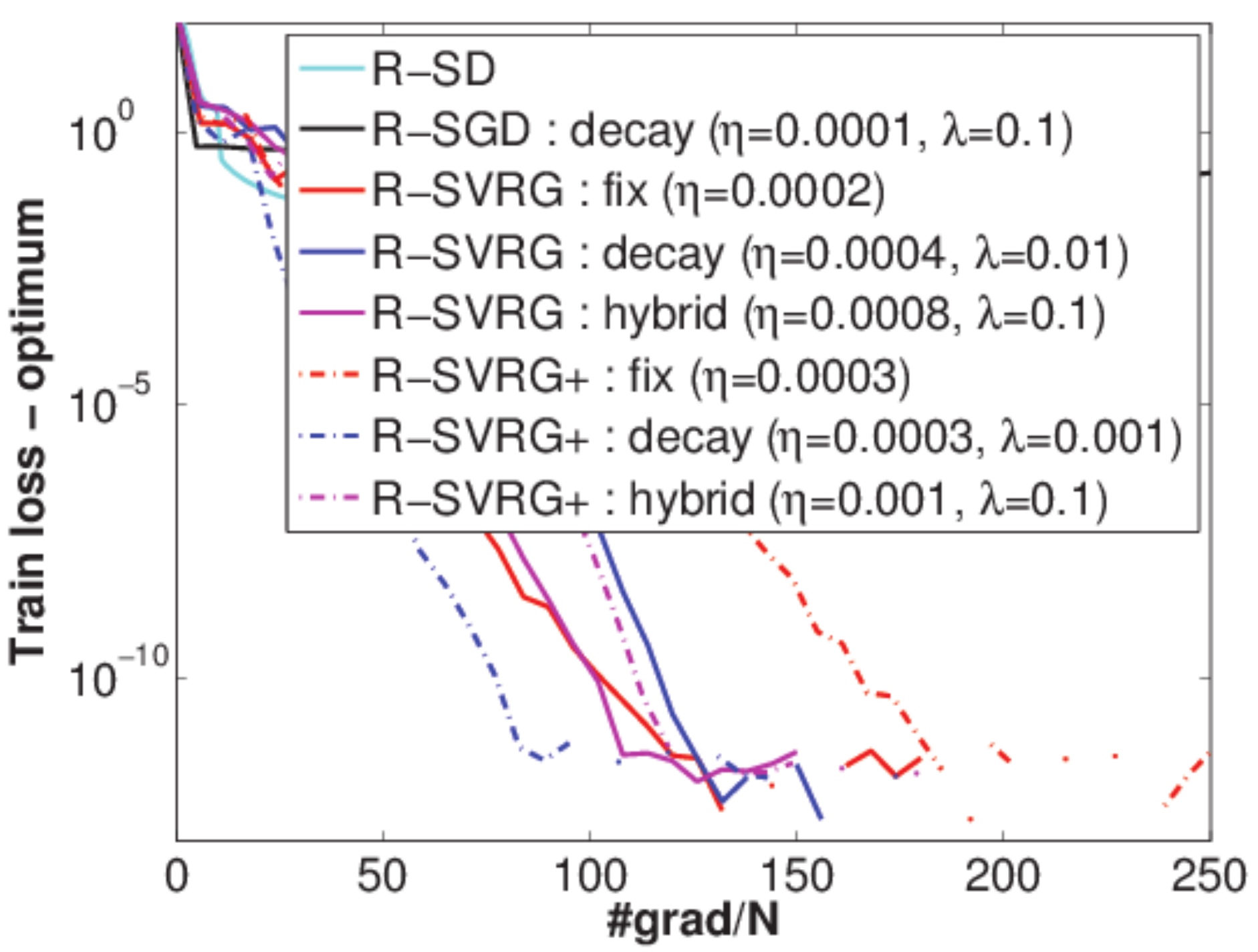}\\
		
		{\small (c-2) Optimality gap. }
		
	\end{center} 
	\end{minipage}
	\hspace*{-0.1cm}
	\begin{minipage}[t]{0.32\textwidth}
	\begin{center}
		\includegraphics[width=\textwidth]{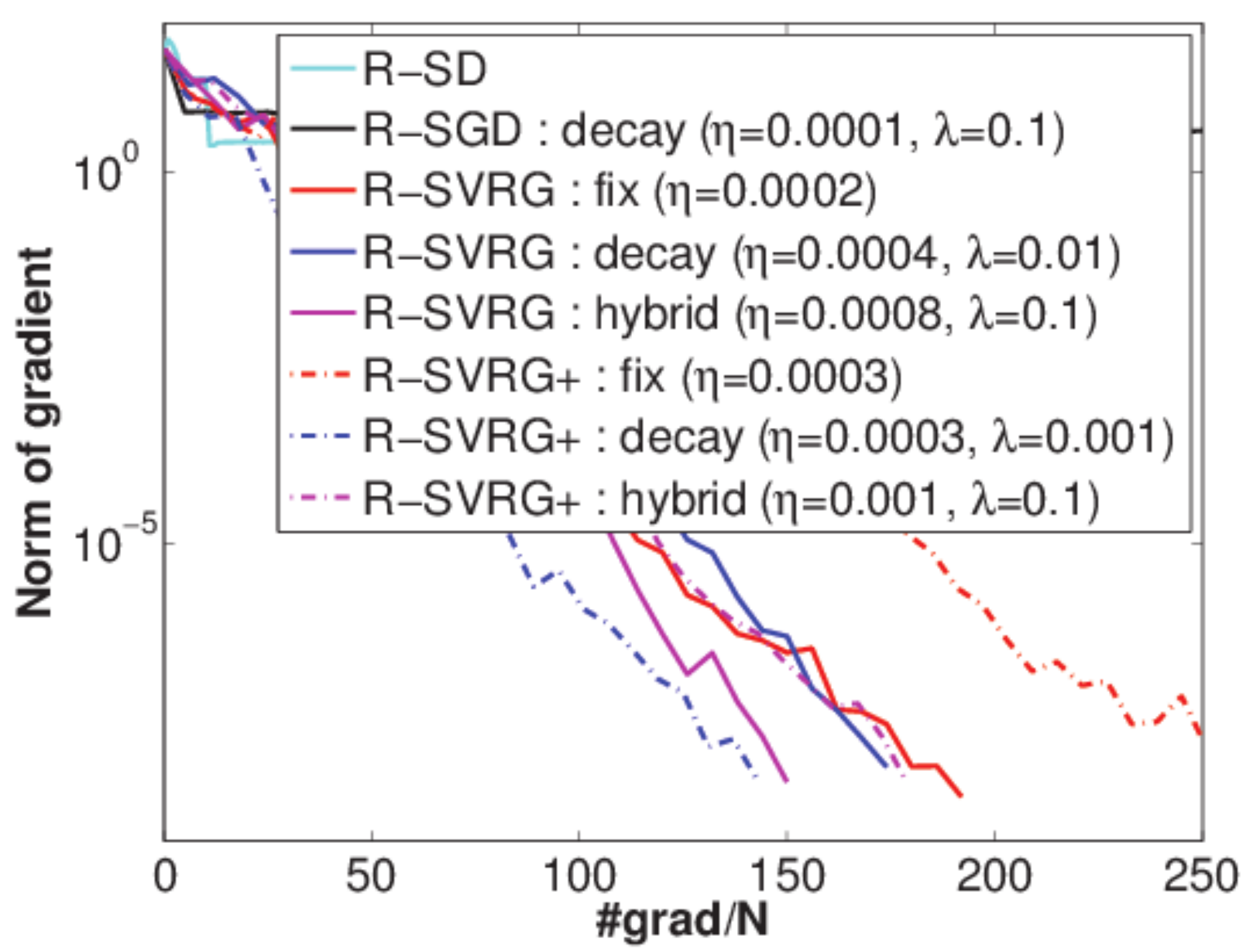}\\
		
		{\small (c-3) Norm of gradient.}
		
	\end{center} 
	\end{minipage}
\vspace*{0.5cm}
	
(c) $N=10000, d=100, r=10$.
	
\caption{The PCA problem. }
\label{Appen_fig:PCA_results}
\end{center}		
\end{figure}

{\bf PCA problem (additional experiments).} We consider the PCA problem of $N=10000$, $d=20$, and $r=10$. Whereas the manuscript provides the results for the case of $r=5$, here we show the results for the case of $r=10$. Figure \ref{Appen_fig:PCA_results}(a) shows the train loss, optimality gap, and the norm of gradient. These results indicate the superior performances of R-SVRG and R-SVRG+. In addition, we consider a larger-scale instance with $d=100$ and $d=20$. The results are shown in Figures \ref{Appen_fig:PCA_results}(b) and \ref{Appen_fig:PCA_results}(c) for two different ranks $r=5$ and $r=10$, respectively. Overall, we find the superior performances of R-SVRG and R-SVRG+.

\begin{figure}[t]
\begin{center}
	\begin{minipage}[t]{0.32\textwidth}
	\begin{center}
		\includegraphics[width=\textwidth]{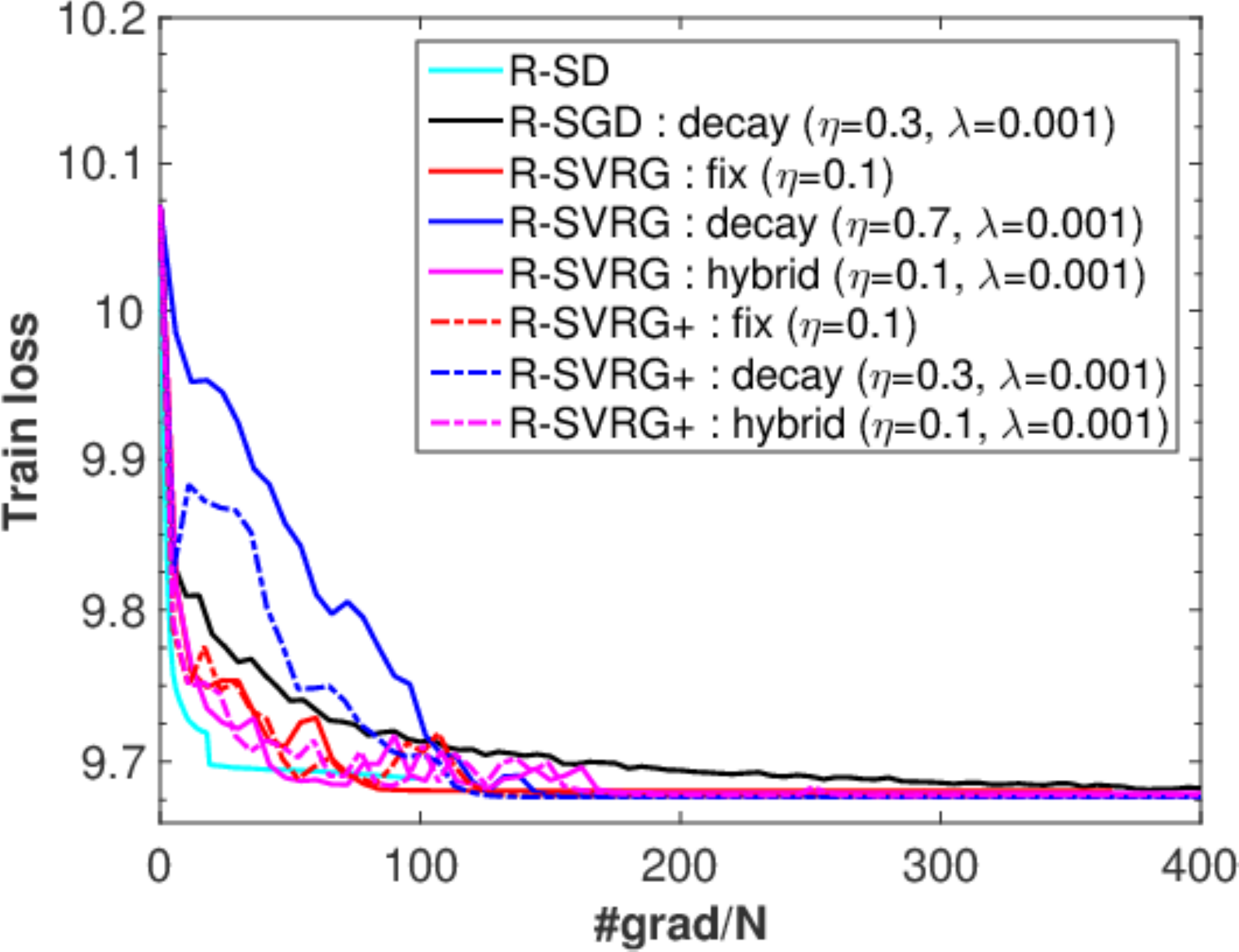}\\
		
		{\small (a-1) Train loss.}
		
	\end{center} 
	\end{minipage}
	\hspace*{-0.1cm}
	\begin{minipage}[t]{0.32\textwidth}
	\begin{center}
		\includegraphics[width=\textwidth]{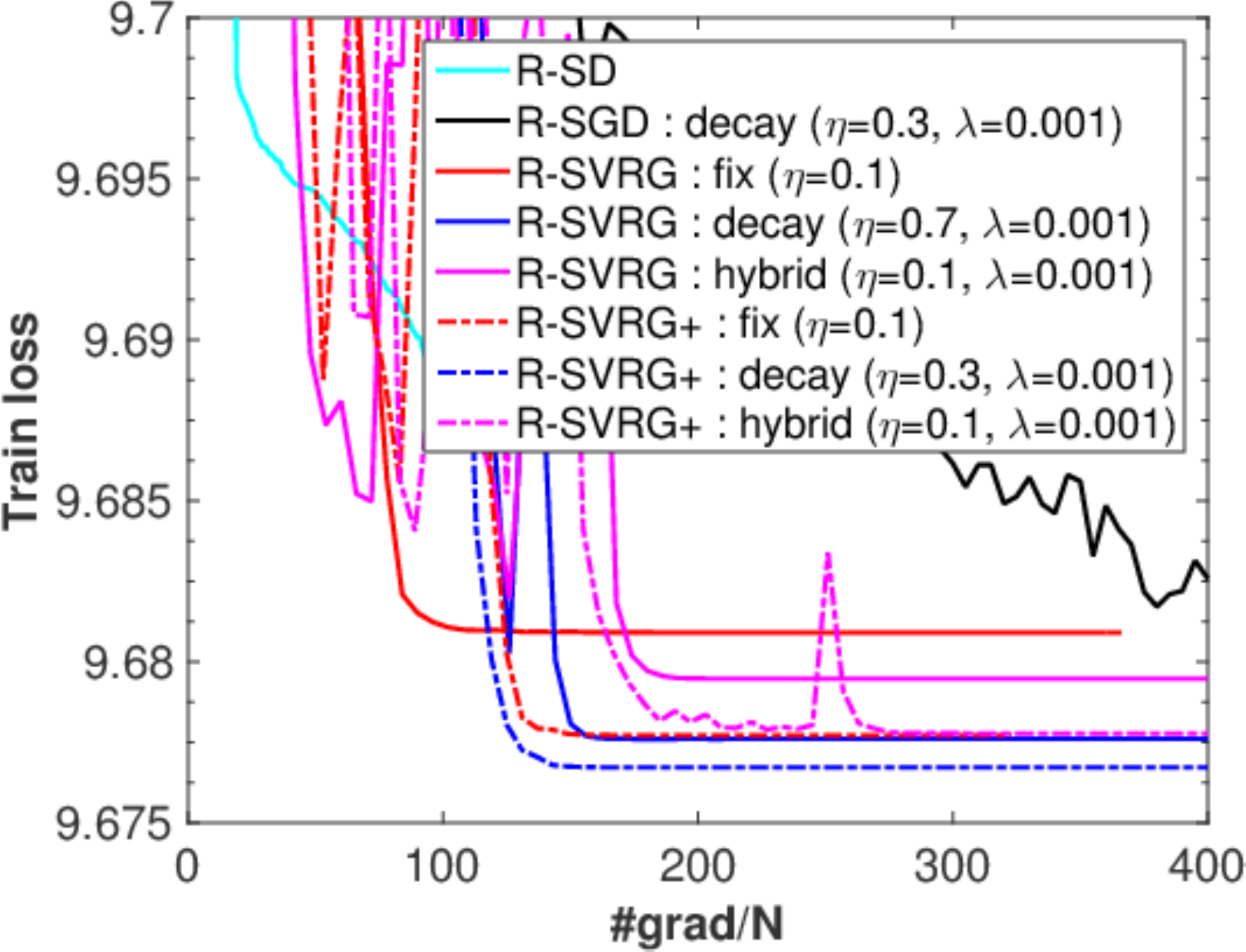}\\
		
		{\small (a-2) Train loss (enlarged). }
		
	\end{center} 
	\end{minipage}
	\hspace*{-0.1cm}
	\begin{minipage}[t]{0.32\textwidth}
	\begin{center}
		\includegraphics[width=\textwidth]{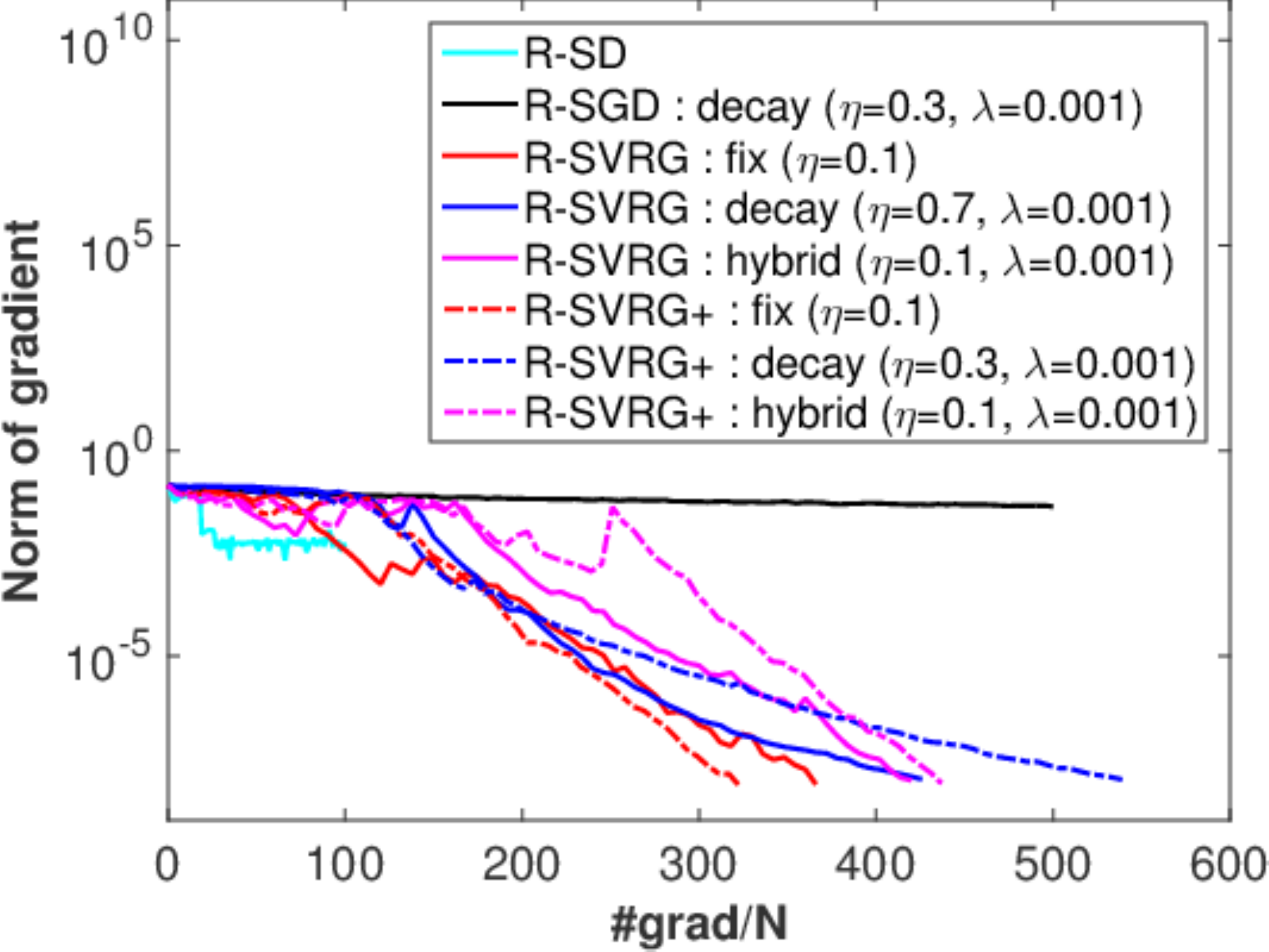}\\
		
		{\small (a-3) Norm of gradient.}
		
	\end{center} 
	\end{minipage}
\vspace*{0.5cm}
	
(a) $N=1000, d=300, r=10$.
\vspace*{0.5cm}

	\begin{minipage}[t]{0.32\textwidth}
	\begin{center}
		\includegraphics[width=\textwidth]{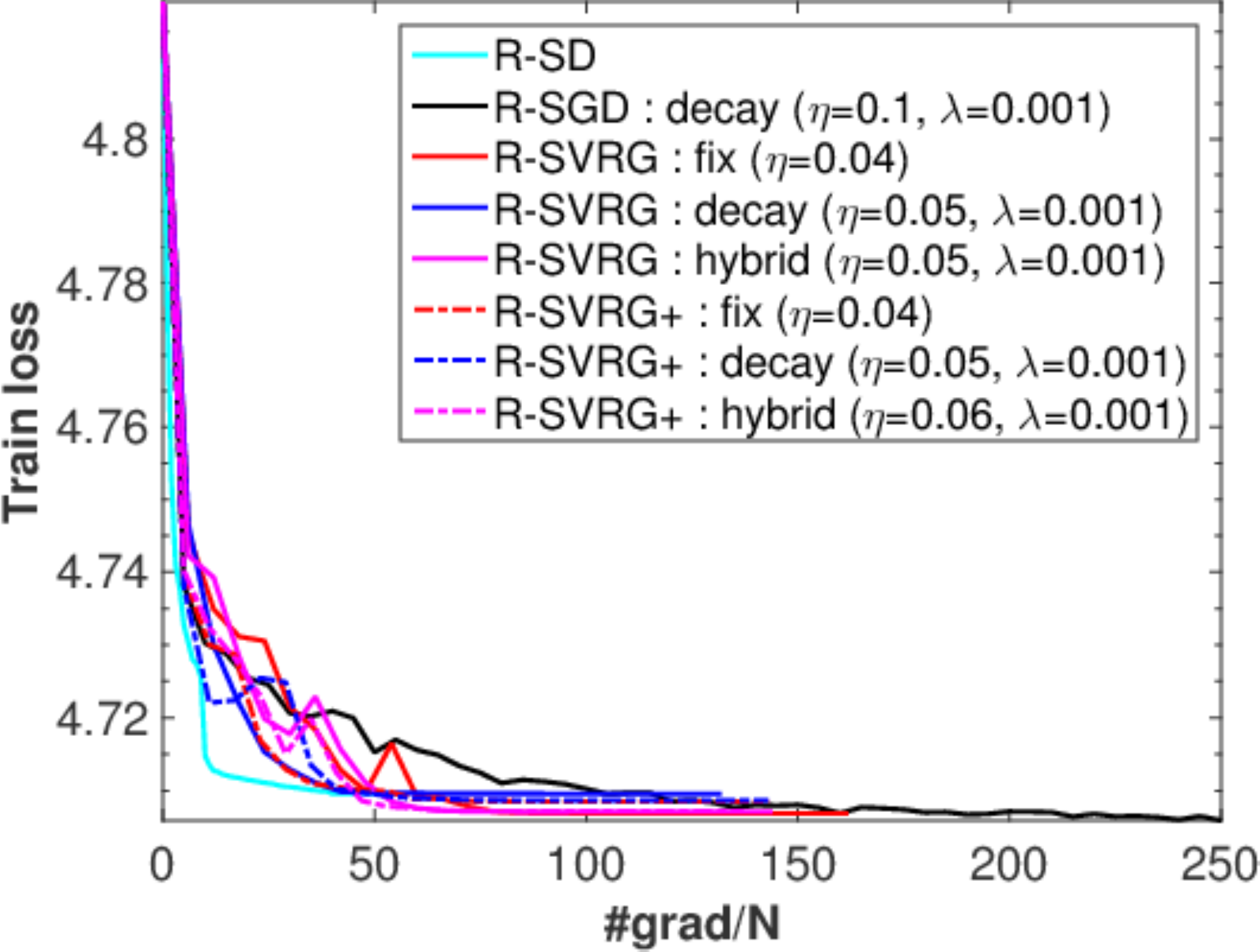}\\
		
		{\small (b-1) Train loss.}
		
	\end{center} 
	\end{minipage}
	\hspace*{-0.1cm}
	\begin{minipage}[t]{0.32\textwidth}
	\begin{center}
		\includegraphics[width=\textwidth]{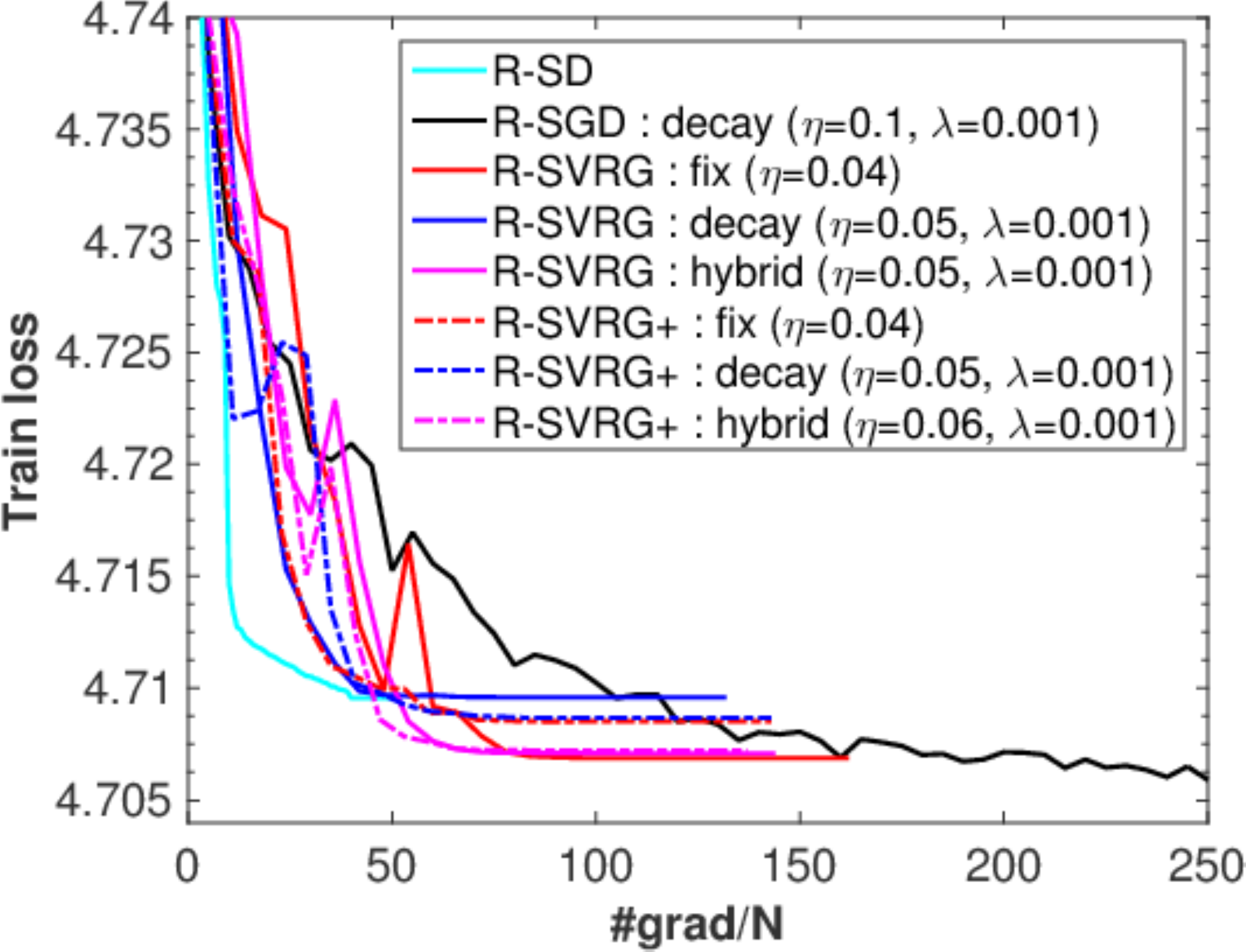}\\
		
		{\small (b-2) Train loss (enlarged). }
		
	\end{center} 
	\end{minipage}
	\hspace*{-0.1cm}
	\begin{minipage}[t]{0.32\textwidth}
	\begin{center}
		\includegraphics[width=\textwidth]{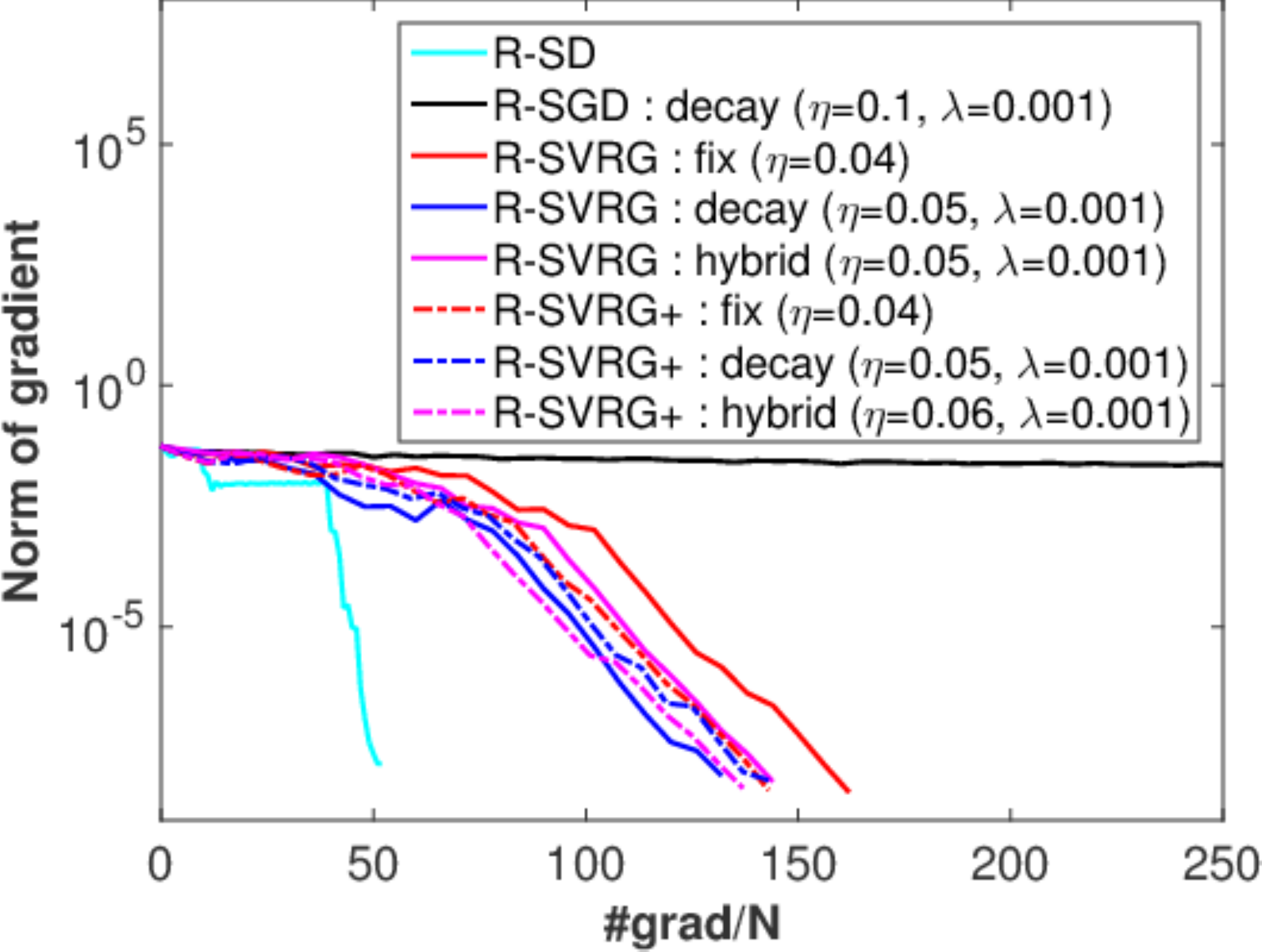}\\
		
		{\small (b-3) Norm of gradient.}
		
	\end{center} 
	\end{minipage}
\vspace*{0.5cm}
	
(b) $N=3000, d=100, r=5$.
\vspace*{0.5cm}

	\begin{minipage}[t]{0.32\textwidth}
	\begin{center}
		\includegraphics[width=\textwidth]{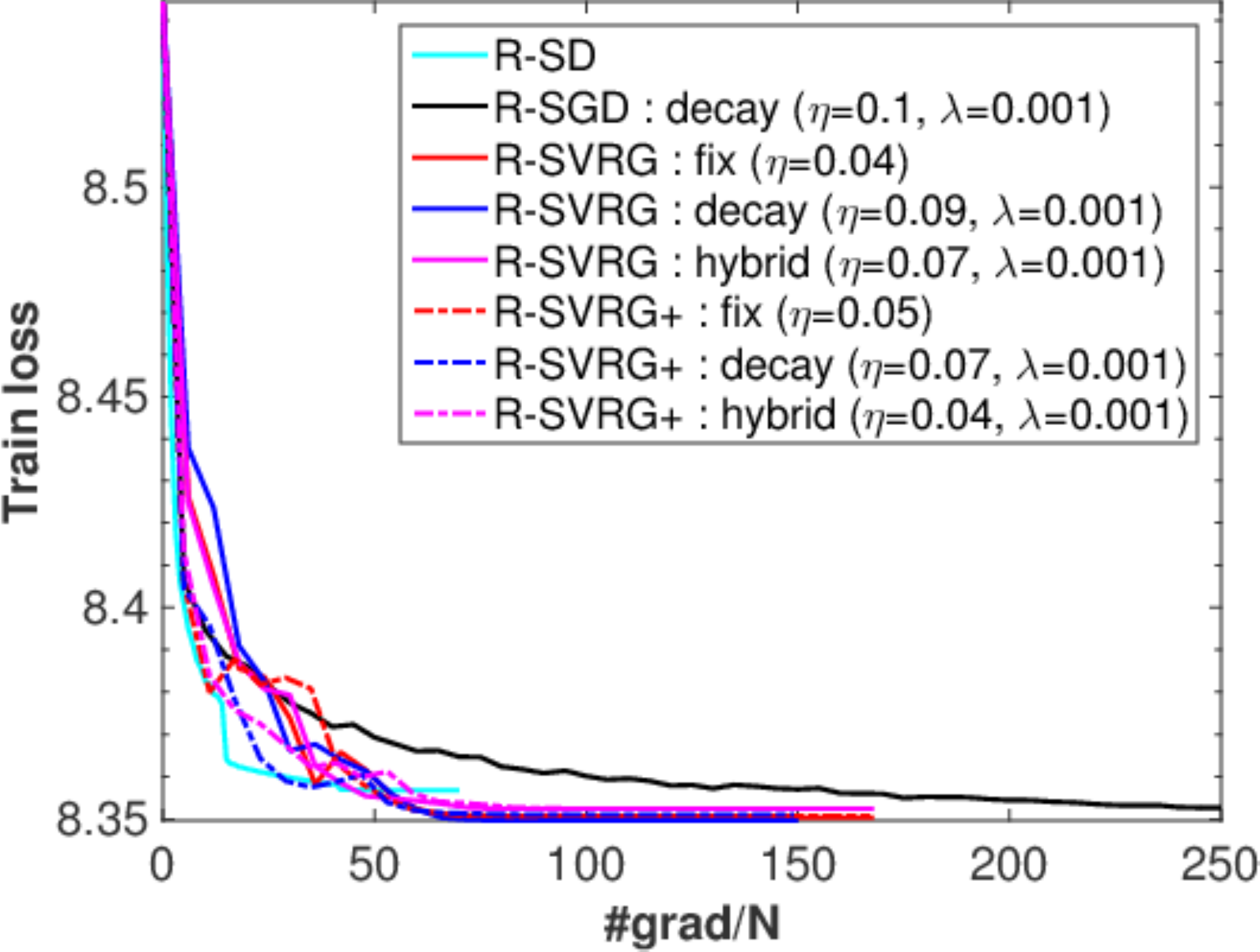}\\
		
		{\small (c-1) Train loss.}
		
	\end{center} 
	\end{minipage}
	\hspace*{-0.1cm}
	\begin{minipage}[t]{0.32\textwidth}
	\begin{center}
		\includegraphics[width=\textwidth]{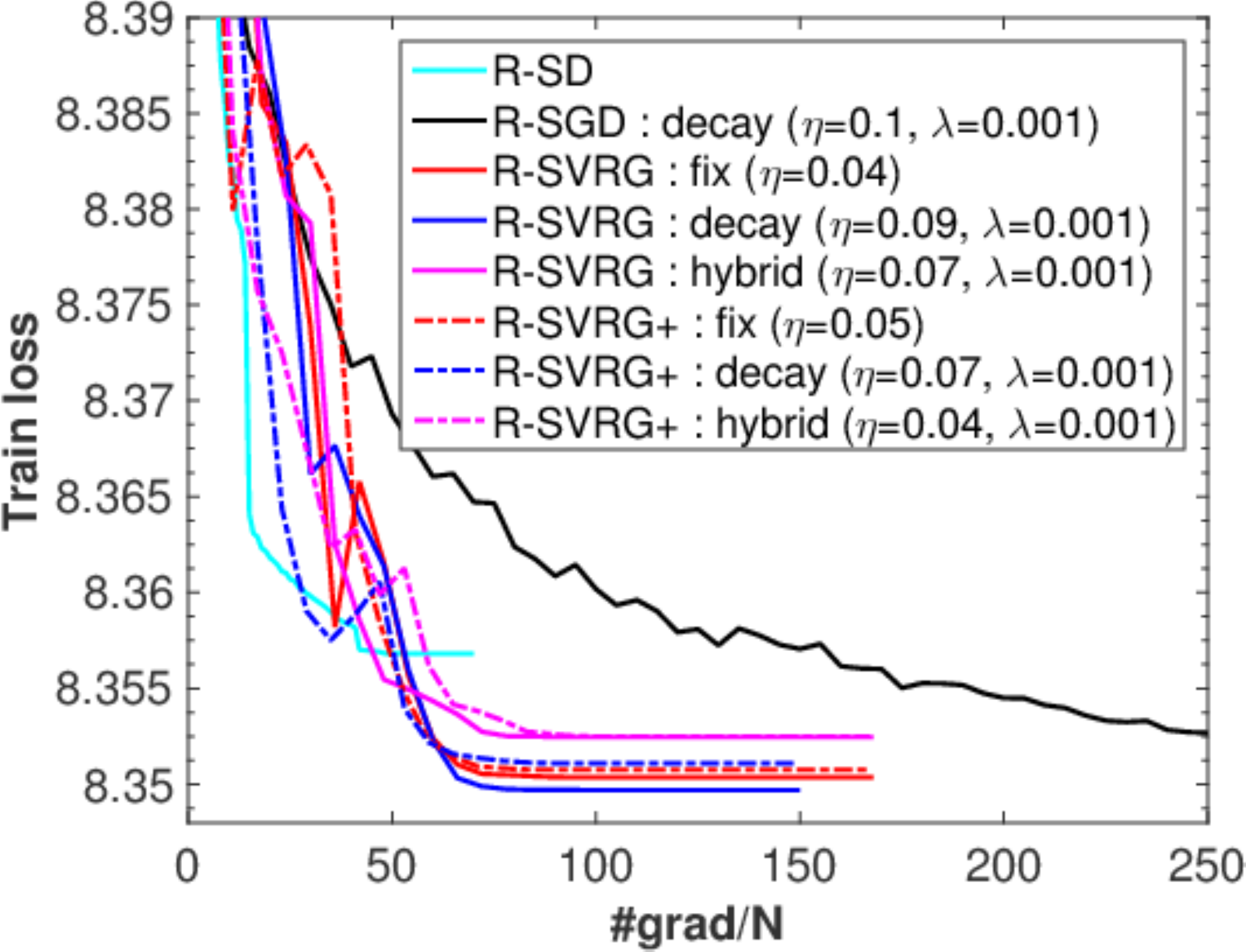}\\
		
		{\small (c-2) Train loss (enlarged). }
		
	\end{center} 
	\end{minipage}
	\hspace*{-0.1cm}
	\begin{minipage}[t]{0.32\textwidth}
	\begin{center}
		\includegraphics[width=\textwidth]{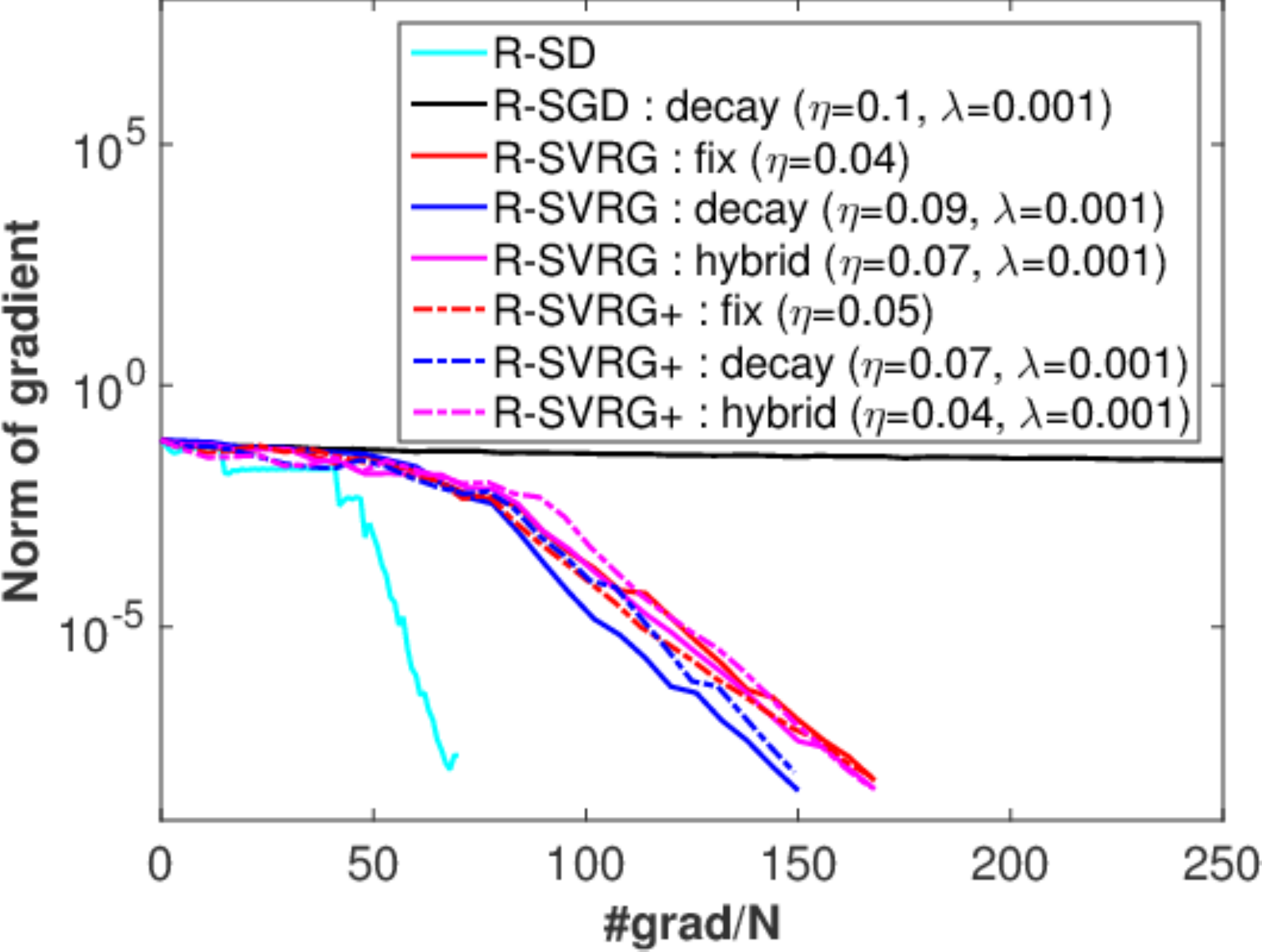}\\
		
		{\small (c-3) Norm of gradient.}
		
	\end{center} 
	\end{minipage}
\vspace*{0.5cm}
	
(c) $N=3000, d=100, r=10$.

\caption{The Karcher mean problem.}
\label{Appen_fig:KarcherMean_results}
\end{center}
\end{figure}

{\bf Karcher mean problem (additional experiments).} The manuscript shows the results for the case of $r=5$, where $N=1000$, $d=300$, Figure \ref{Appen_fig:KarcherMean_results}(a) shows the results of $r=10$. In this instance, R-SVRG+ shows superior performance than R-SVRG in terms of the final loss values. Furthermore, Figures \ref{Appen_fig:KarcherMean_results}(b) and (c) shows the results for the case with $N=1000$ and $d=100$ and with $r=5$ and  $r=10$, respectively. R-SVRG outperforms R-SGD and the final loss of R-SVRG is less than that of R-SD. 

\begin{figure}[t]
\begin{center}
	\begin{minipage}[t]{0.32\textwidth}
	\begin{center}
		\includegraphics[width=\textwidth]{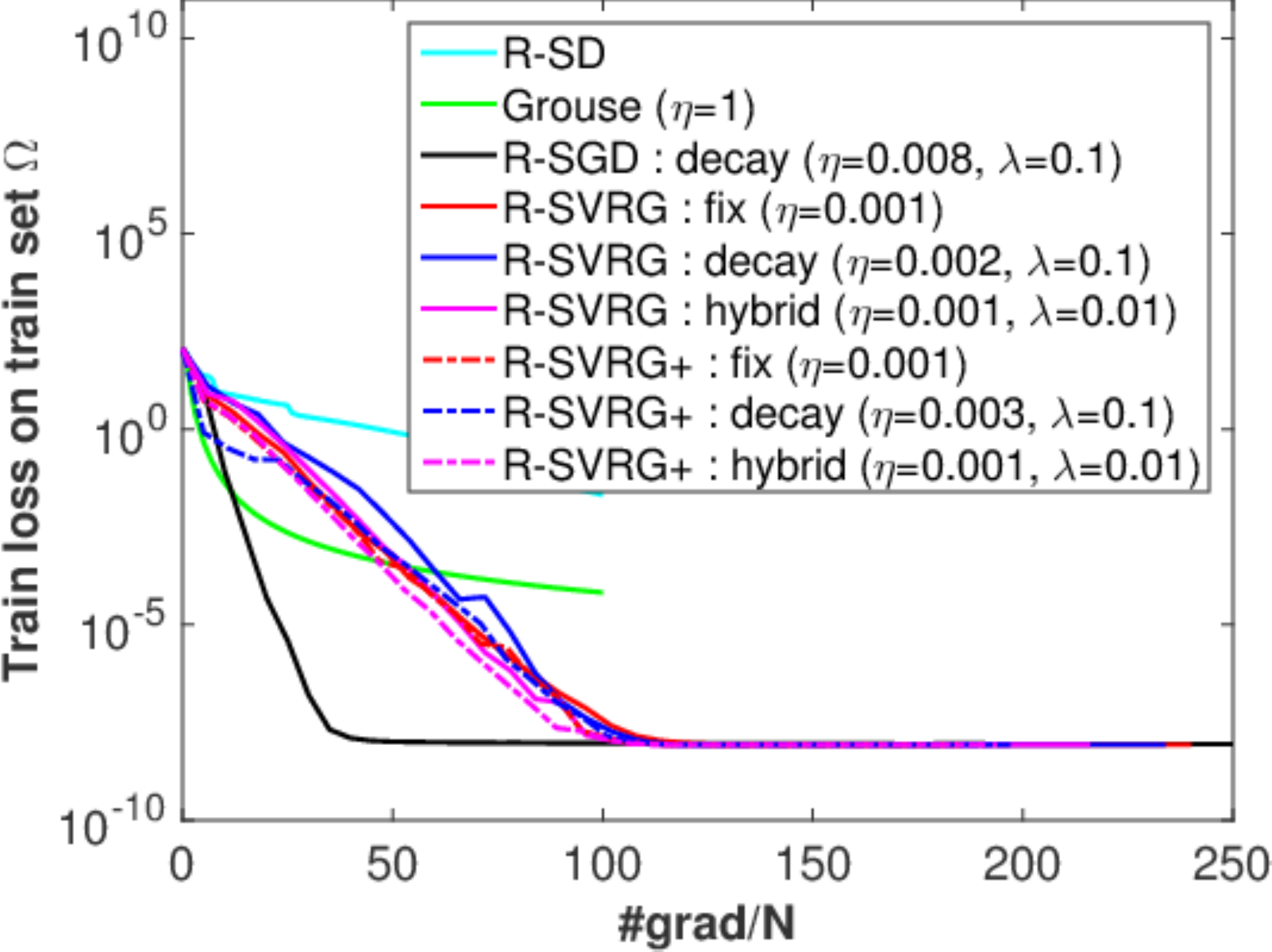}\\
		
		{\small (a-1) Train loss.}
		
	\end{center} 
	\end{minipage}
	\hspace*{-0.1cm}
	\begin{minipage}[t]{0.32\textwidth}
	\begin{center}
		\includegraphics[width=\textwidth]{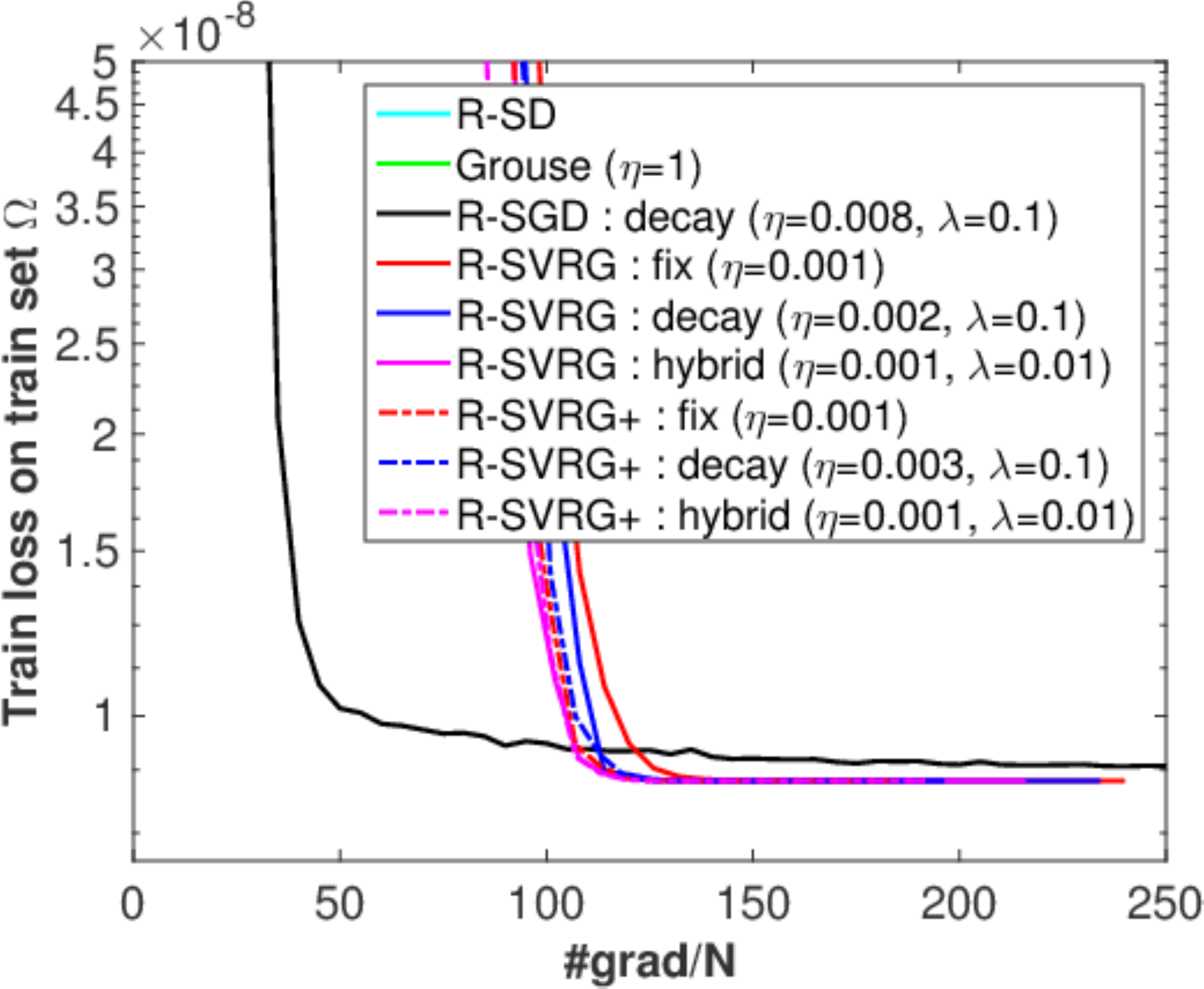}\\
		
		{\small (a-2) Train loss (enlarged). }
		
	\end{center} 
	\end{minipage}
	\hspace*{-0.1cm}
	\begin{minipage}[t]{0.32\textwidth}
	\begin{center}
		\includegraphics[width=\textwidth]{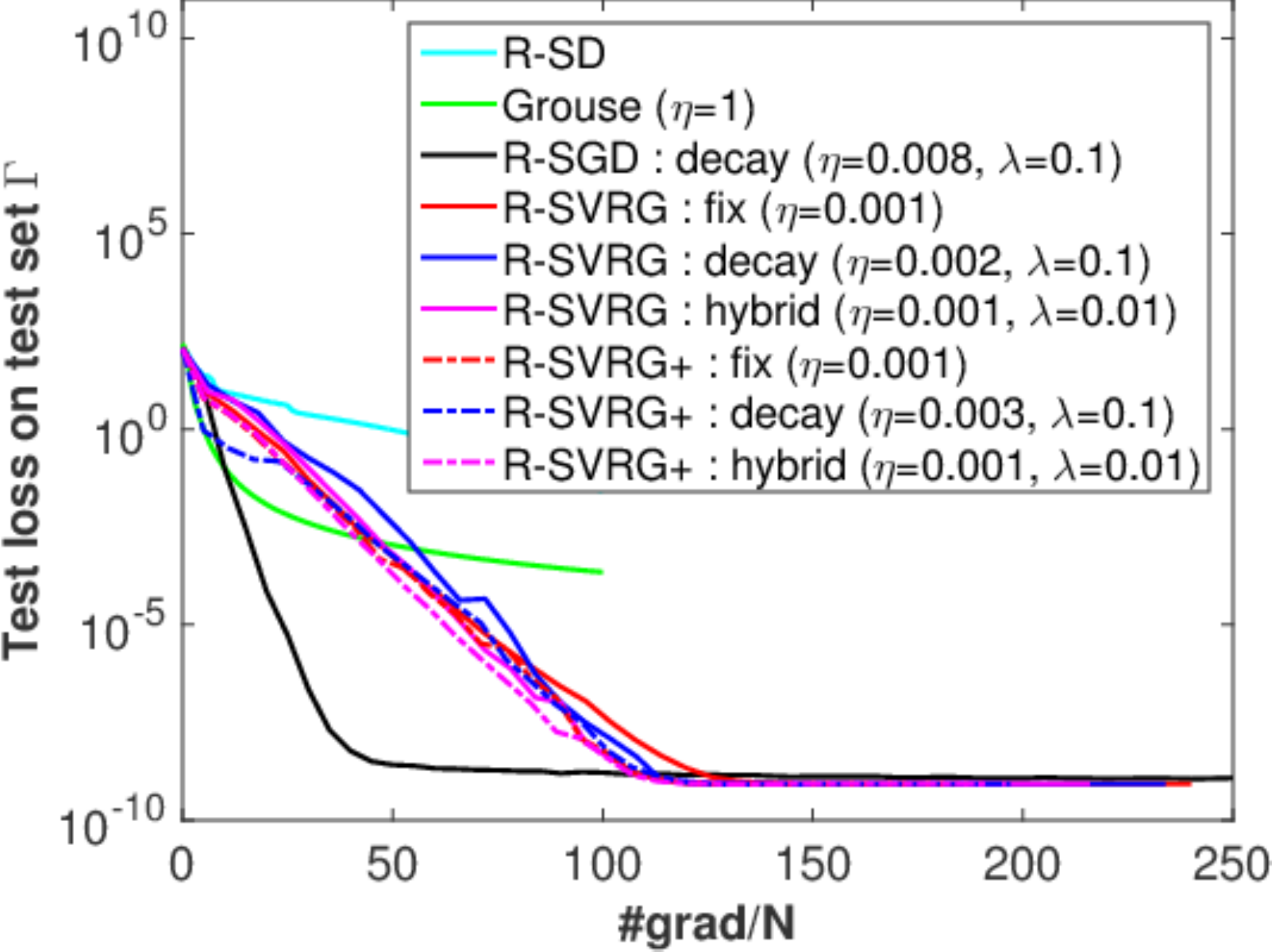}\\
		
		{\small (a-3) Test loss.}
		
	\end{center} 
	\end{minipage}
	\vspace*{0.4cm}
	
	\hspace*{-0.1cm}
	\begin{minipage}[t]{0.32\textwidth}
	\begin{center}
		\includegraphics[width=\textwidth]{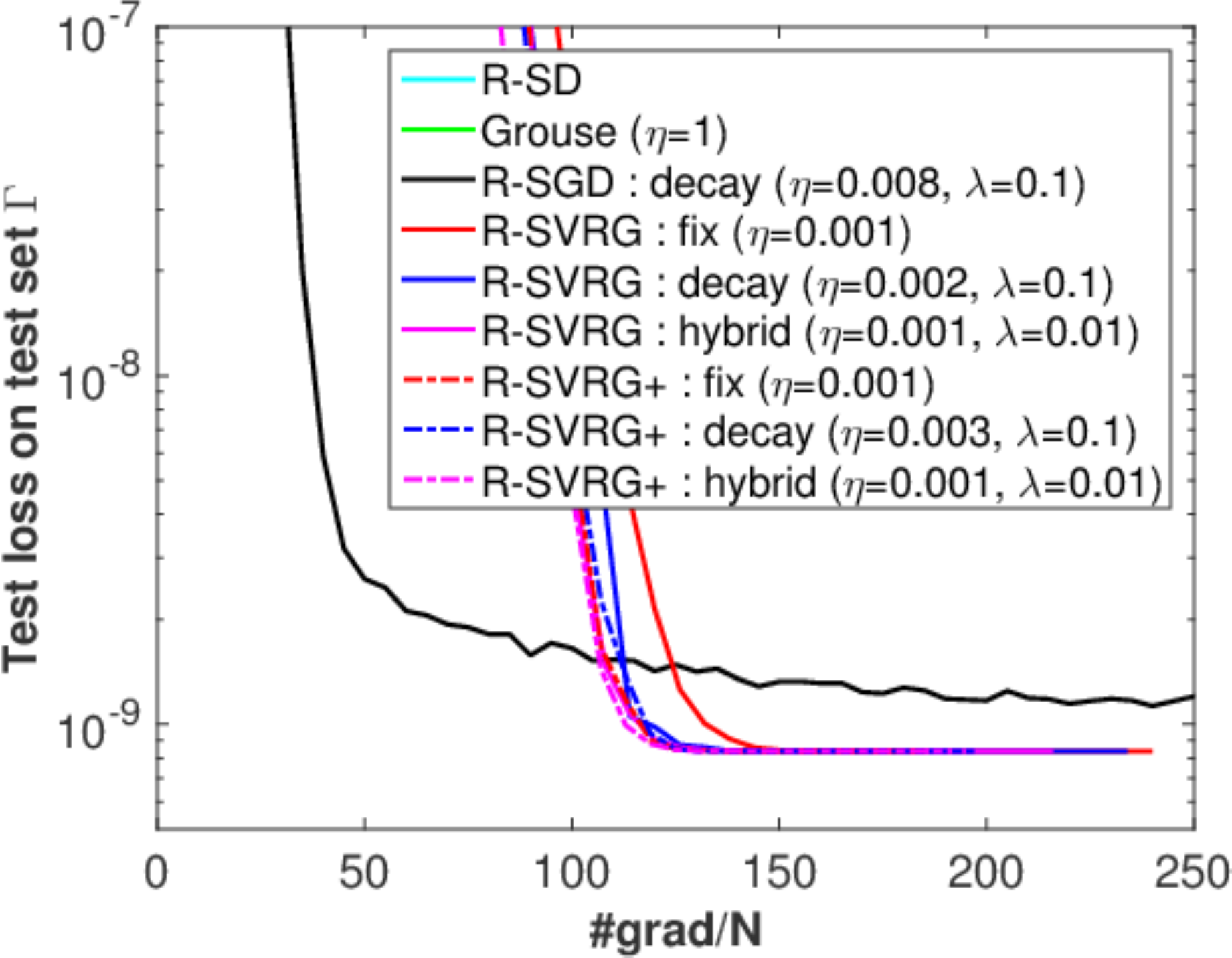}\\
		
		{\small (a-4) Test loss (enlarged).}
		
	\end{center} 
	\end{minipage}
	\begin{minipage}[t]{0.32\textwidth}
	\begin{center}
		\includegraphics[width=\textwidth]{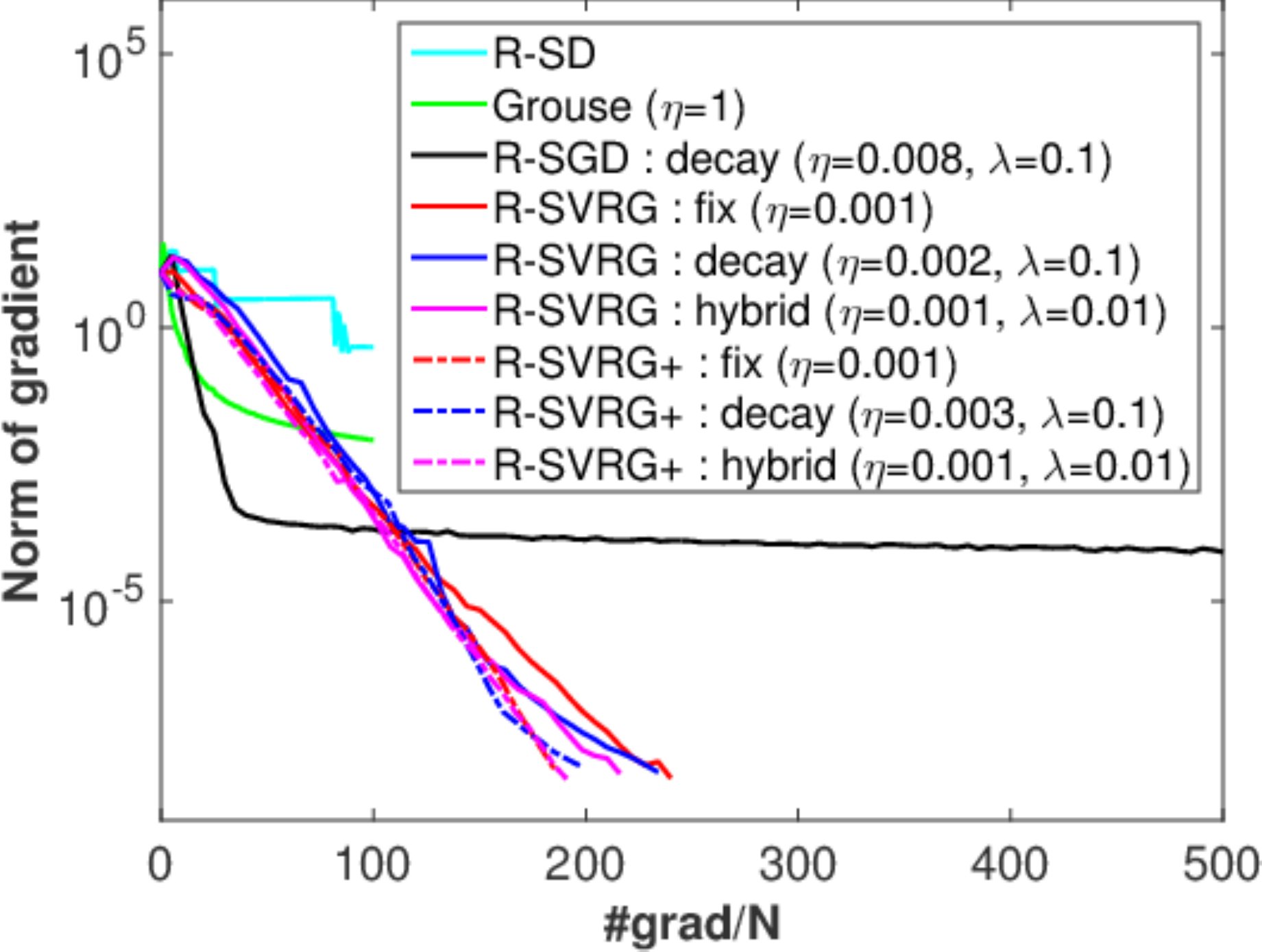}\\
		
		{\small (a-5) Norm of gradient.}
		
	\end{center} 
	\end{minipage}
\vspace*{0.5cm}
	
(a) $r=5$.
\vspace*{0.5cm}

	\begin{minipage}[t]{0.32\textwidth}
	\begin{center}
		\includegraphics[width=\textwidth]{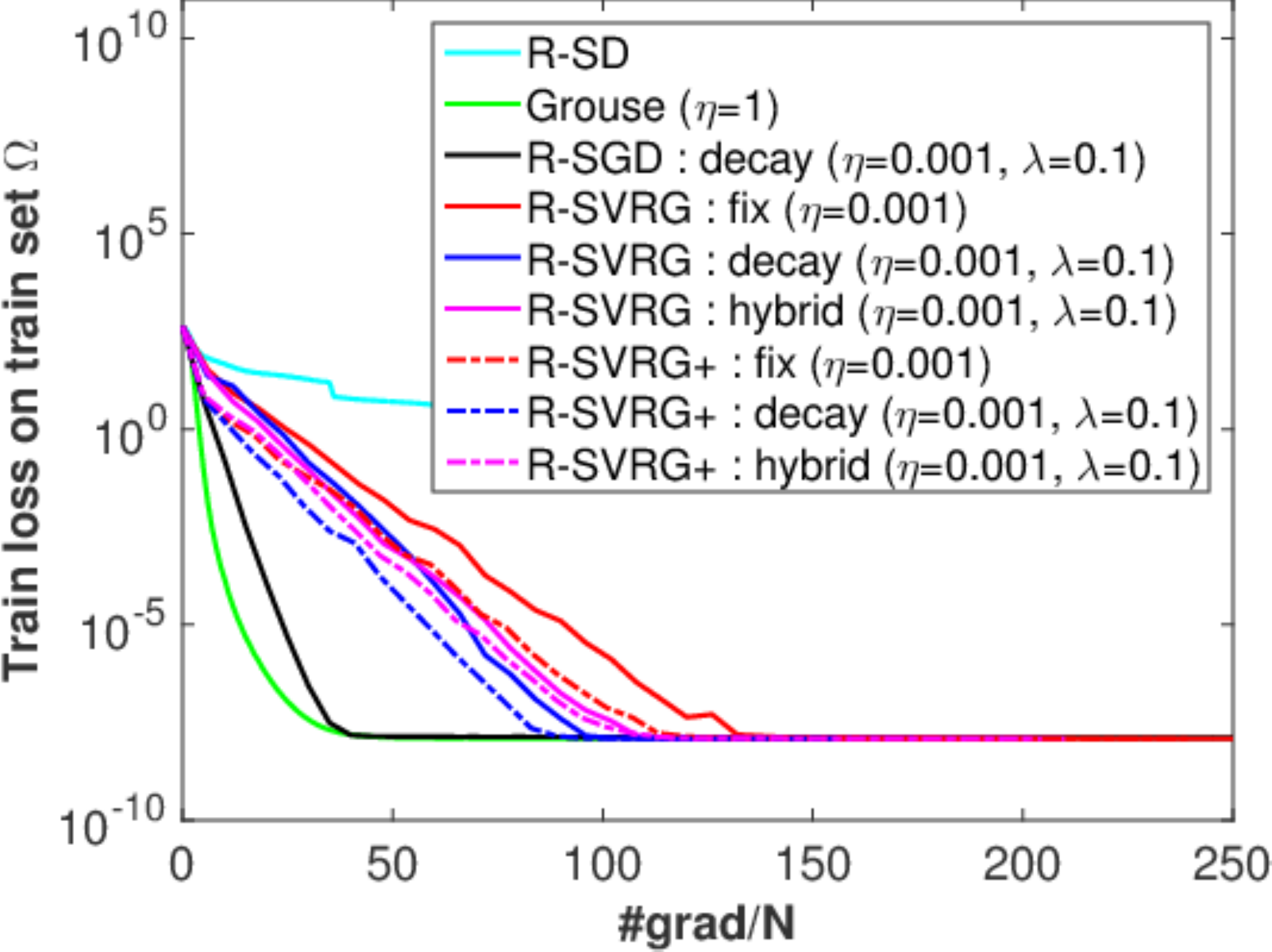}\\
		
		{\small (b-1) Train loss.}
		
	\end{center} 
	\end{minipage}
	\hspace*{-0.1cm}
	\begin{minipage}[t]{0.32\textwidth}
	\begin{center}
		\includegraphics[width=\textwidth]{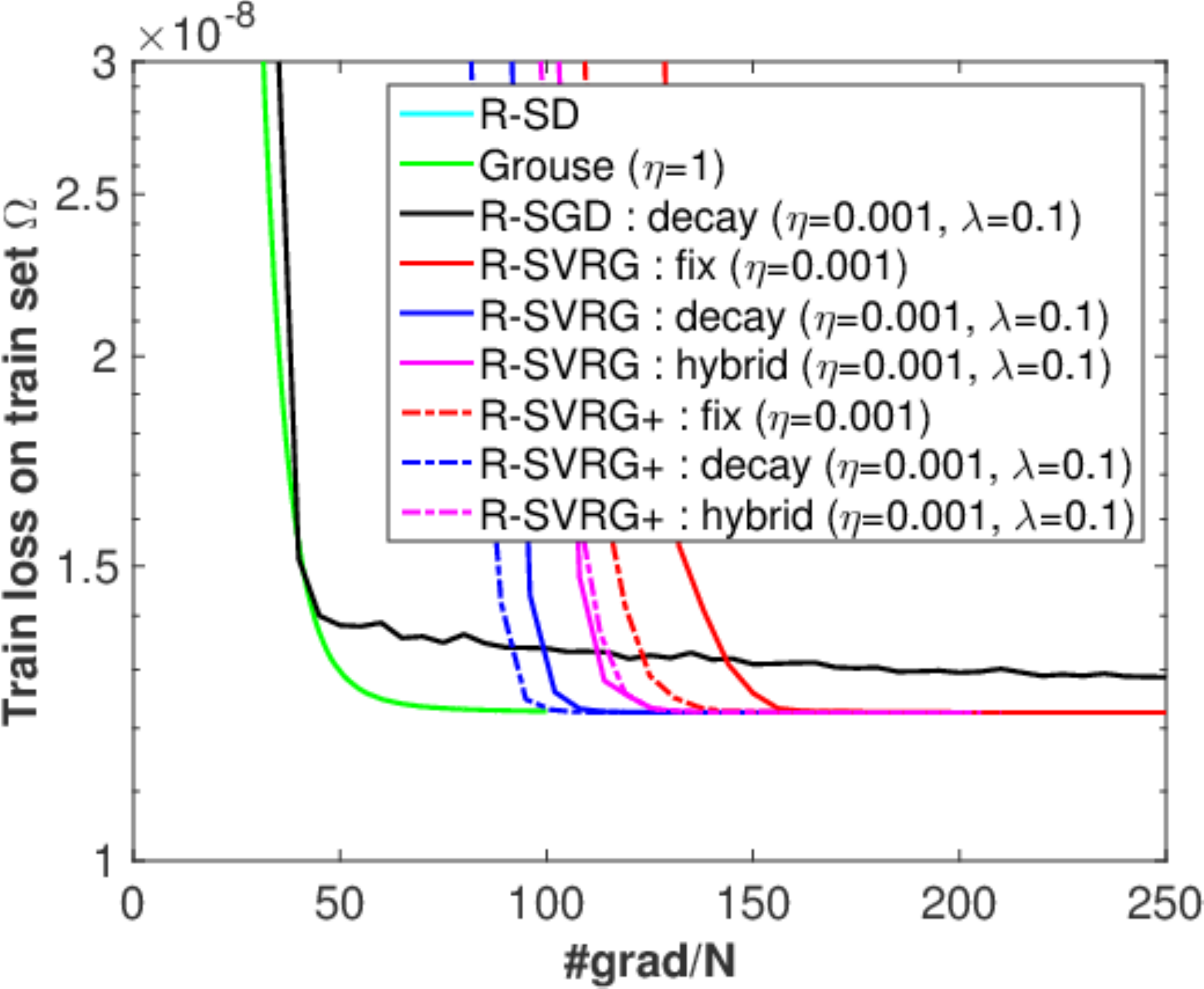}\\
		
		{\small (b-2) Train loss (enlarged). }
		
	\end{center} 
	\end{minipage}
	\hspace*{-0.1cm}
	\begin{minipage}[t]{0.32\textwidth}
	\begin{center}
		\includegraphics[width=\textwidth]{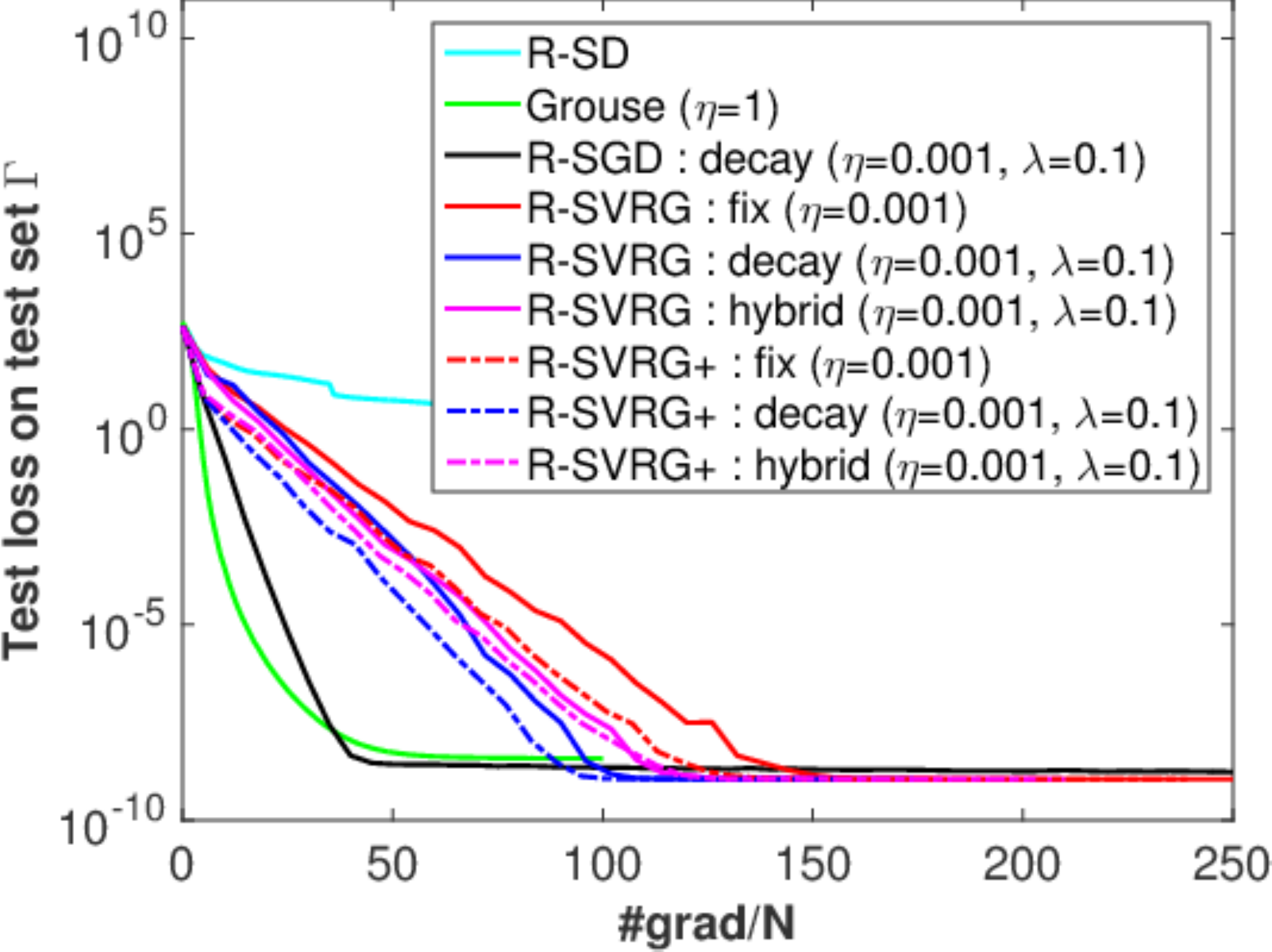}\\
		
		{\small (b-3) Test loss.}
		
	\end{center} 
	\end{minipage}
	\vspace*{0.4cm}
	
	\hspace*{-0.1cm}
	\begin{minipage}[t]{0.32\textwidth}
	\begin{center}
		\includegraphics[width=\textwidth]{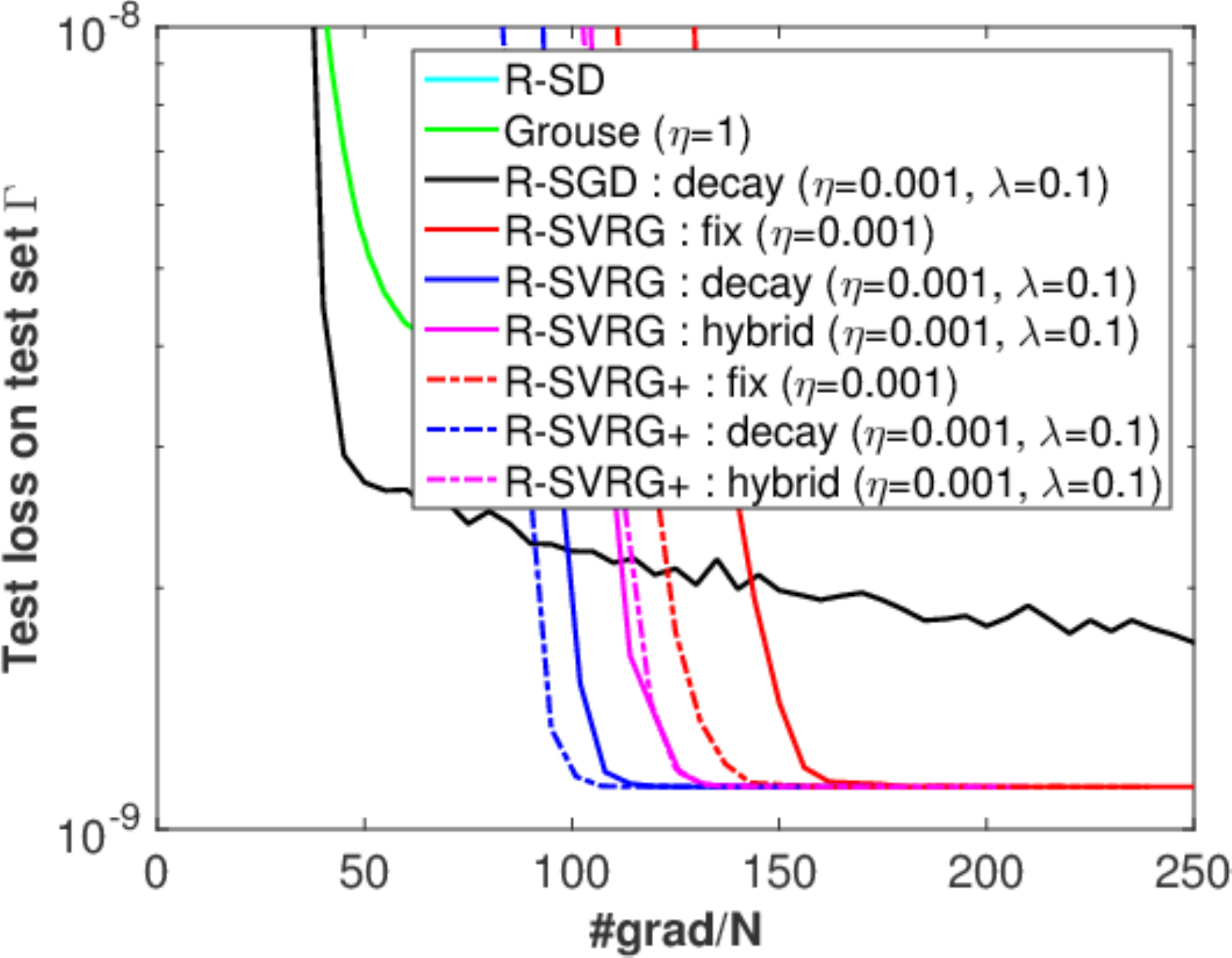}\\
		
		{\small (b-4) Test loss (enlarged).}
		
	\end{center} 
	\end{minipage}
	\begin{minipage}[t]{0.32\textwidth}
	\begin{center}
		\includegraphics[width=\textwidth]{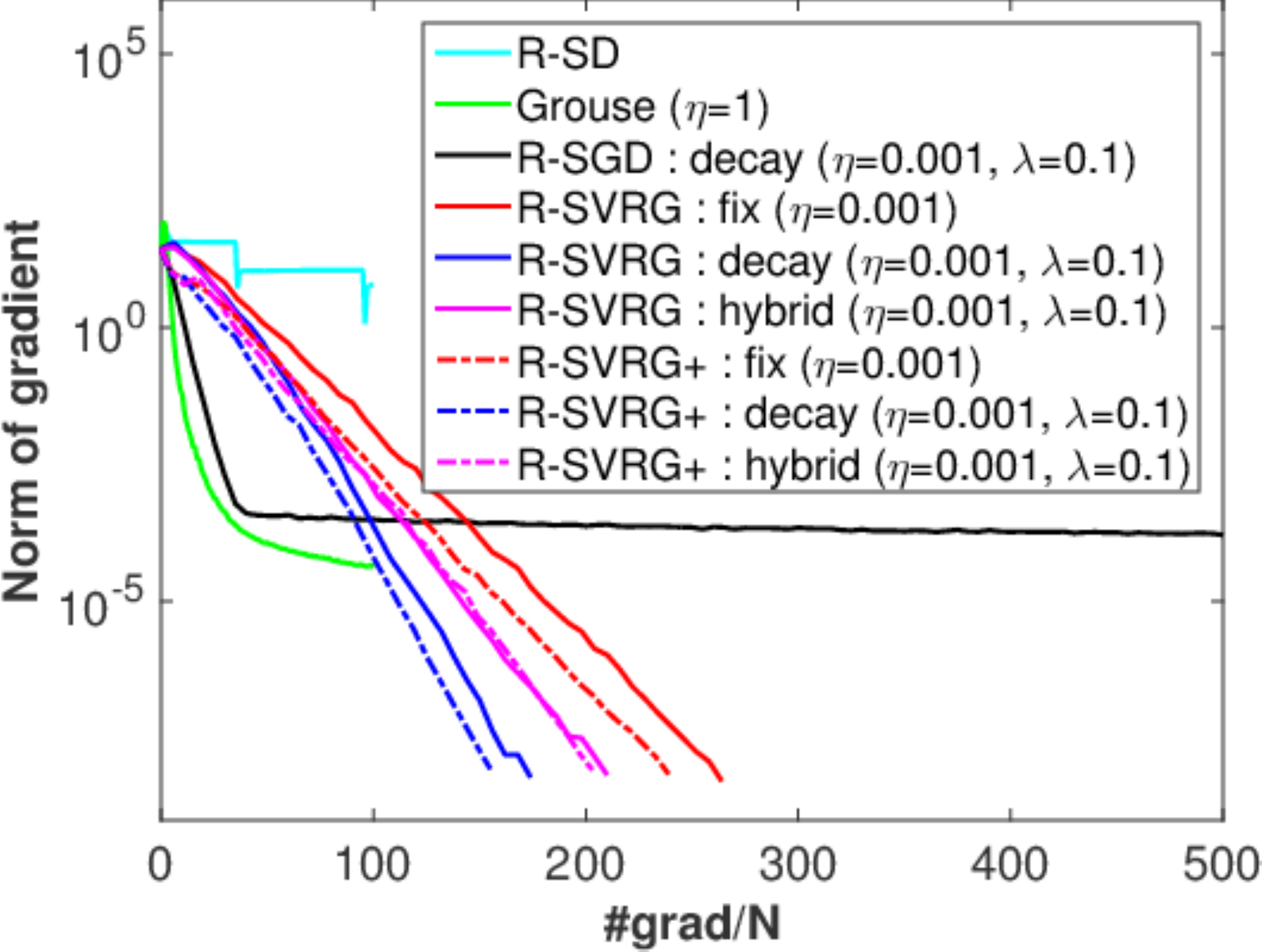}\\
		
		{\small (b-5) Norm of gradient.}
		
	\end{center} 
	\end{minipage}
\vspace*{0.5cm}
	
(b) $r=10$.
	
\caption{Low-rank matrix completion problem (synthetic dataset: $N=1000$, $d=500$).}
\label{Appen_fig:MC_results_synthetic_N_1000_d_500}
\end{center}	
\end{figure}

\begin{figure}[t]
\begin{center}
	\begin{minipage}[t]{0.32\textwidth}
	\begin{center}
		\includegraphics[width=\textwidth]{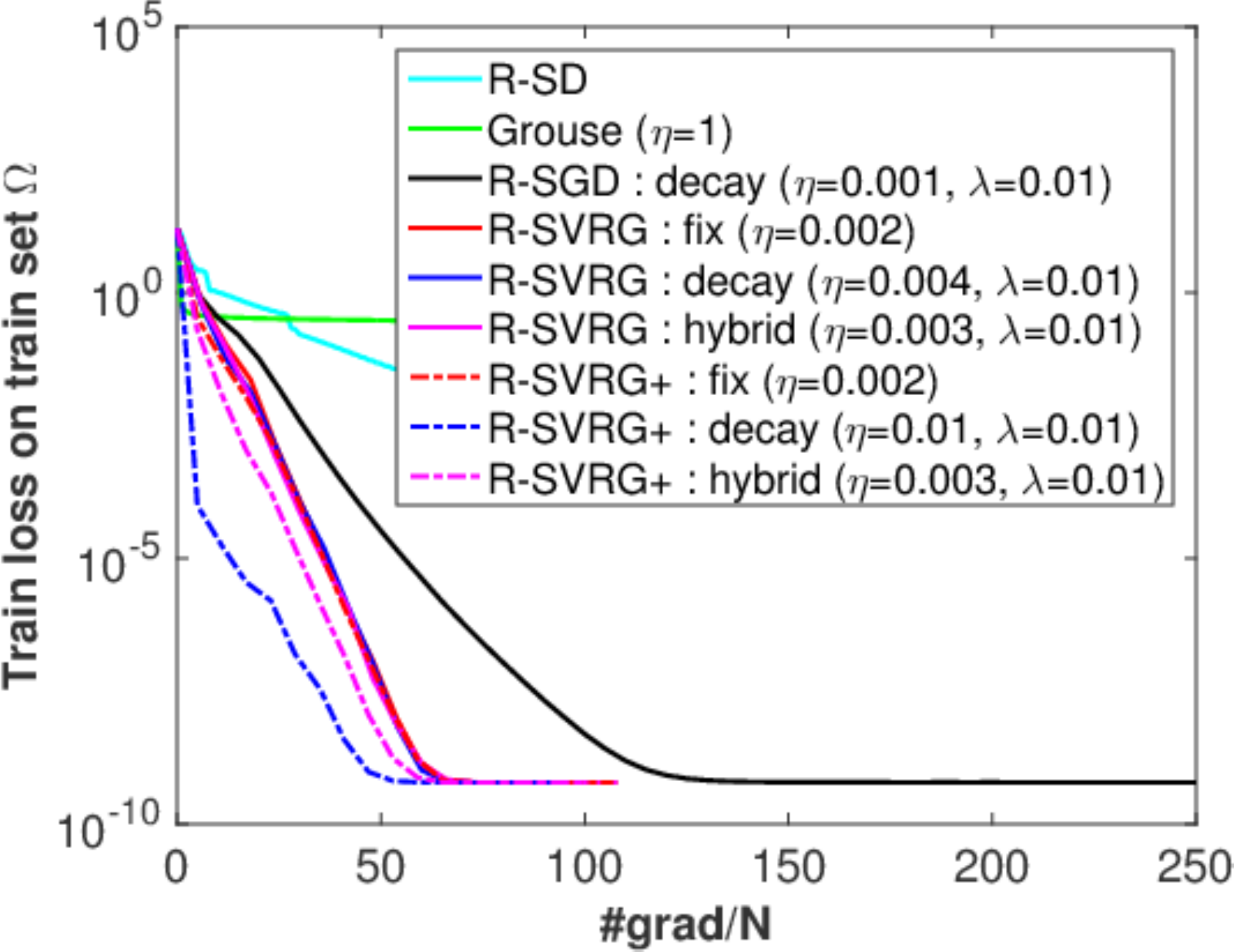}\\
		
		{\small (a-1) Train loss.}
		
	\end{center} 
	\end{minipage}
	\hspace*{-0.1cm}
	\begin{minipage}[t]{0.32\textwidth}
	\begin{center}
		\includegraphics[width=\textwidth]{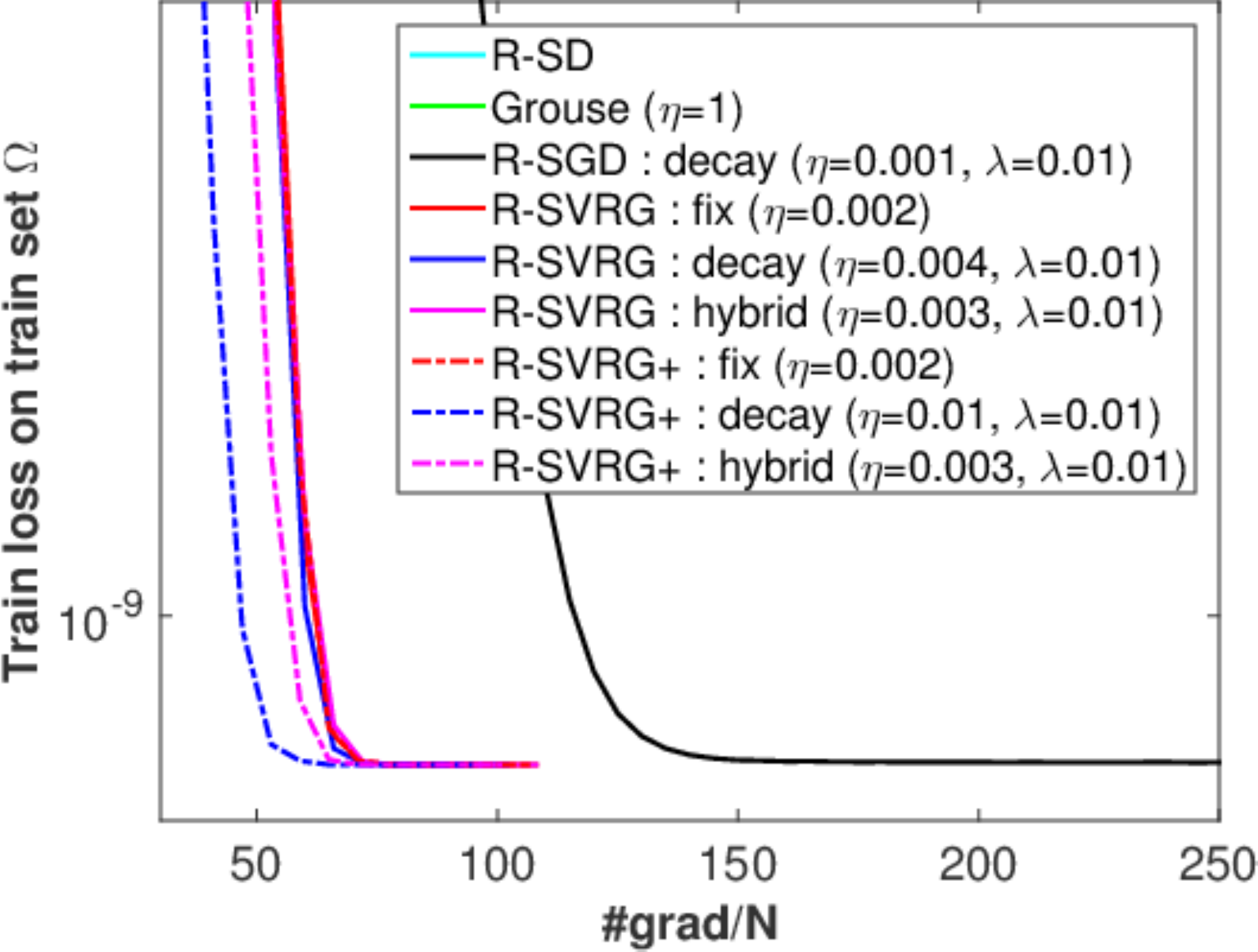}\\
		
		{\small (a-2) Train loss (enlarged). }
		
	\end{center} 
	\end{minipage}
	\hspace*{-0.1cm}
	\begin{minipage}[t]{0.32\textwidth}
	\begin{center}
		\includegraphics[width=\textwidth]{results_pdf/mc/mc_test_MSE_N5000_d500_r5.pdf}\\
		
		{\small (a-3) Test loss.}
		
	\end{center} 
	\end{minipage}
	\vspace*{0.4cm}
	
	\hspace*{-0.1cm}
	\begin{minipage}[t]{0.32\textwidth}
	\begin{center}
		\includegraphics[width=\textwidth]{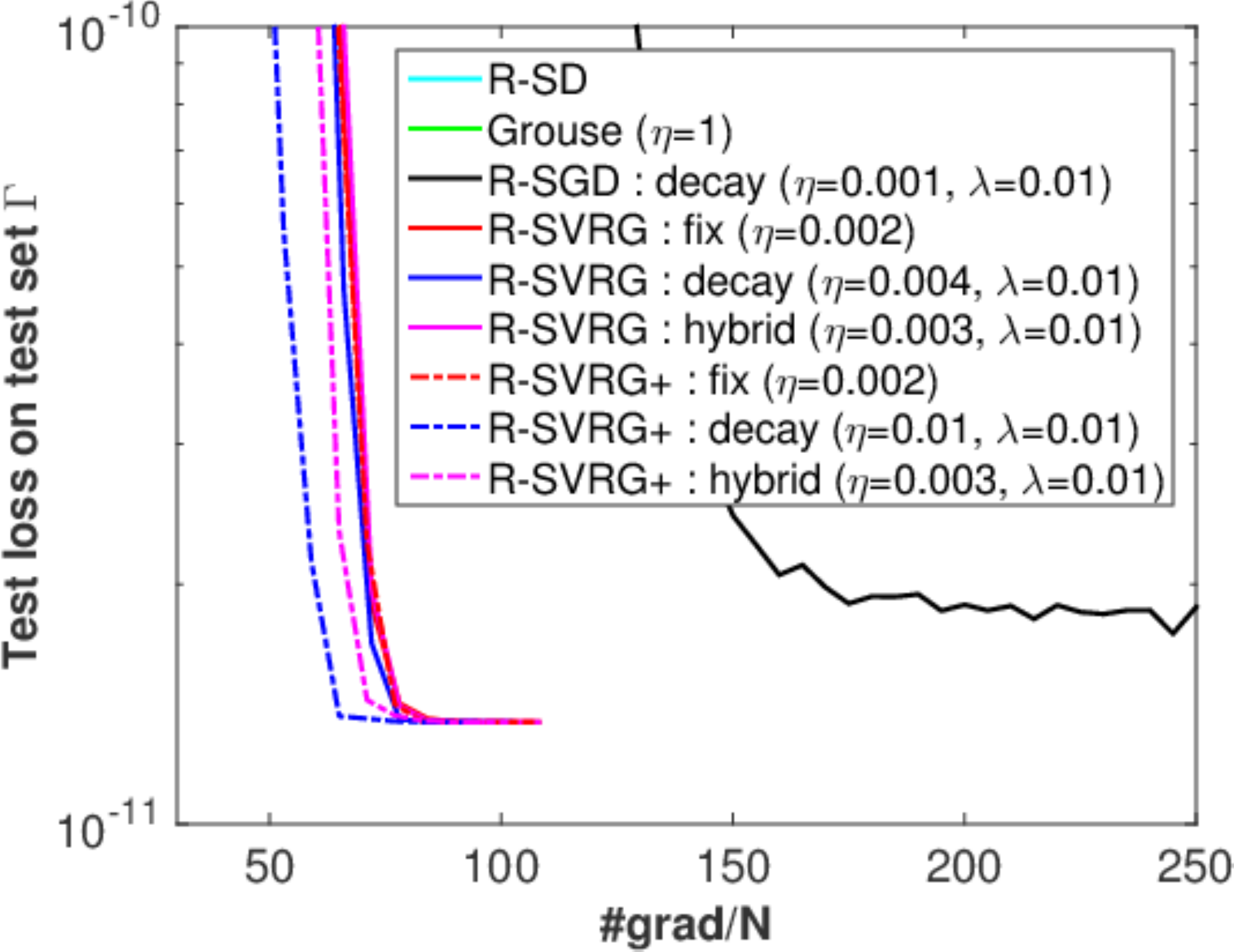}\\
		
		{\small (a-4) Test loss (enlarged).}
		
	\end{center} 
	\end{minipage}
	\begin{minipage}[t]{0.32\textwidth}
	\begin{center}
		\includegraphics[width=\textwidth]{results_pdf/mc/mc_gnorm_N5000_d500_r5.pdf}\\
		
		{\small (a-5) Norm of gradient.}
		
	\end{center} 
	\end{minipage}
\vspace*{0.5cm}
	
(a) $r=5$.
\vspace*{0.5cm}

	\begin{minipage}[t]{0.32\textwidth}
	\begin{center}
		\includegraphics[width=\textwidth]{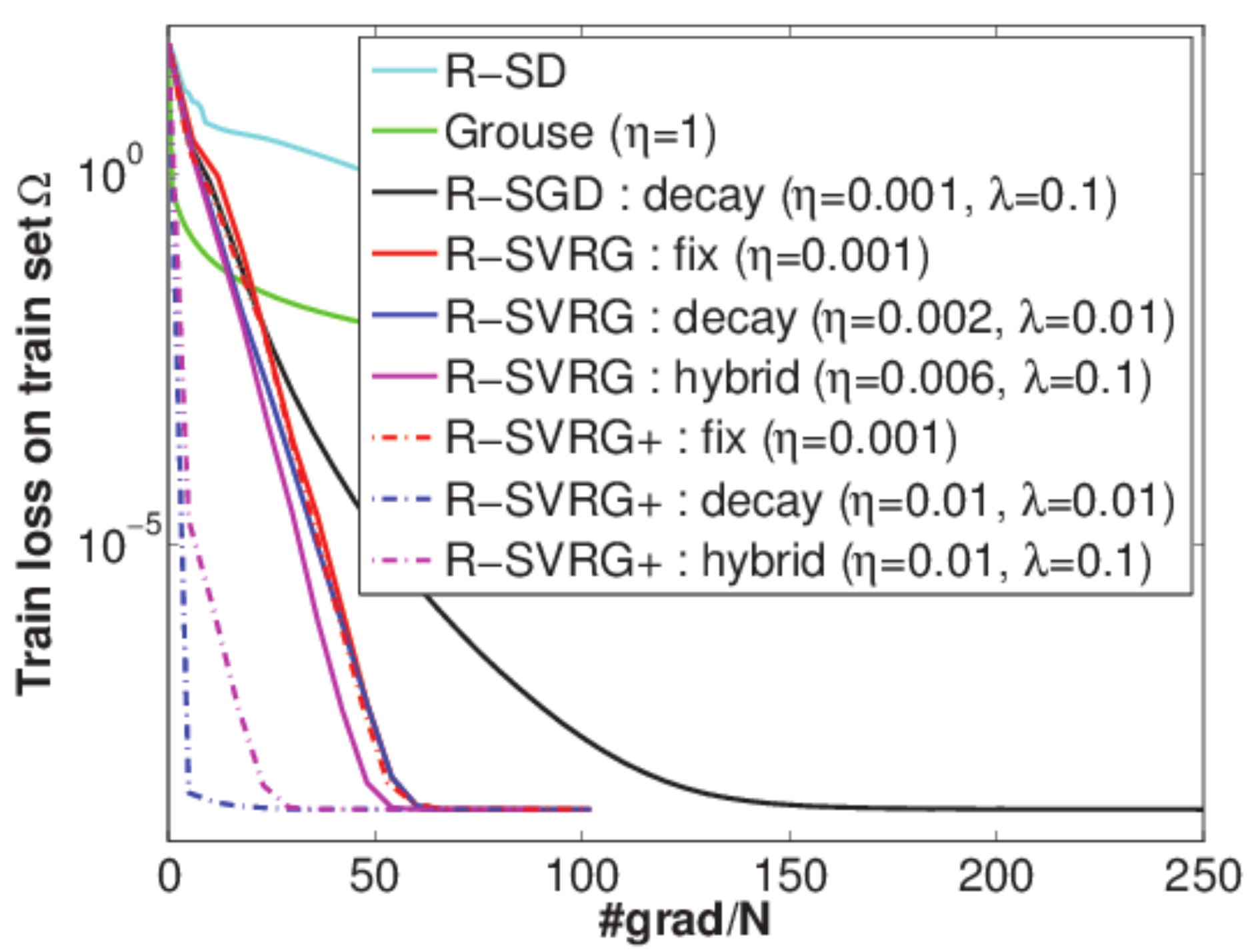}\\
		
		{\small (b-1) Train loss.}
		
	\end{center} 
	\end{minipage}
	\hspace*{-0.1cm}
	\begin{minipage}[t]{0.32\textwidth}
	\begin{center}
		\includegraphics[width=\textwidth]{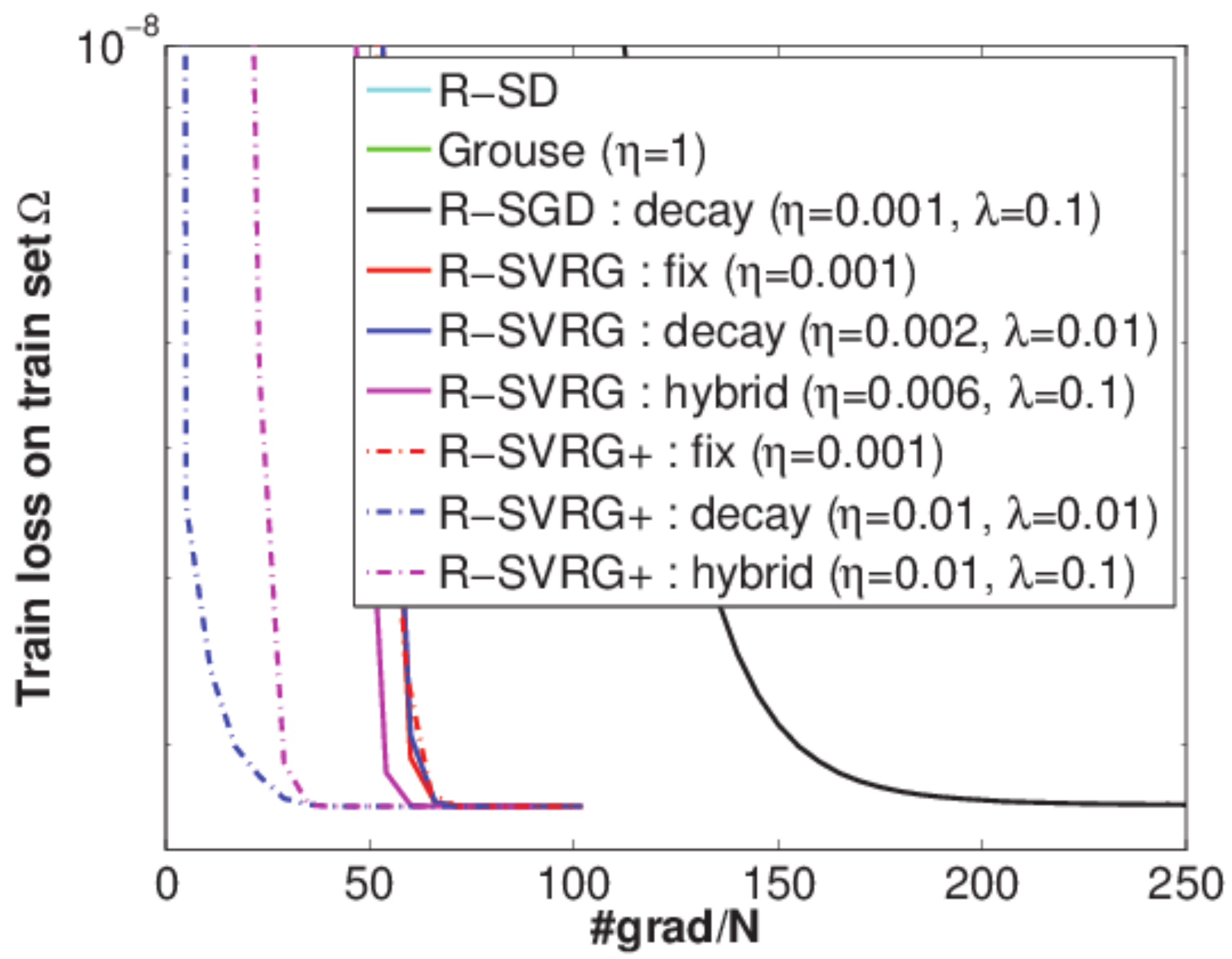}\\
		
		{\small (b-2) Train loss (enlarged). }
		
	\end{center} 
	\end{minipage}
	\hspace*{-0.1cm}
	\begin{minipage}[t]{0.32\textwidth}
	\begin{center}
		\includegraphics[width=\textwidth]{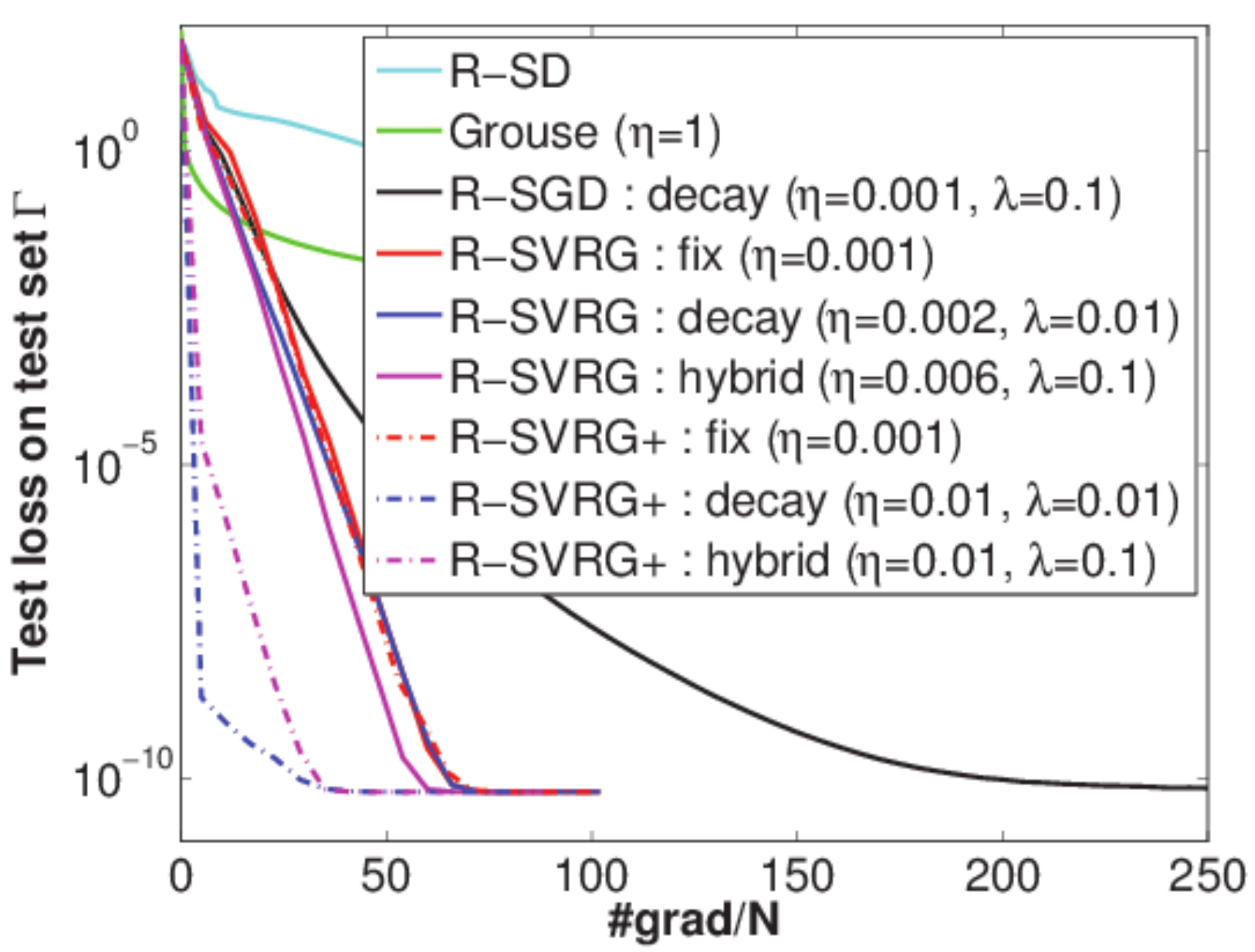}\\
		
		{\small (b-3) Test loss.}
		
	\end{center} 
	\end{minipage}
	\vspace*{0.4cm}
	
	\hspace*{-0.1cm}
	\begin{minipage}[t]{0.32\textwidth}
	\begin{center}
		\includegraphics[width=\textwidth]{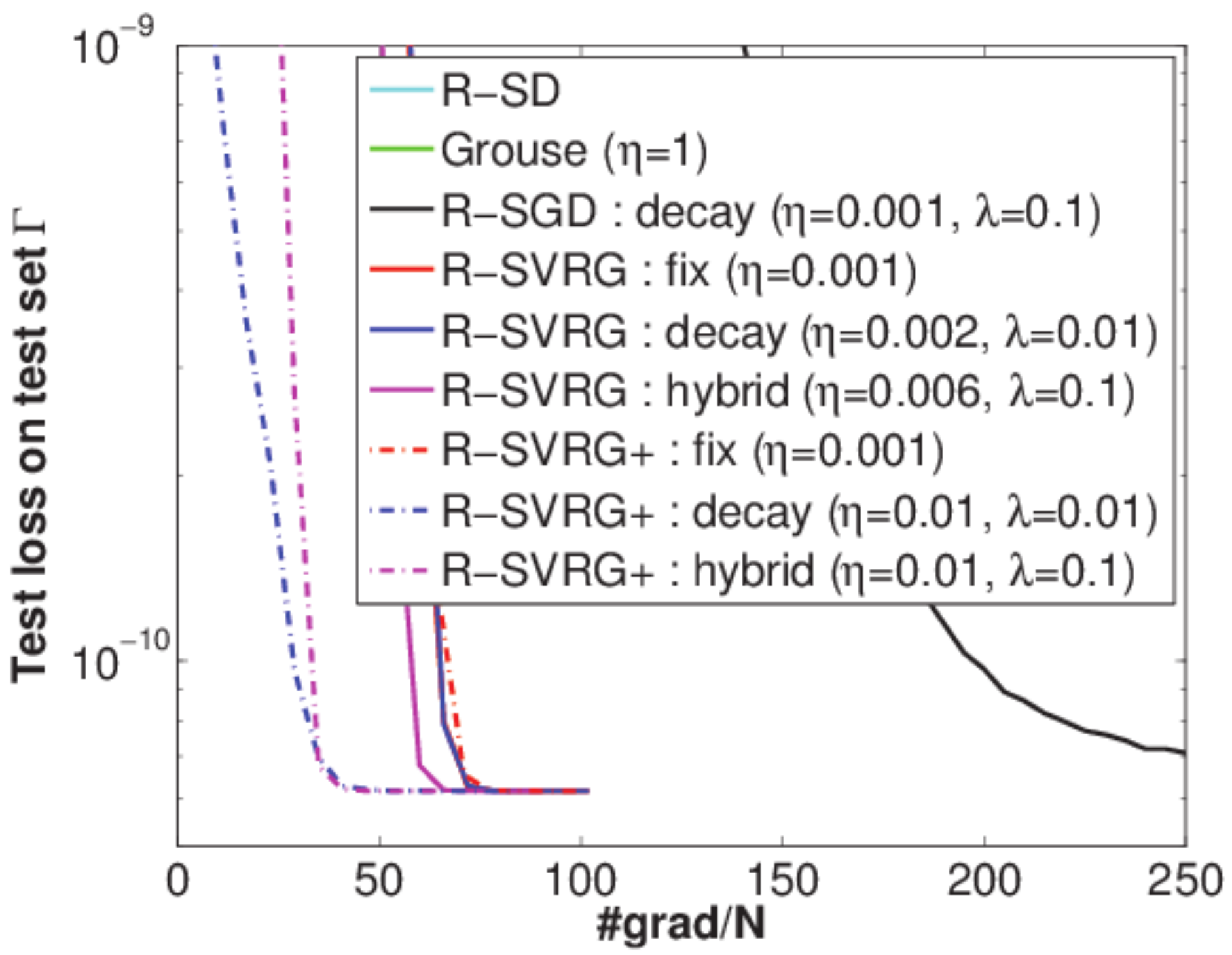}\\
		
		{\small (b-4) Test loss (enlarged).}
		
	\end{center} 
	\end{minipage}
	\begin{minipage}[t]{0.32\textwidth}
	\begin{center}
		\includegraphics[width=\textwidth]{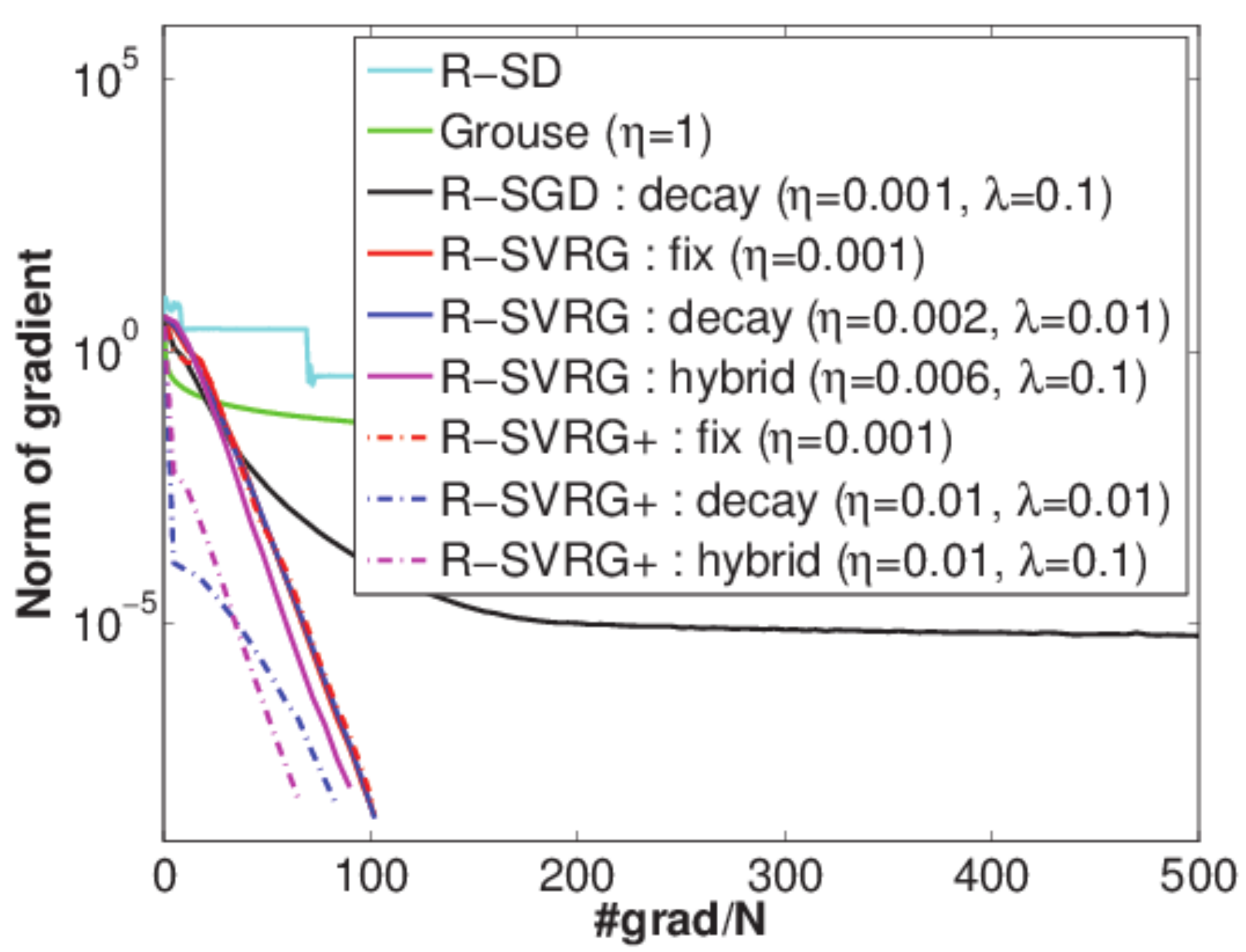}\\
		
		{\small (b-5) Norm of gradient.}
		
	\end{center} 
	\end{minipage}
\vspace*{0.5cm}
	
(b) $r=10$.

\caption{The low-rank matrix completion problem (synthetic dataset: $N=5000$, $d=500$).}
\label{Appen_fig:MC_results_synthetic_N_5000_d_500}
\end{center}	
\end{figure}

{\bf Matrix completion problem (additional experiments).} We show the additional results for the smaller instances $N=1000$, $d=500$, and $r=5$ in Figure \ref{Appen_fig:MC_results_synthetic_N_1000_d_500}(a). R-SGD and Grouse decrease very fast in the beginning, but R-SVRG(+) converges to lower values. Figure \ref{Appen_fig:MC_results_synthetic_N_1000_d_500}(b) also shows the case of $r=10$. Although Grouse indicates the fastest convergence, and gives the lowest values in the train loss as the same R-SVRG(+), R-SVRG(+) outperforms Grouse and R-SGD in test loss. In addition, we show all the results of for $N=5000$, $d=500$, and $r=5$ in Figure \ref{Appen_fig:MC_results_synthetic_N_5000_d_500}(a). These experiments are identical to those in the manuscript.  The results show the superior performance of our proposed algorithms. Furthermore, we consider a higher rank of  $r=10$ in Figure \ref{Appen_fig:MC_results_synthetic_N_5000_d_500}(b). The results also show that R-SVRG yield better performances than Grouse and R-SGD. 

\begin{figure}[t]
\begin{center}
	\begin{minipage}[t]{0.32\textwidth}
	\begin{center}
		\includegraphics[width=\textwidth]{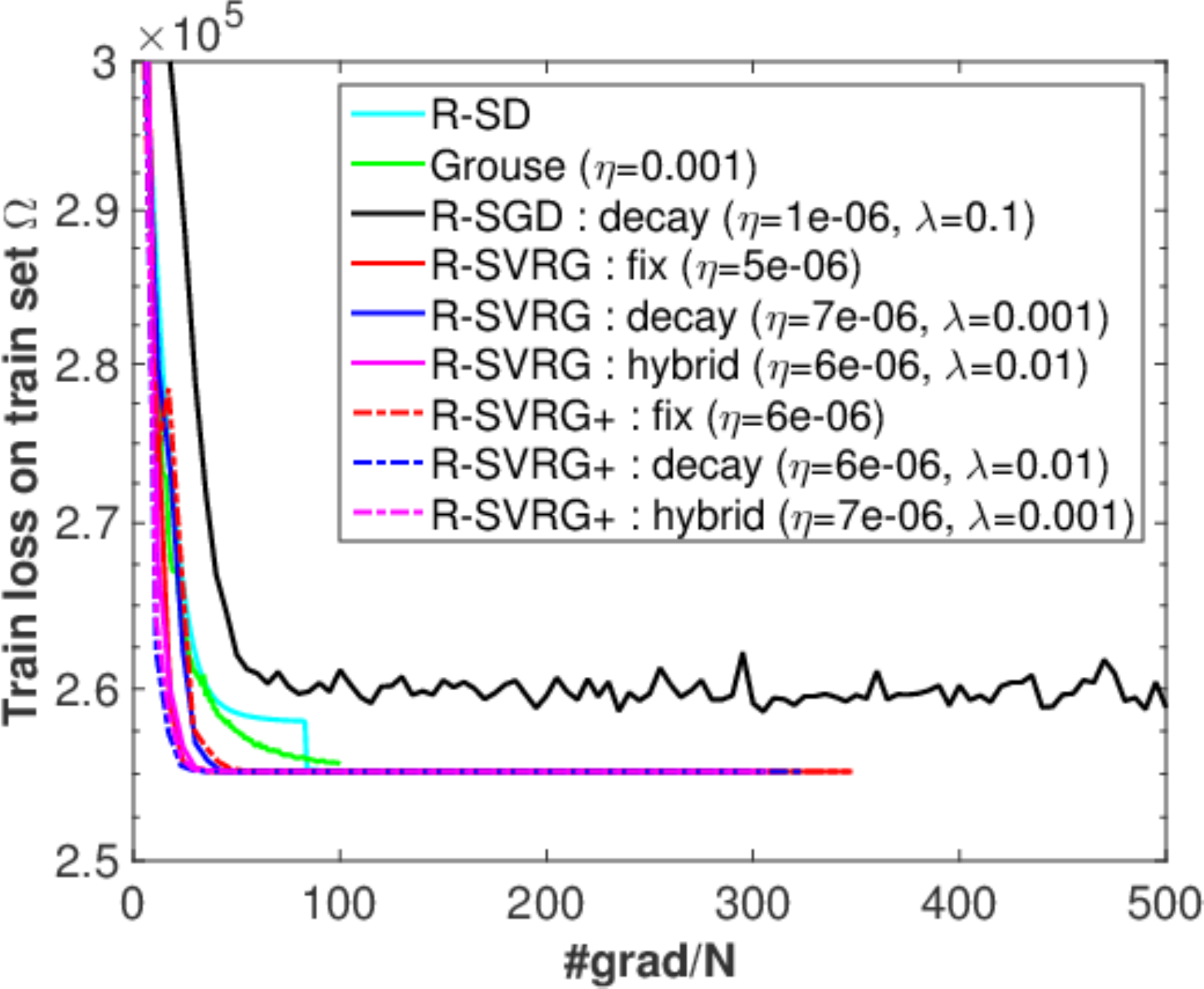}\\
		
		{\small (a-1) Train loss.}
		
	\end{center} 
	\end{minipage}
	\hspace*{-0.1cm}
	\begin{minipage}[t]{0.32\textwidth}
	\begin{center}
		\includegraphics[width=\textwidth]{results_pdf/mc_jester/mc_jester_train_MSE_enlarge_N100_d24983_r5.pdf}\\
		
		{\small (a-2) Train loss (enlarged). }
		
	\end{center} 
	\end{minipage}
	\hspace*{-0.1cm}
	\begin{minipage}[t]{0.32\textwidth}
	\begin{center}
		\includegraphics[width=\textwidth]{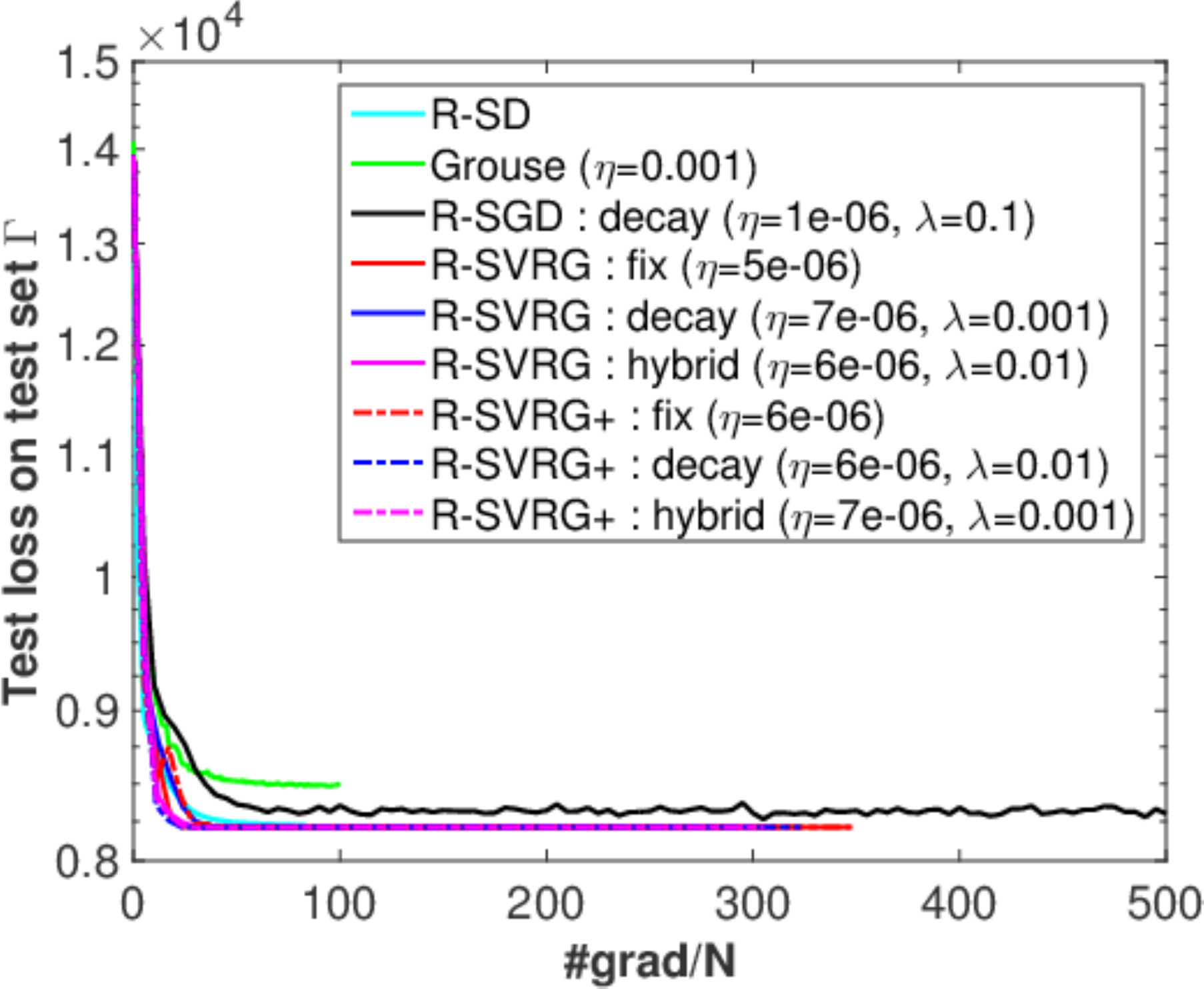}\\
		
		{\small (a-3) Test loss.}
		
	\end{center} 
	\end{minipage}
	\vspace*{0.4cm}
	
	\hspace*{-0.1cm}
	\begin{minipage}[t]{0.32\textwidth}
	\begin{center}
		\includegraphics[width=\textwidth]{results_pdf/mc_jester/mc_jester_test_MSE_enlarge_N100_d24983_r5.pdf}\\
		
		{\small (a-4) Test loss (enlarged).}
		
	\end{center} 
	\end{minipage}
	\begin{minipage}[t]{0.32\textwidth}
	\begin{center}
		\includegraphics[width=\textwidth]{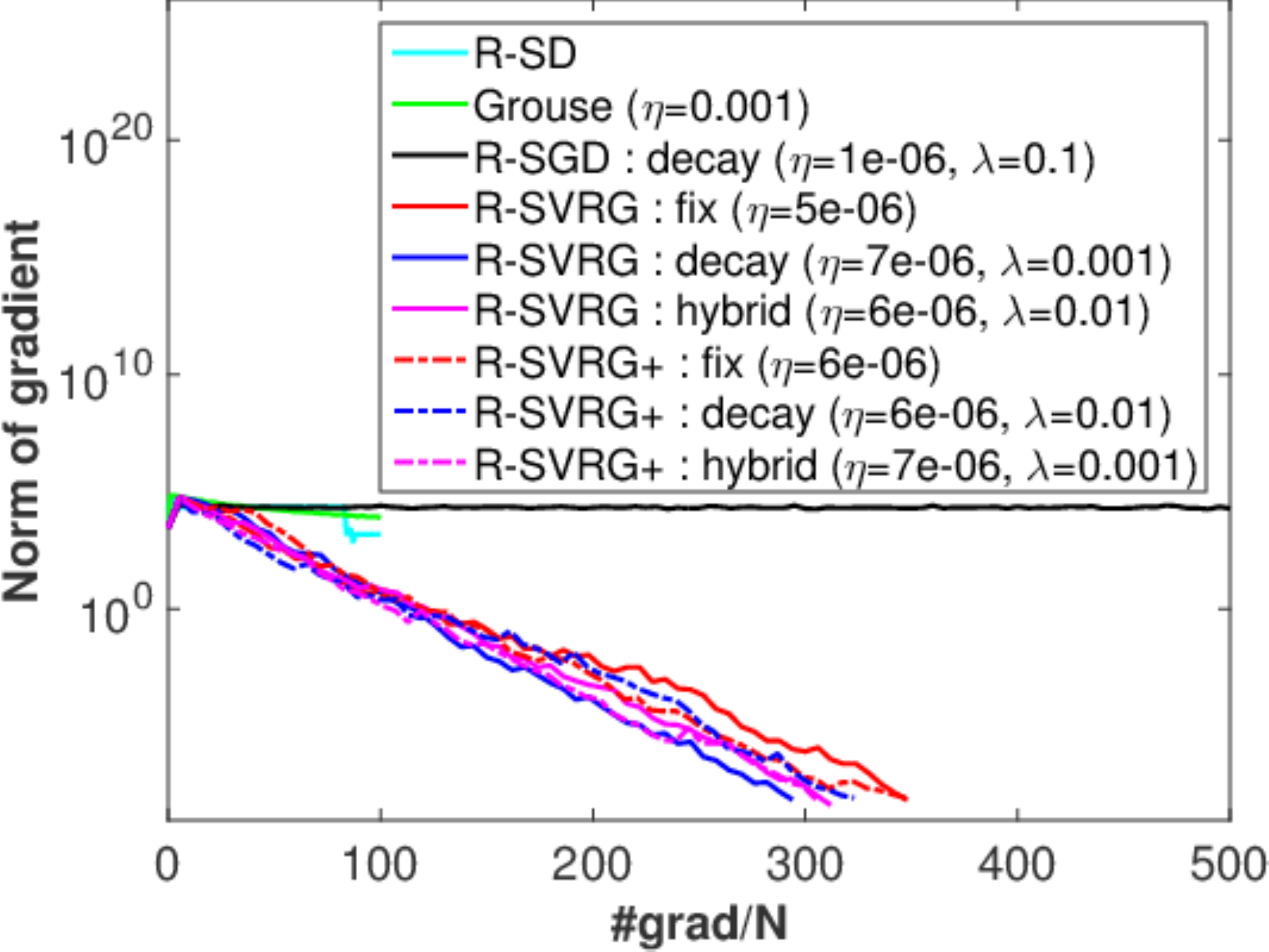}\\
		
		{\small (a-5) Norm of gradient.}
		
	\end{center} 
	\end{minipage}
\vspace*{0.5cm}
	
(a) $r=5$.
\vspace*{0.5cm}

	\begin{minipage}[t]{0.32\textwidth}
	\begin{center}
		\includegraphics[width=\textwidth]{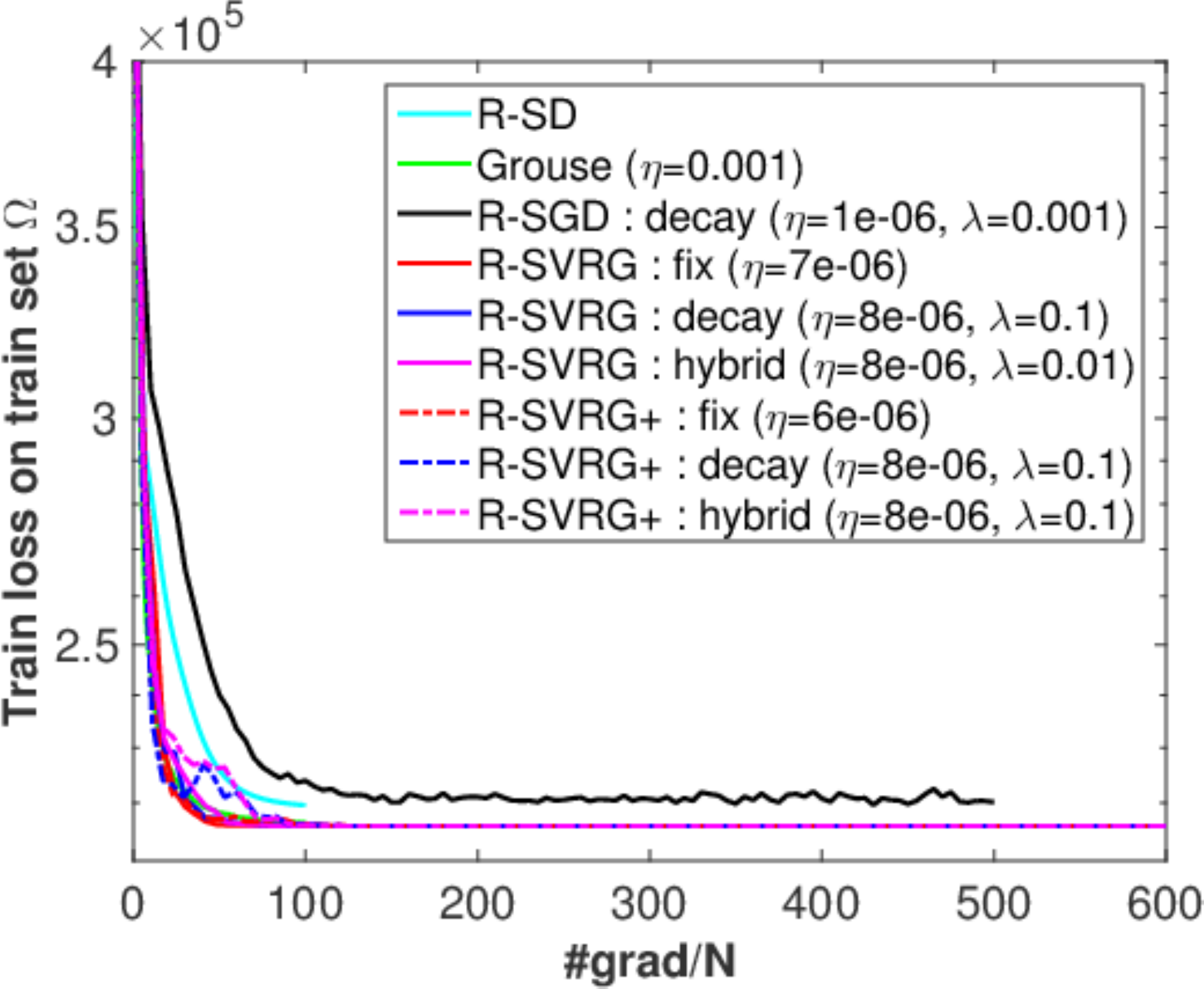}\\
		
		{\small (b-1) Train loss.}
		
	\end{center} 
	\end{minipage}
	\hspace*{-0.1cm}
	\begin{minipage}[t]{0.32\textwidth}
	\begin{center}
		\includegraphics[width=\textwidth]{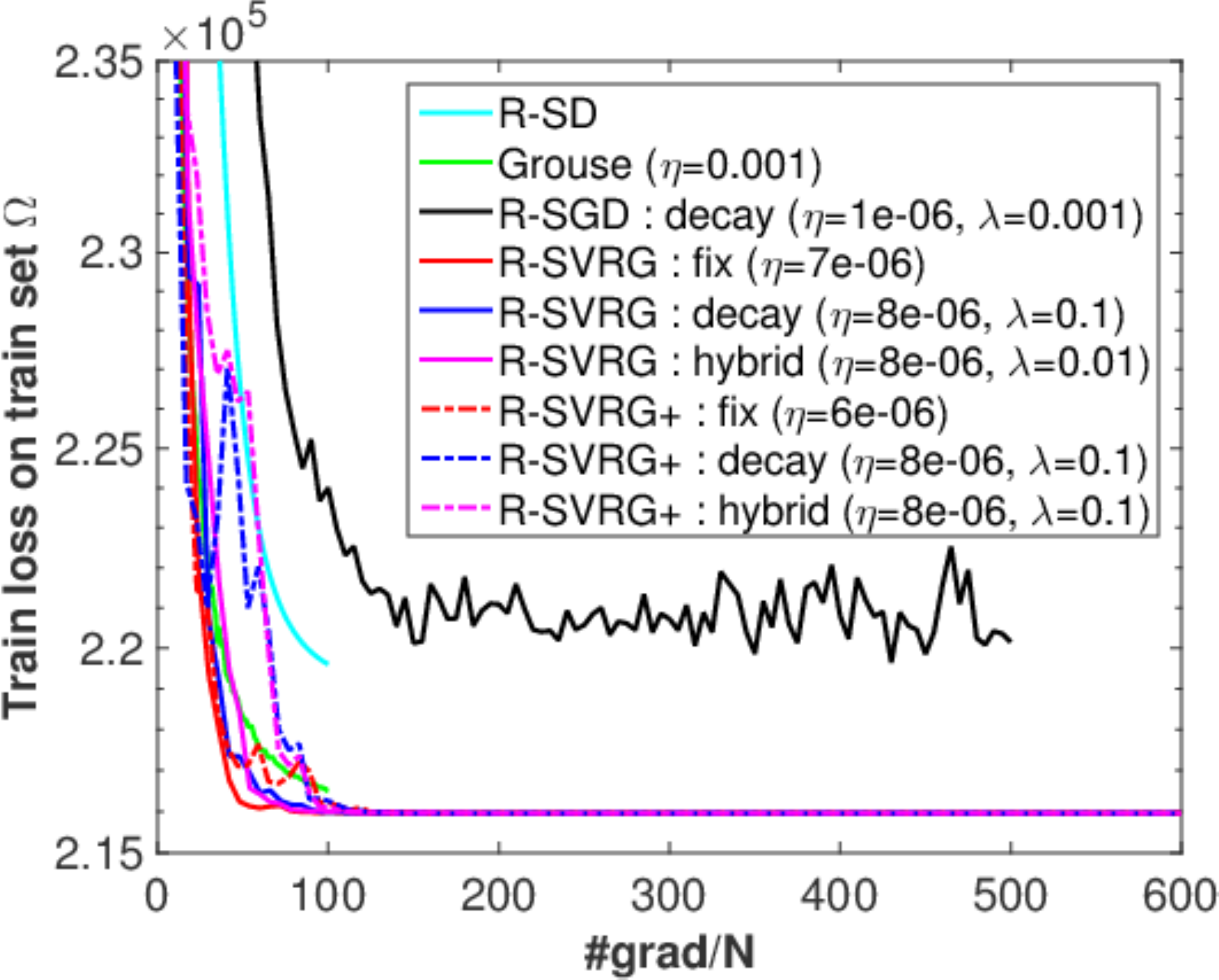}\\
		
		{\small (b-2) Train loss (enlarged). }
		
	\end{center} 
	\end{minipage}
	\hspace*{-0.1cm}
	\begin{minipage}[t]{0.32\textwidth}
	\begin{center}
		\includegraphics[width=\textwidth]{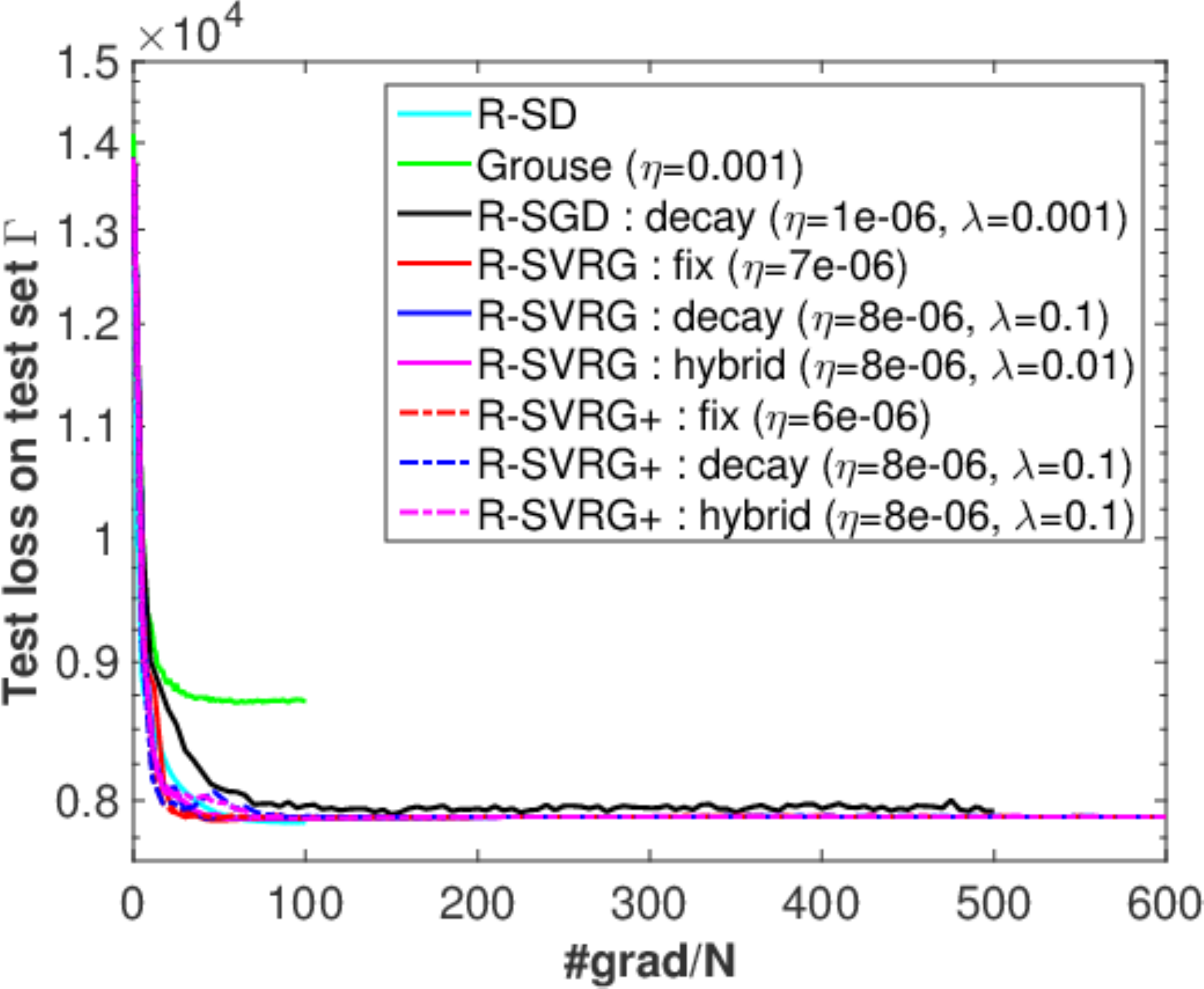}\\
		
		{\small (b-3) Test loss.}
		
	\end{center} 
	\end{minipage}
	\vspace*{0.4cm}
	
	\hspace*{-0.1cm}
	\begin{minipage}[t]{0.32\textwidth}
	\begin{center}
		\includegraphics[width=\textwidth]{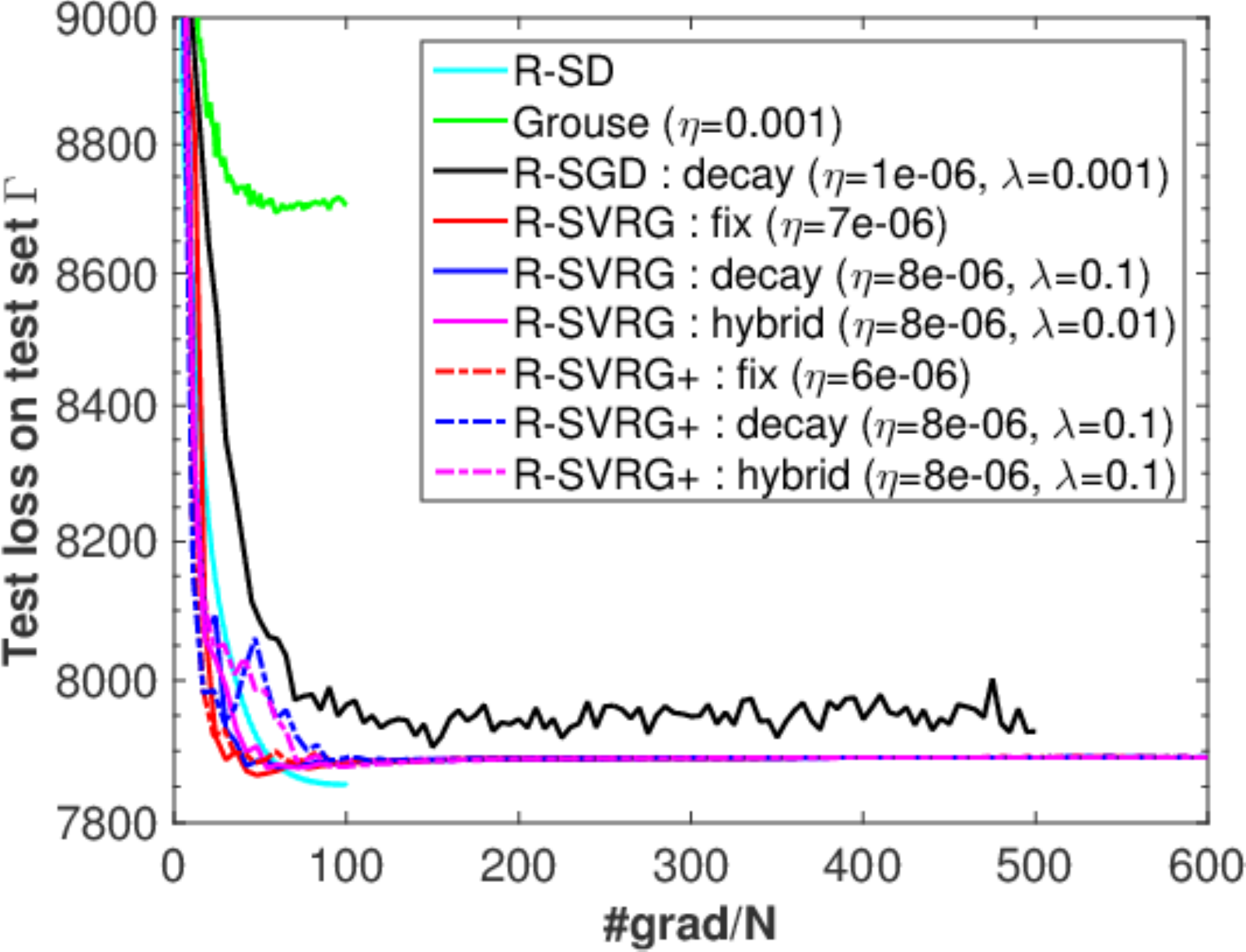}\\
		
		{\small (b-4) Test loss (enlarged).}
		
	\end{center} 
	\end{minipage}
	\begin{minipage}[t]{0.32\textwidth}
	\begin{center}
		\includegraphics[width=\textwidth]{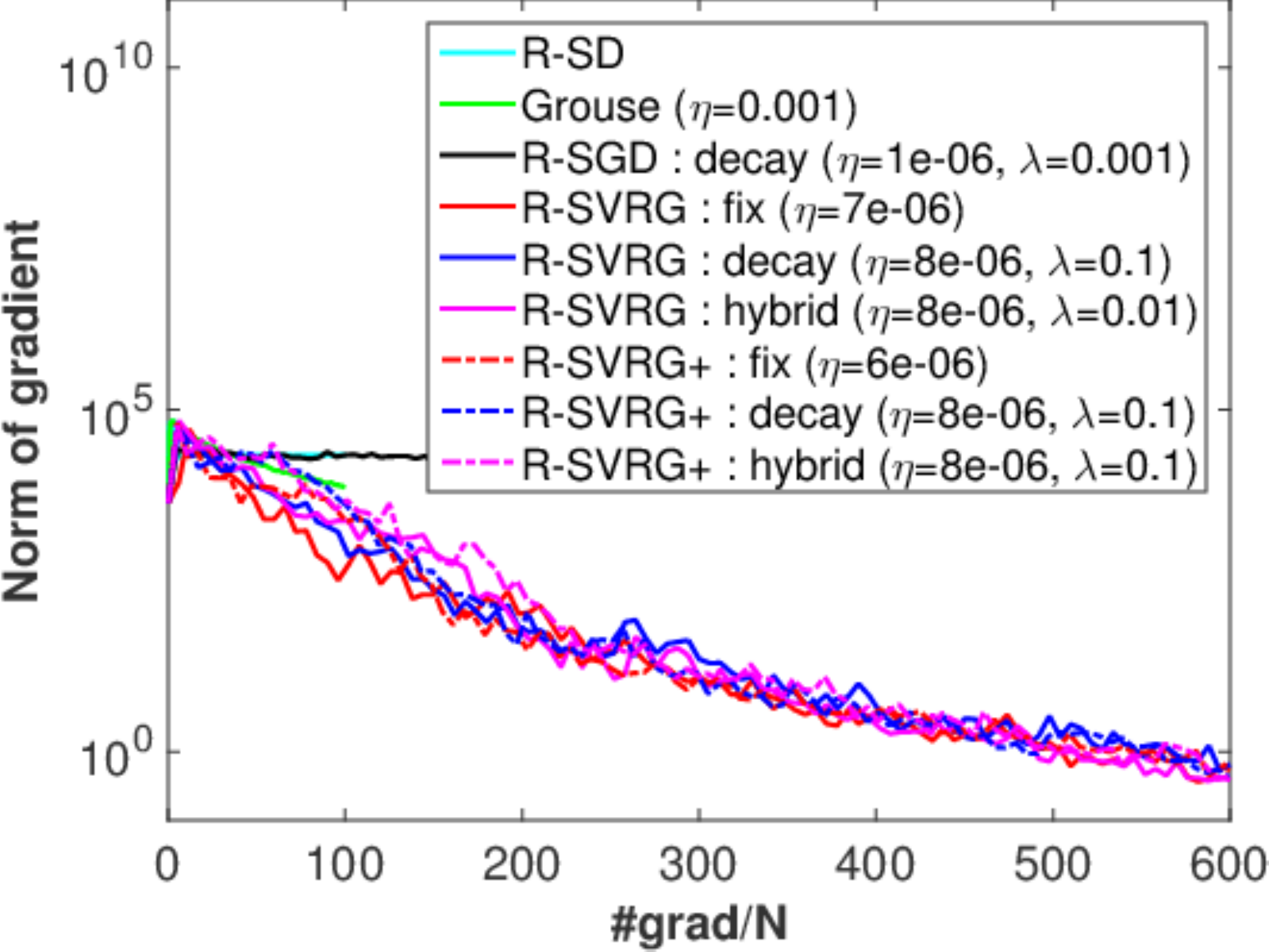}\\
		
		{\small (b-5) Norm of gradient.}
		
	\end{center} 
	\end{minipage}
\vspace*{0.5cm}
	
(b) $r=10$.
\vspace*{0.5cm}	
\caption{The low-rank matrix completion problem (Jester dataset).}
\label{Appen_fig:MC_jester}
\end{center}	
\end{figure}

Next, we show additional results on the Jester dataset 1. We first show all the results in Figure \ref{Appen_fig:MC_jester}(a) for the case of $r=5$, some of which are shown in the manuscript. Figure \ref{Appen_fig:MC_jester}(b) with a larger rank $r=10$. Overall, our proposed R-SVRG and R-SVRG+ indicate much better convergence than R-SD, R-SGD, and Grouse. 

\begin{figure}[t]
\begin{center}
%
%
	\begin{minipage}[t]{0.32\textwidth}
	\begin{center}
		\includegraphics[width=\textwidth]{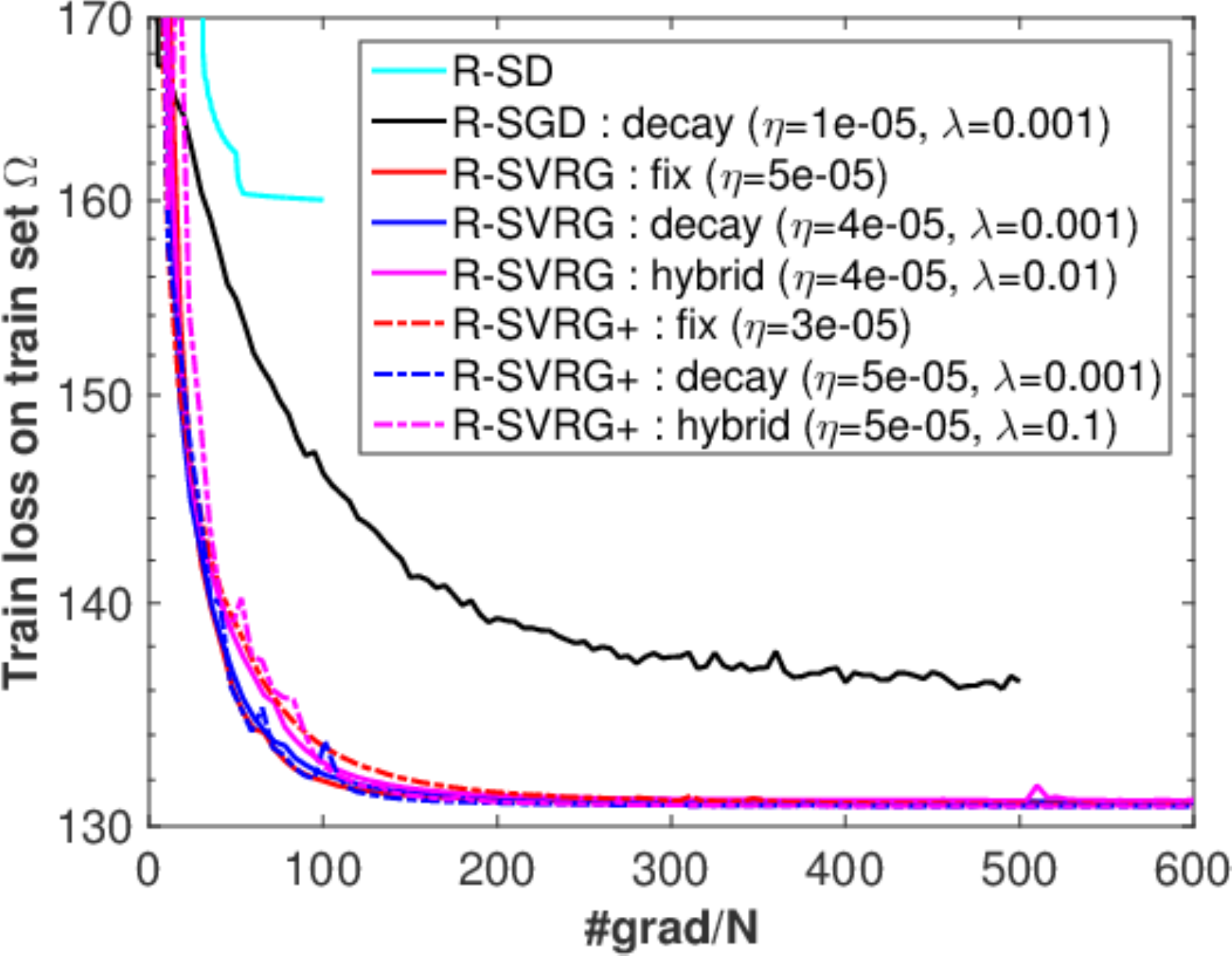}\\
		
		{\small (a-1) Train loss (enlarged). }
		
	\end{center} 
	\end{minipage}
	\hspace*{-0.1cm}
	\begin{minipage}[t]{0.32\textwidth}
	\begin{center}
		\includegraphics[width=\textwidth]{results_pdf/mc_movielens/mc_movielens_train_MSE_enlarge_2_N3952_d6040_r5.pdf}\\
		
		{\small (a-2) Train loss (enlarge 2).}
		
	\end{center} 
	\end{minipage}
	\vspace*{0.4cm}
	
	\hspace*{-0.1cm}
	\begin{minipage}[t]{0.32\textwidth}
	\begin{center}
		\includegraphics[width=\textwidth]{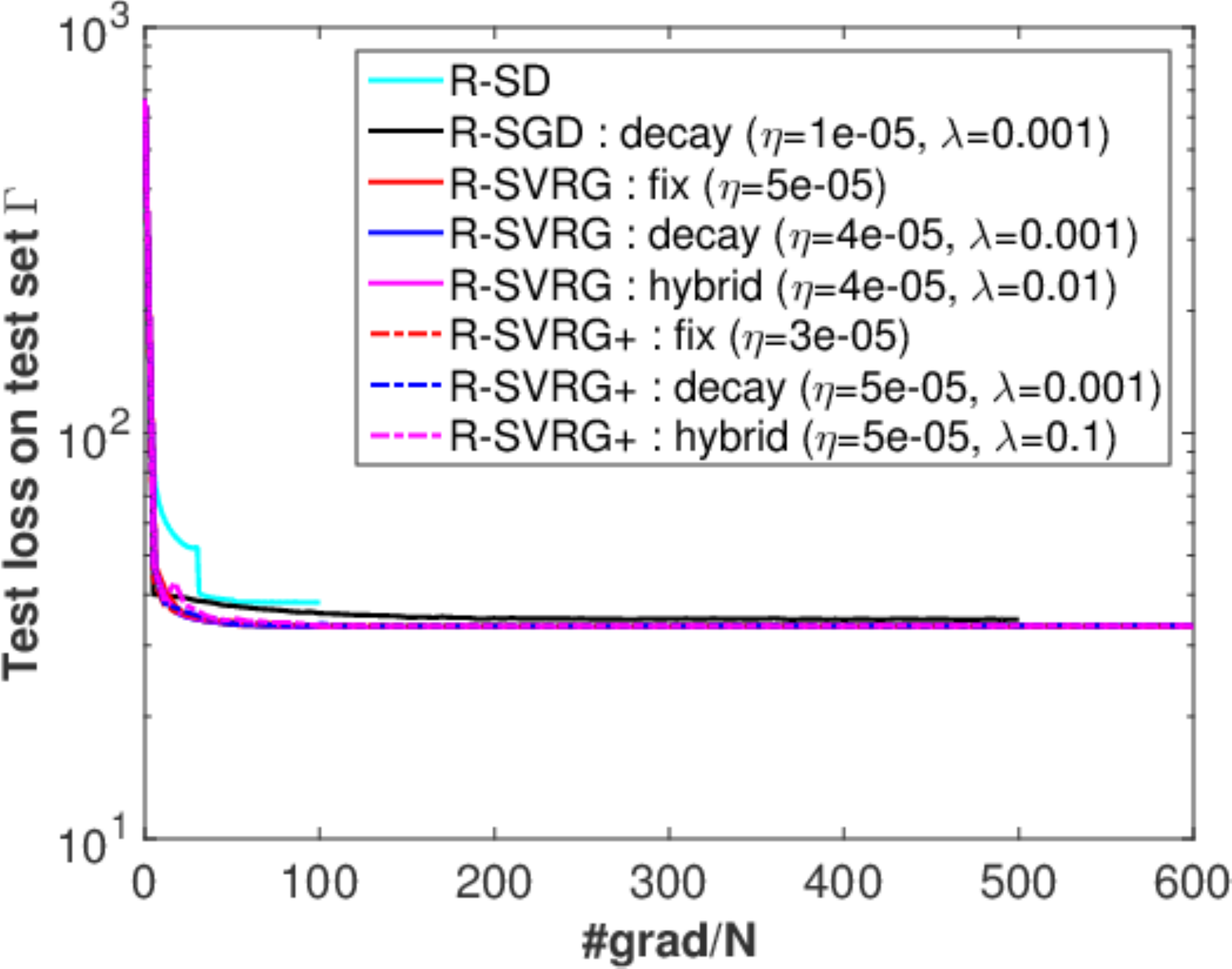}\\
		
		{\small (a-3) Test loss (enlarged).}
		
	\end{center} 
	\end{minipage}
	\begin{minipage}[t]{0.32\textwidth}
	\begin{center}
		\includegraphics[width=\textwidth]{results_pdf/mc_movielens/mc_movielens_test_MSE_enlarge_N3952_d6040_r5.pdf}\\
		
		{\small (a-4) Test loss (enlarged).}
		
	\end{center} 
	\end{minipage}
	\hspace*{-0.1cm}
	\begin{minipage}[t]{0.32\textwidth}
	\begin{center}
		\includegraphics[width=\textwidth]{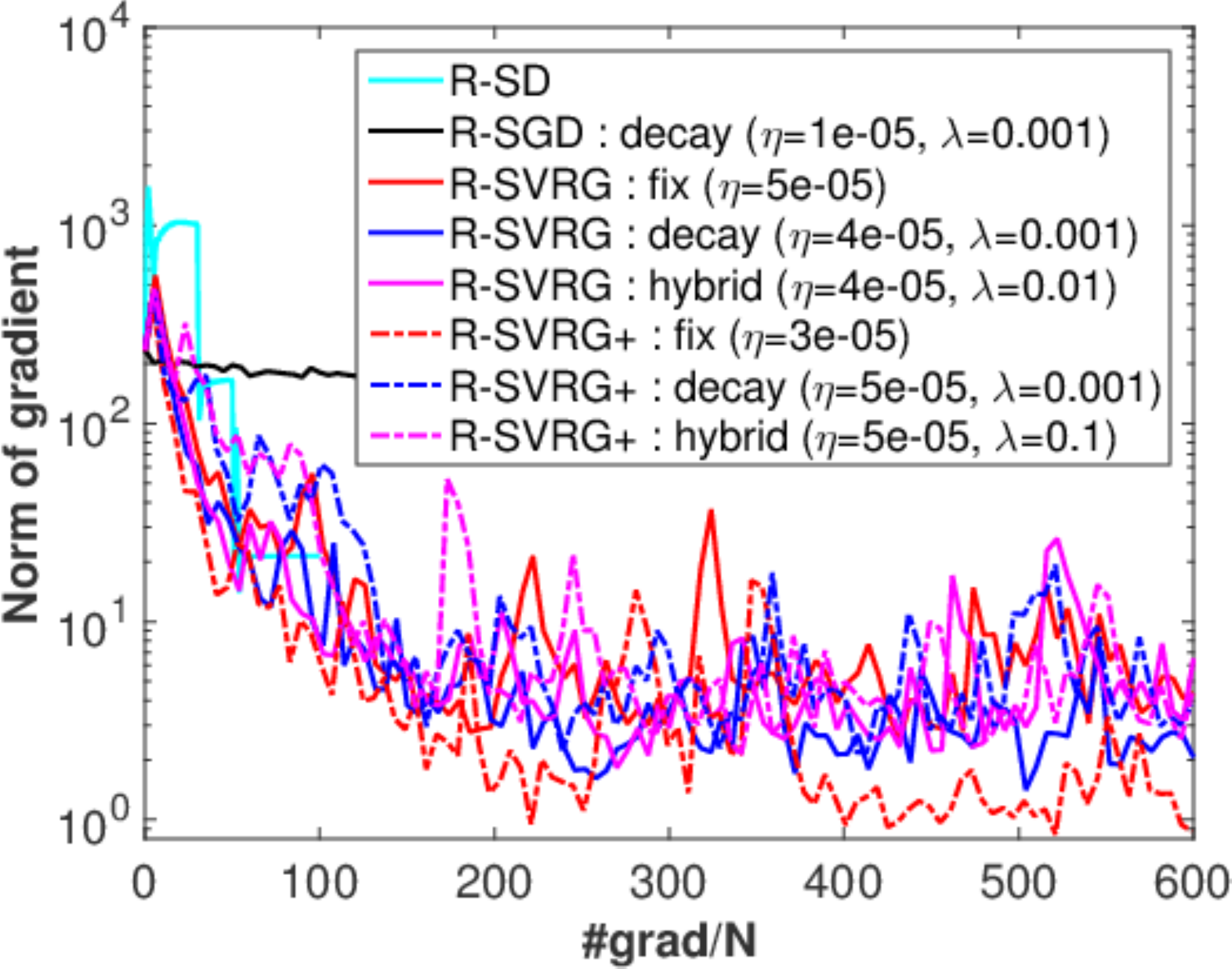}\\
		
		{\small (a-5) Norm of gradient.}
		
	\end{center} 
	\end{minipage}
\vspace*{0.5cm}
	
(a) $r=5$.
\vspace*{0.5cm}

%
%
	\begin{minipage}[t]{0.32\textwidth}
	\begin{center}
		\includegraphics[width=\textwidth]{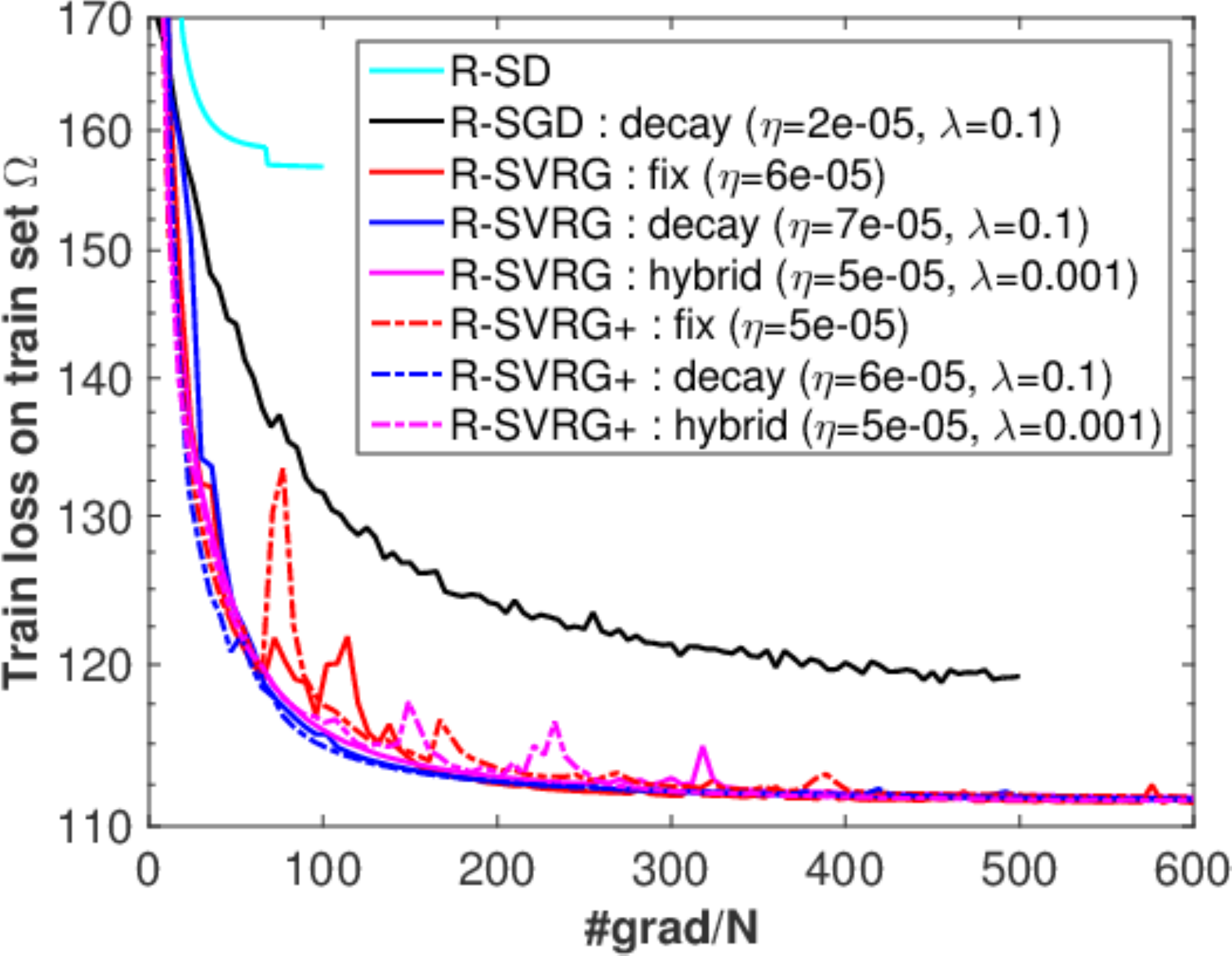}\\
		
		{\small (b-1) Train loss (enlarged). }
		
	\end{center} 
	\end{minipage}
	\hspace*{-0.1cm}
	\begin{minipage}[t]{0.32\textwidth}
	\begin{center}
		\includegraphics[width=\textwidth]{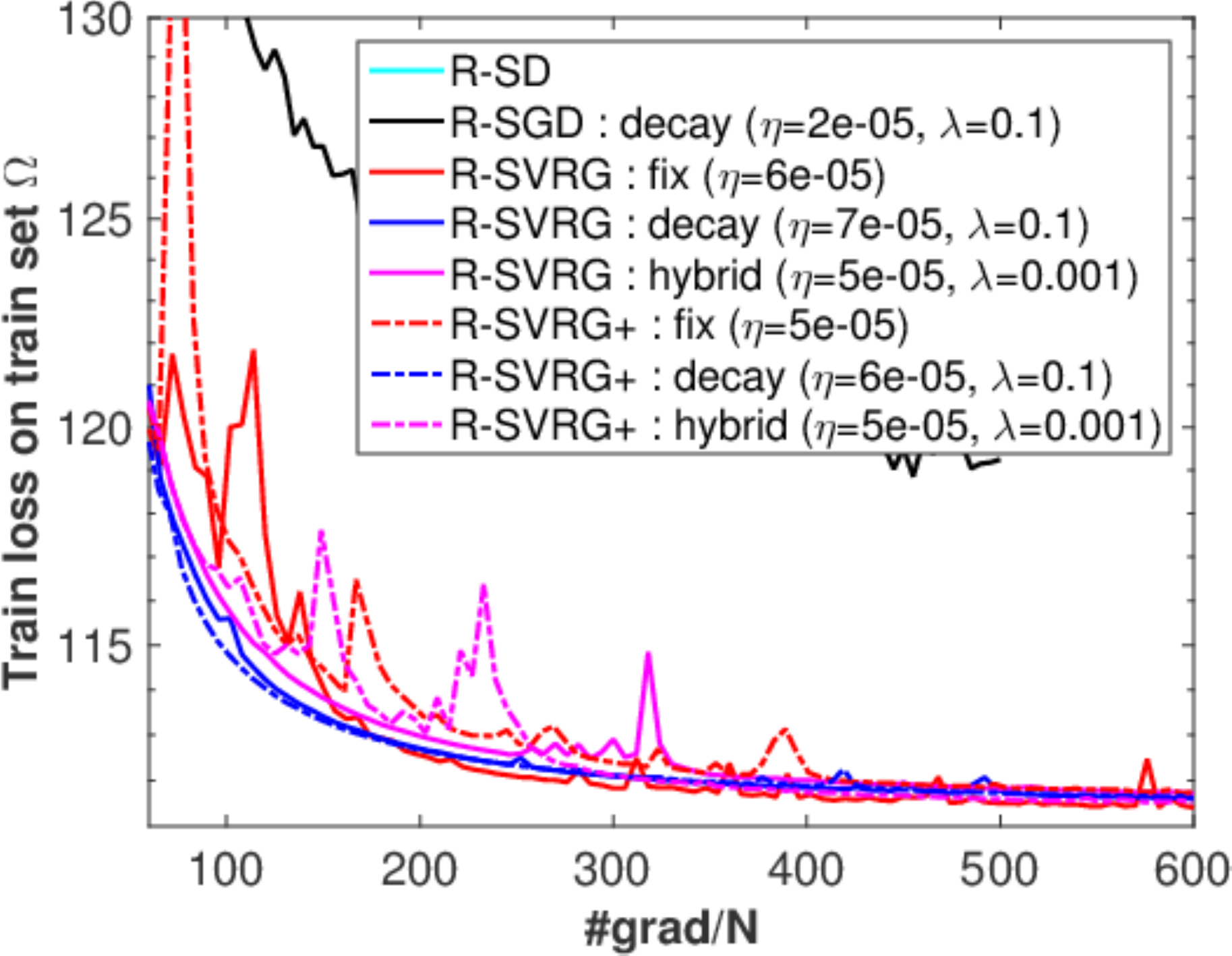}\\
		
		{\small (b-2) Train loss (enlarge 2).}
		
	\end{center} 
	\end{minipage}
	\vspace*{0.4cm}
	
	\hspace*{-0.1cm}
	\begin{minipage}[t]{0.32\textwidth}
	\begin{center}
		\includegraphics[width=\textwidth]{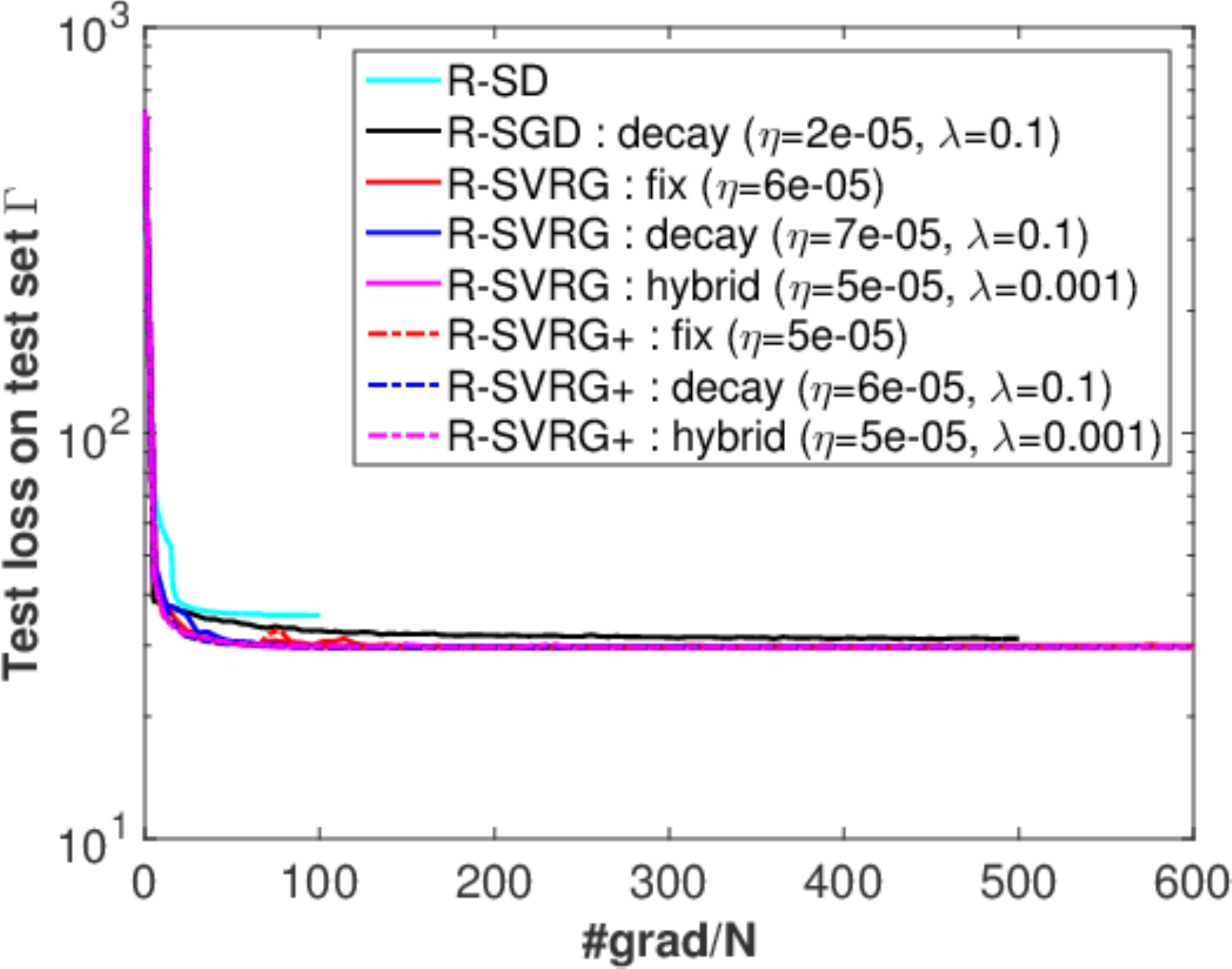}\\
		
		{\small (b-3) Test loss.}
		
	\end{center} 
	\end{minipage}
	\begin{minipage}[t]{0.32\textwidth}
	\begin{center}
		\includegraphics[width=\textwidth]{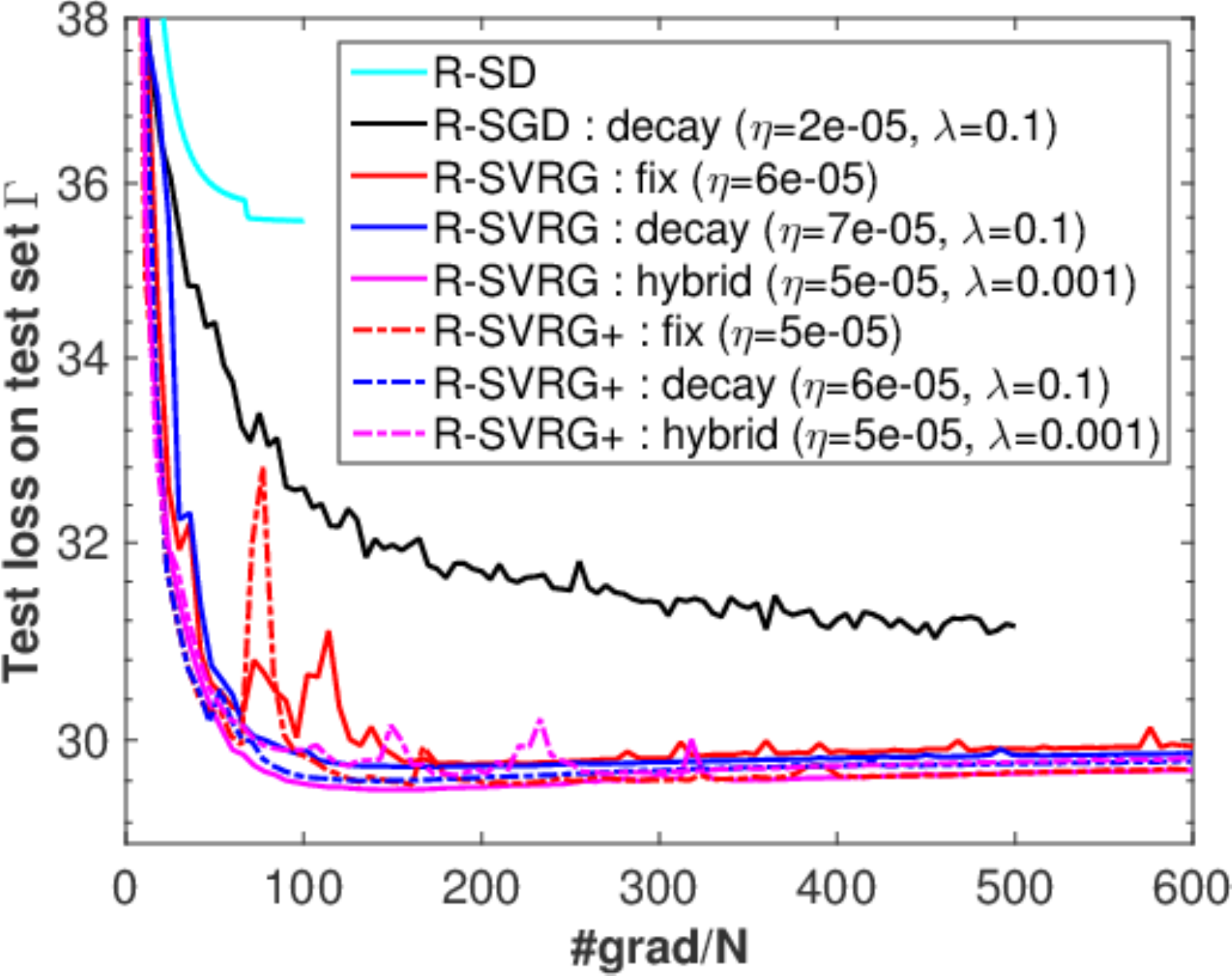}\\
		
		{\small (b-4) Test loss (enlarged).}
		
	\end{center} 
	\end{minipage}
	\hspace*{-0.1cm}
	\begin{minipage}[t]{0.32\textwidth}
	\begin{center}
		\includegraphics[width=\textwidth]{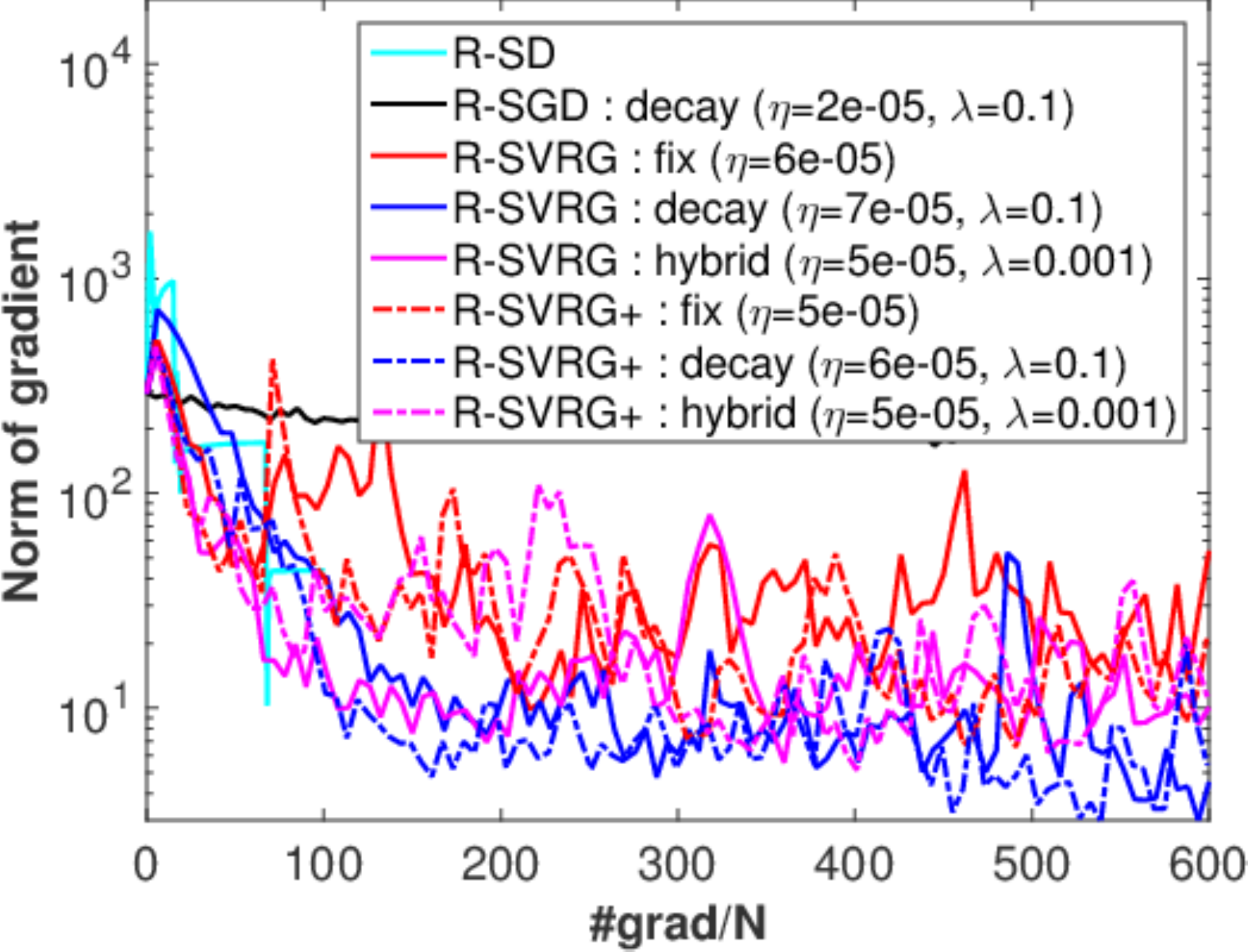}\\
		
		{\small (b-5) Norm of gradient.}
		
	\end{center} 
	\end{minipage}
\vspace*{0.5cm}
	
(b) $r=10$.	
\caption{The low-rank matrix completion problem (MovieLens-1M dataset).}
\label{Appen_fig:MC_movielens}
\end{center}	
\end{figure}

Finally, we show results on the MovieLens-1M dataset. Figure \ref{Appen_fig:MC_movielens}(a) shows the results for the rank $5$. Figures \ref{Appen_fig:MC_movielens}(a-2) and (a-4) are identical to those in the manuscript. We also show results with larger rank $r=10$ case in Figure \ref{Appen_fig:MC_movielens}(b). Once again, our proposed R-SVRG and R-SVRG+ show better results than R-SD and R-SGD.

\begin{figure}[t]
\begin{center}
	\begin{minipage}[t]{0.32\textwidth}
	\begin{center}
		\includegraphics[width=\textwidth]{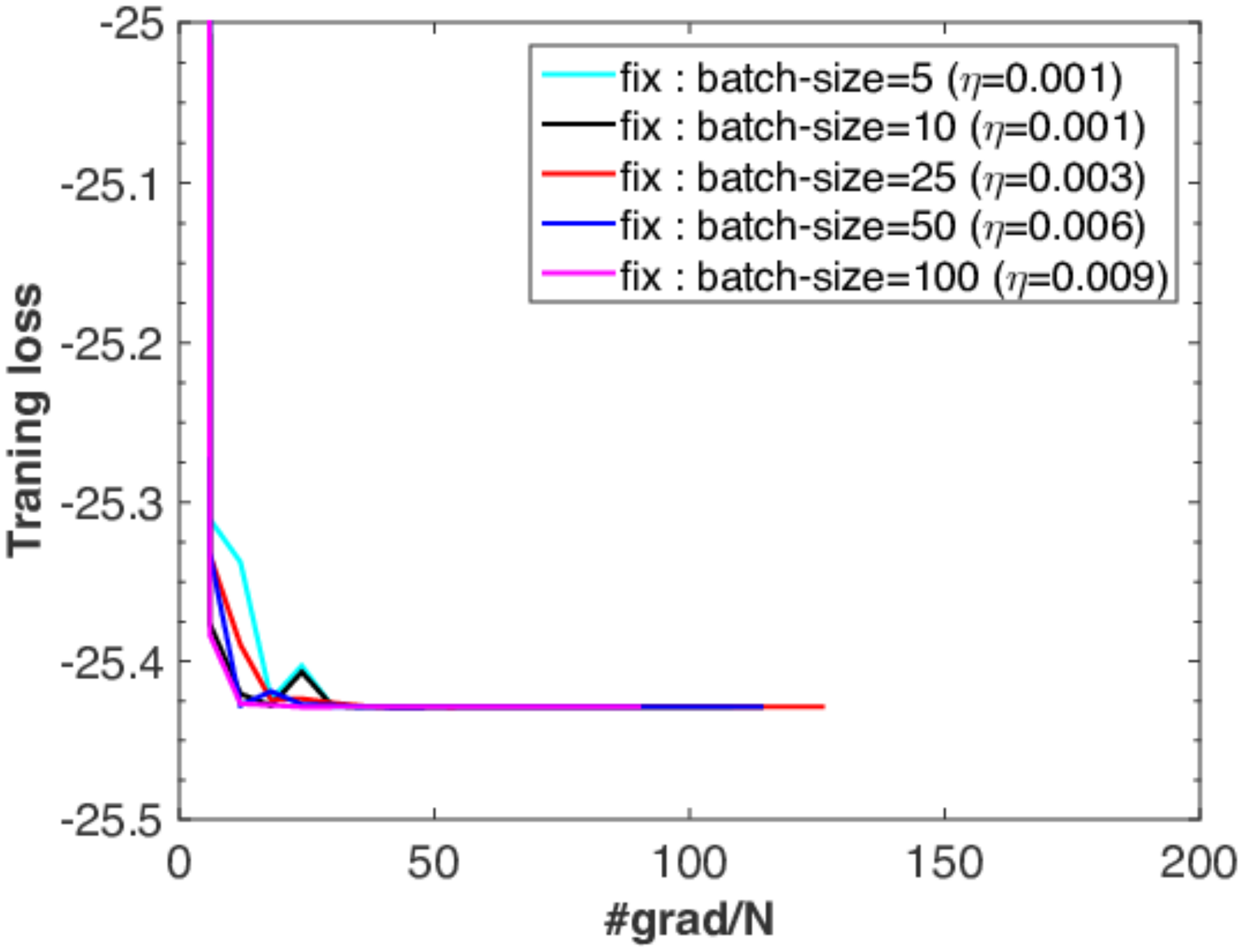}\\
		\vspace*{-1.3cm}
				
		{\small (a-1) Train loss (enlarged).}
		
	\end{center} 
	\end{minipage}
	\hspace*{-0.1cm}
	\begin{minipage}[t]{0.32\textwidth}
	\begin{center}
		\includegraphics[width=\textwidth]{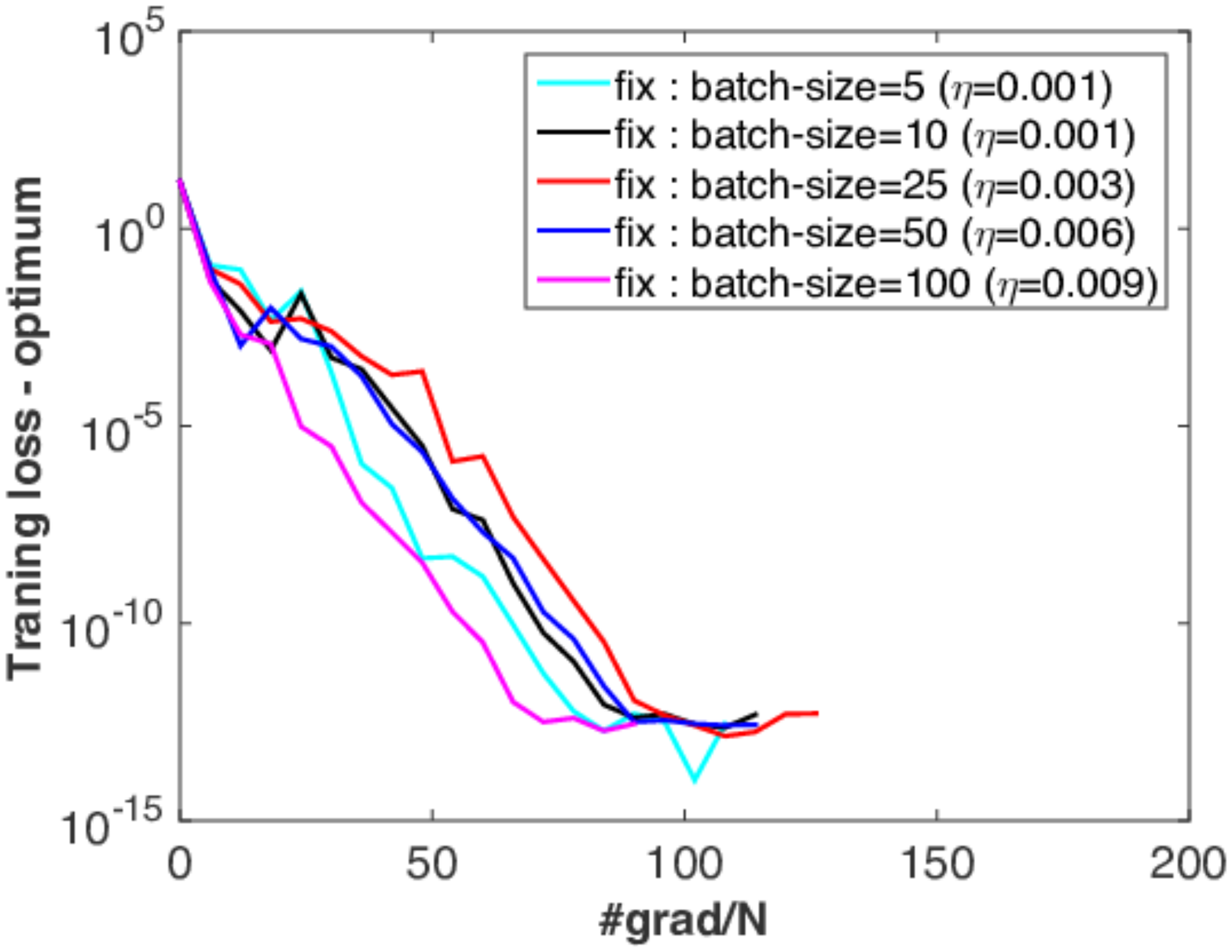}\\
		\vspace*{-1.3cm}
				
		{\small (a-2) Optimality gap. }

	\end{center} 
	\end{minipage}
	\hspace*{-0.1cm}
	\begin{minipage}[t]{0.32\textwidth}
	\begin{center}
		\includegraphics[width=\textwidth]{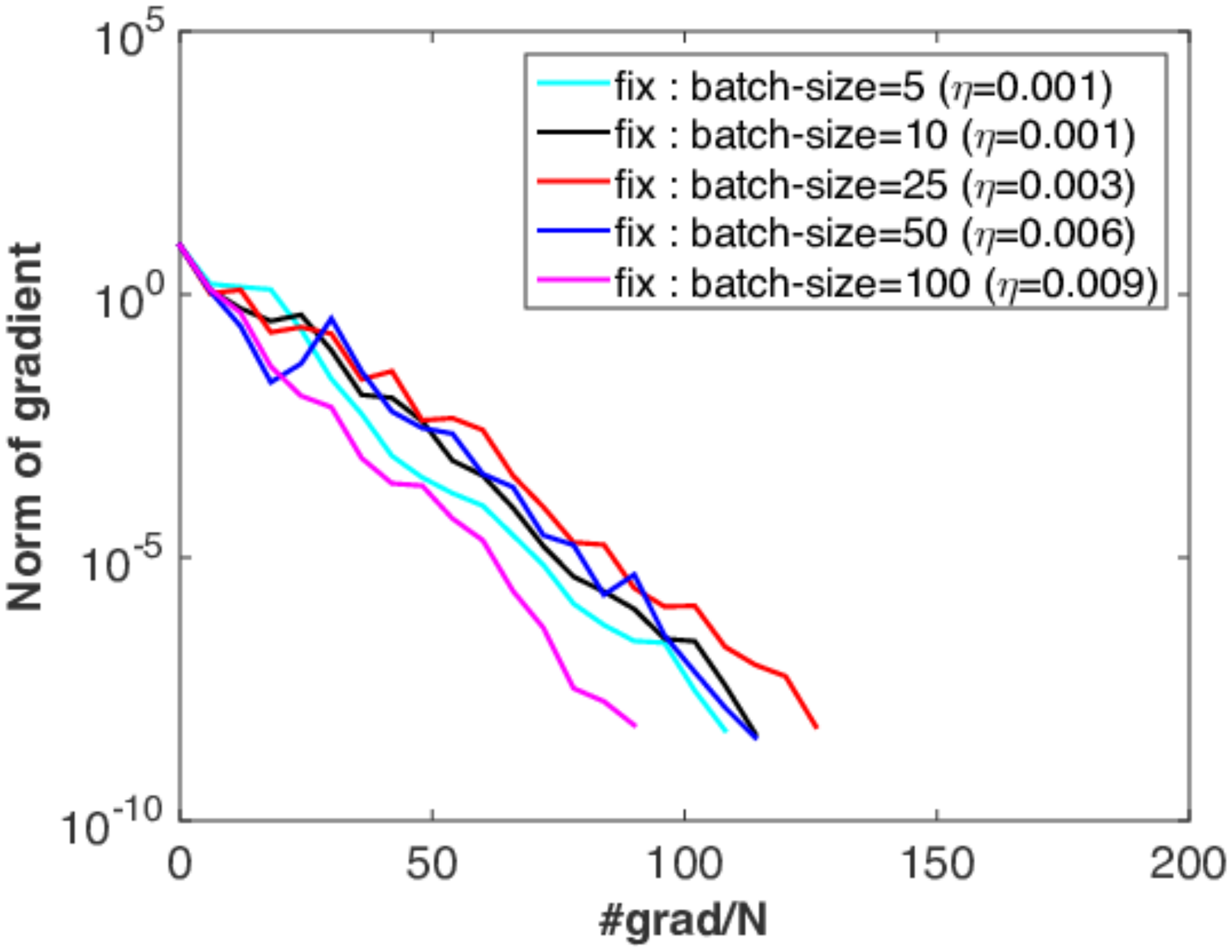}\\
		\vspace*{-1.3cm}		
		
		{\small (a-3) Norm of gradient.}
				
	\end{center} 
	\end{minipage}
\vspace*{-0.5cm}

(a) R-SVRG with fixed step-size.
\vspace*{-1.3cm}

	\begin{minipage}[t]{0.32\textwidth}
	\begin{center}
		\includegraphics[width=\textwidth]{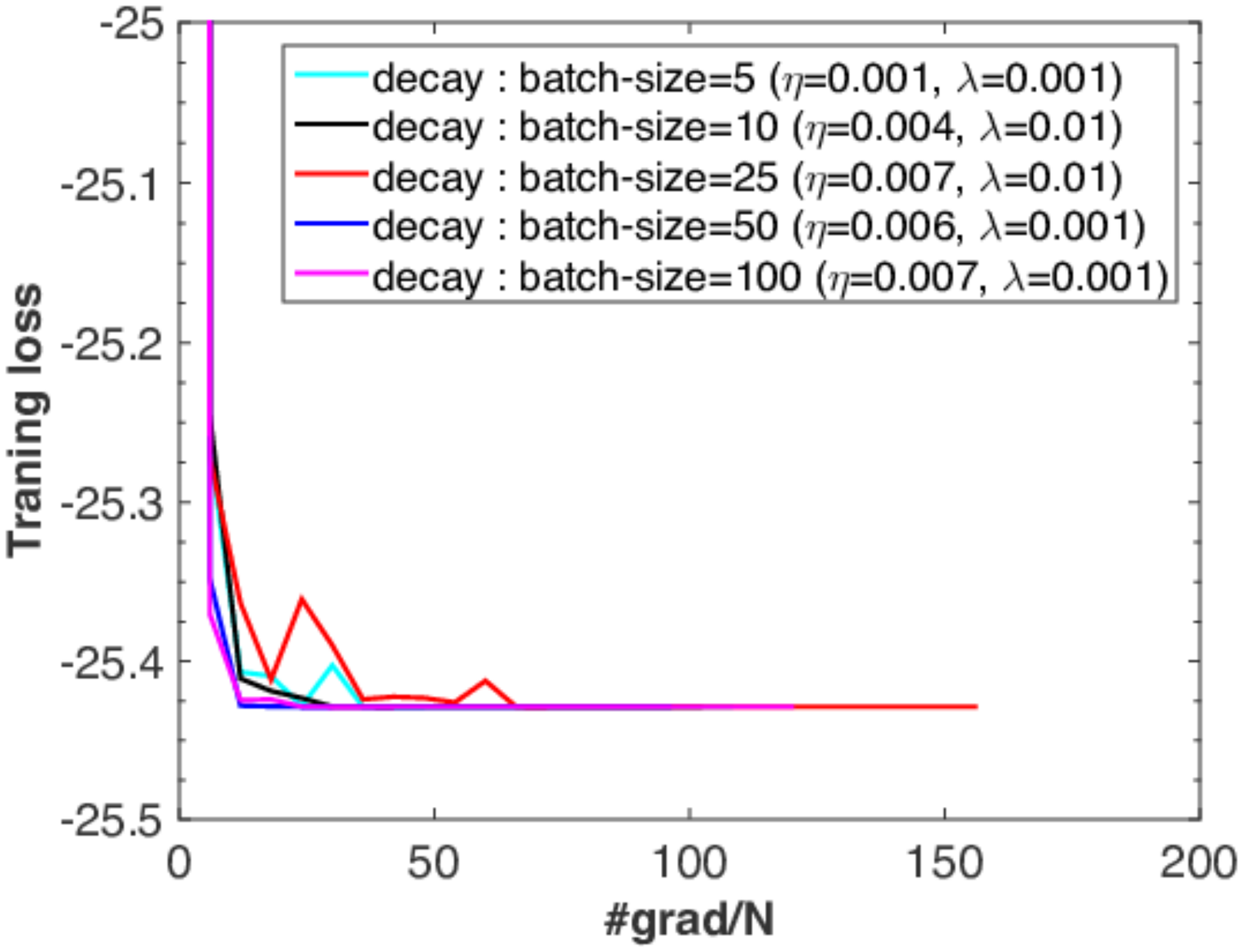}\\
		\vspace*{-1.3cm}
				
		{\small (b-1) Train loss (enlarged).}
		
	\end{center} 
	\end{minipage}
	\hspace*{-0.1cm}
	\begin{minipage}[t]{0.32\textwidth}
	\begin{center}
		\includegraphics[width=\textwidth]{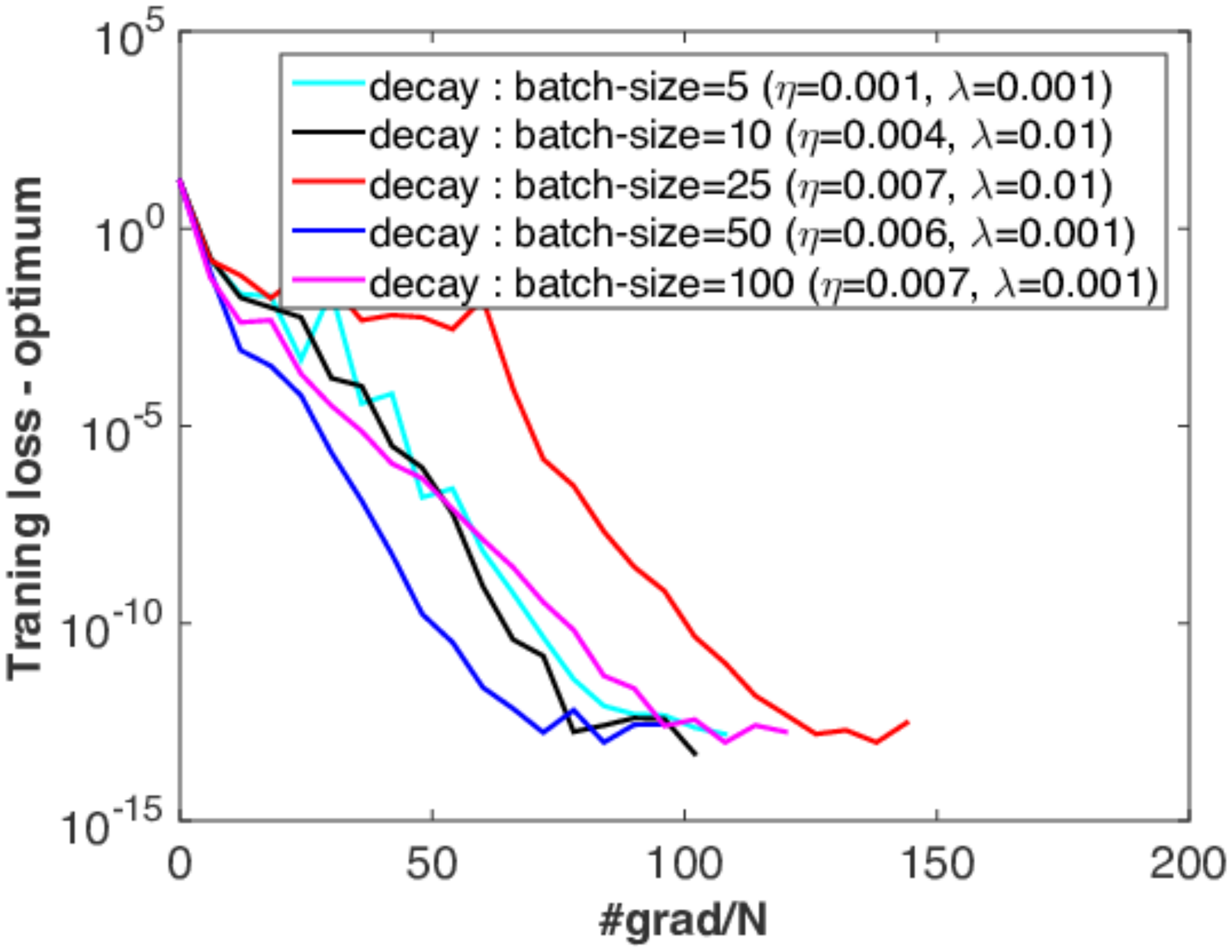}\\
		\vspace*{-1.3cm}
				
		{\small (b-2) Optimality gap. }
				
	\end{center} 
	\end{minipage}
	\hspace*{-0.1cm}
	\begin{minipage}[t]{0.32\textwidth}
	\begin{center}
		\includegraphics[width=\textwidth]{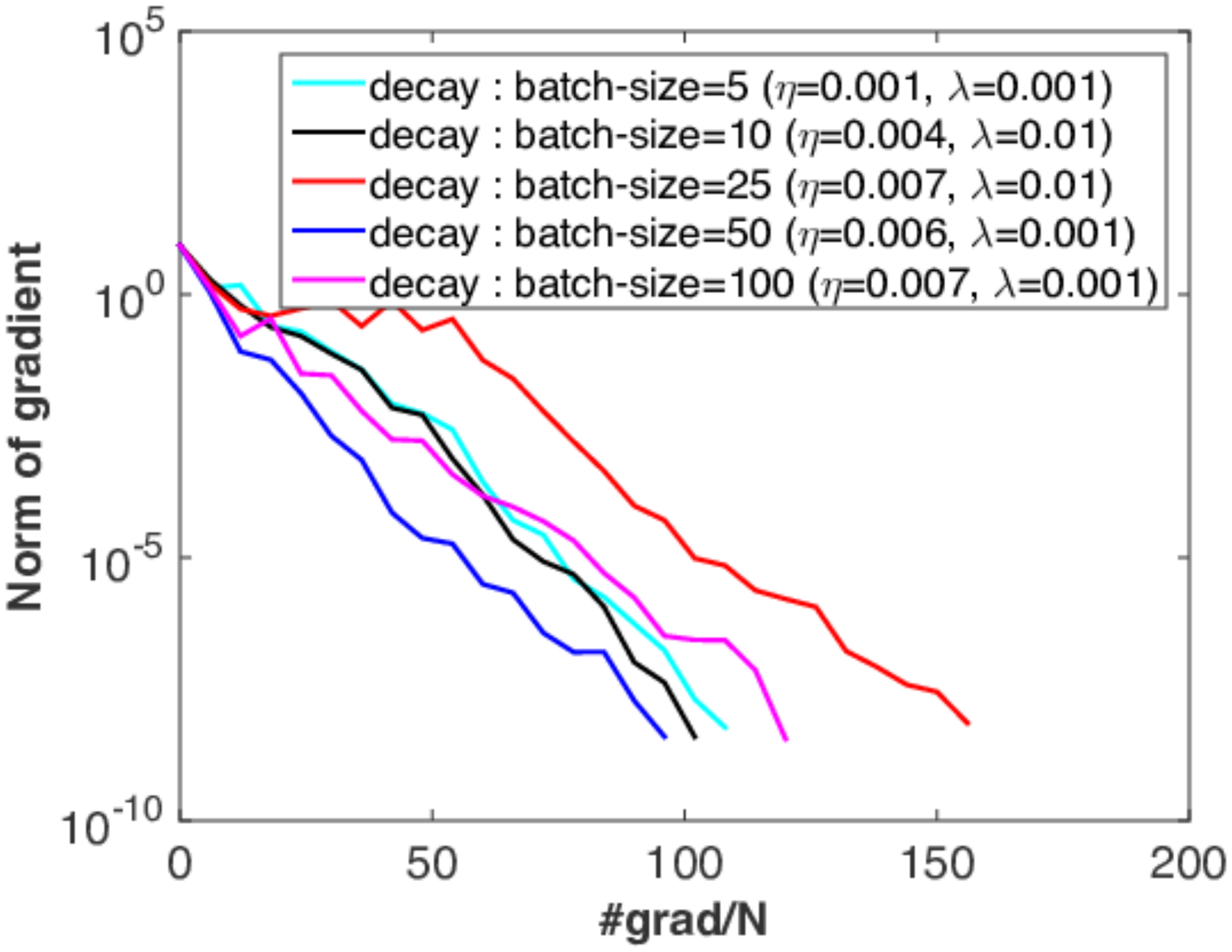}\\
		\vspace*{-1.3cm}
				
		{\small (b-3) Norm of gradient.}
				
	\end{center} 
	\end{minipage}
\vspace*{-0.5cm}
	
(b) R-SVRG with decay step-size.
\vspace*{-1.3cm}

	\begin{minipage}[t]{0.32\textwidth}
	\begin{center}
		\includegraphics[width=\textwidth]{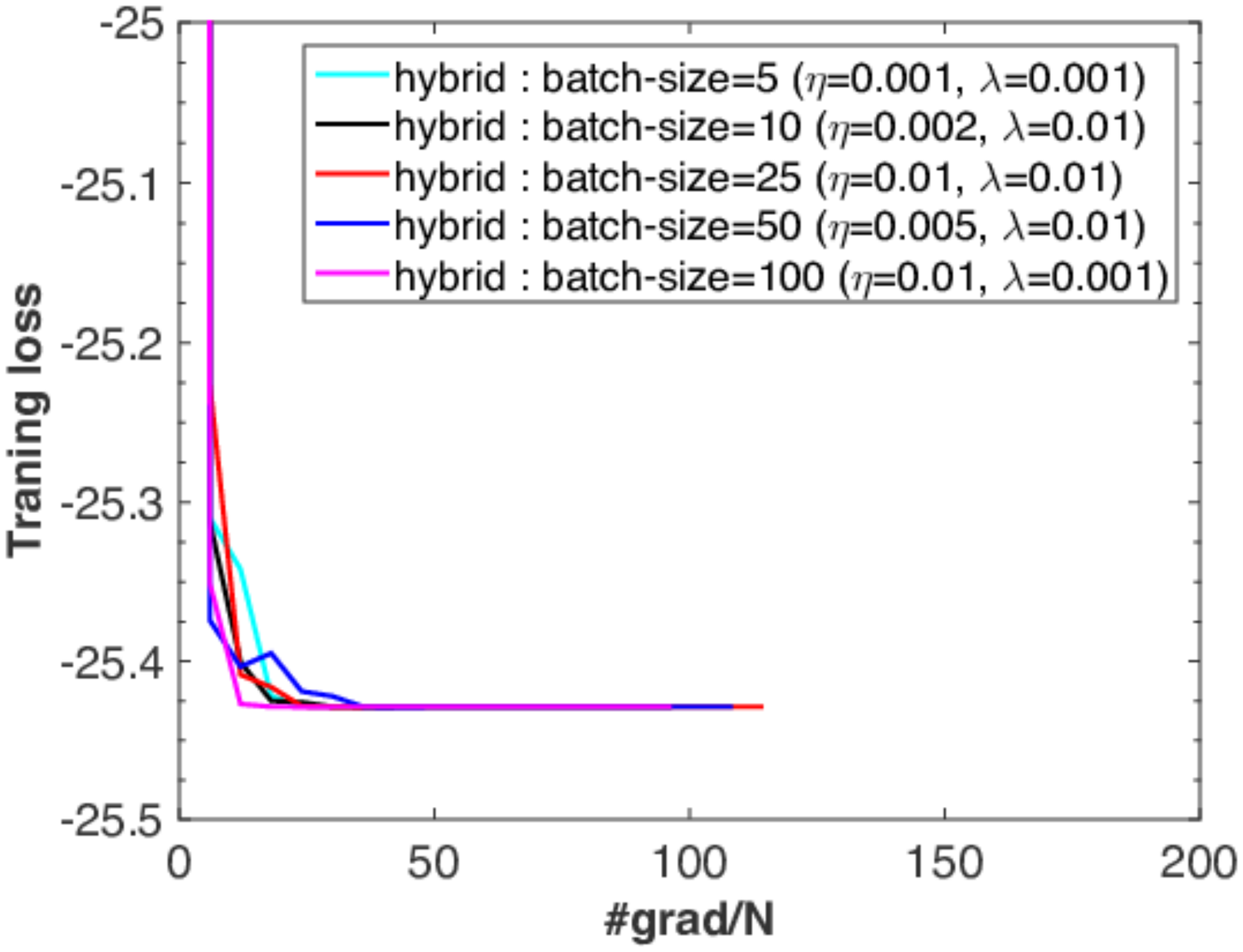}\\
		\vspace*{-1.3cm}
				
		{\small (c-1) Train loss (enlarged).}
		
	\end{center} 
	\end{minipage}
	\hspace*{-0.1cm}
	\begin{minipage}[t]{0.32\textwidth}
	\begin{center}
		\includegraphics[width=\textwidth]{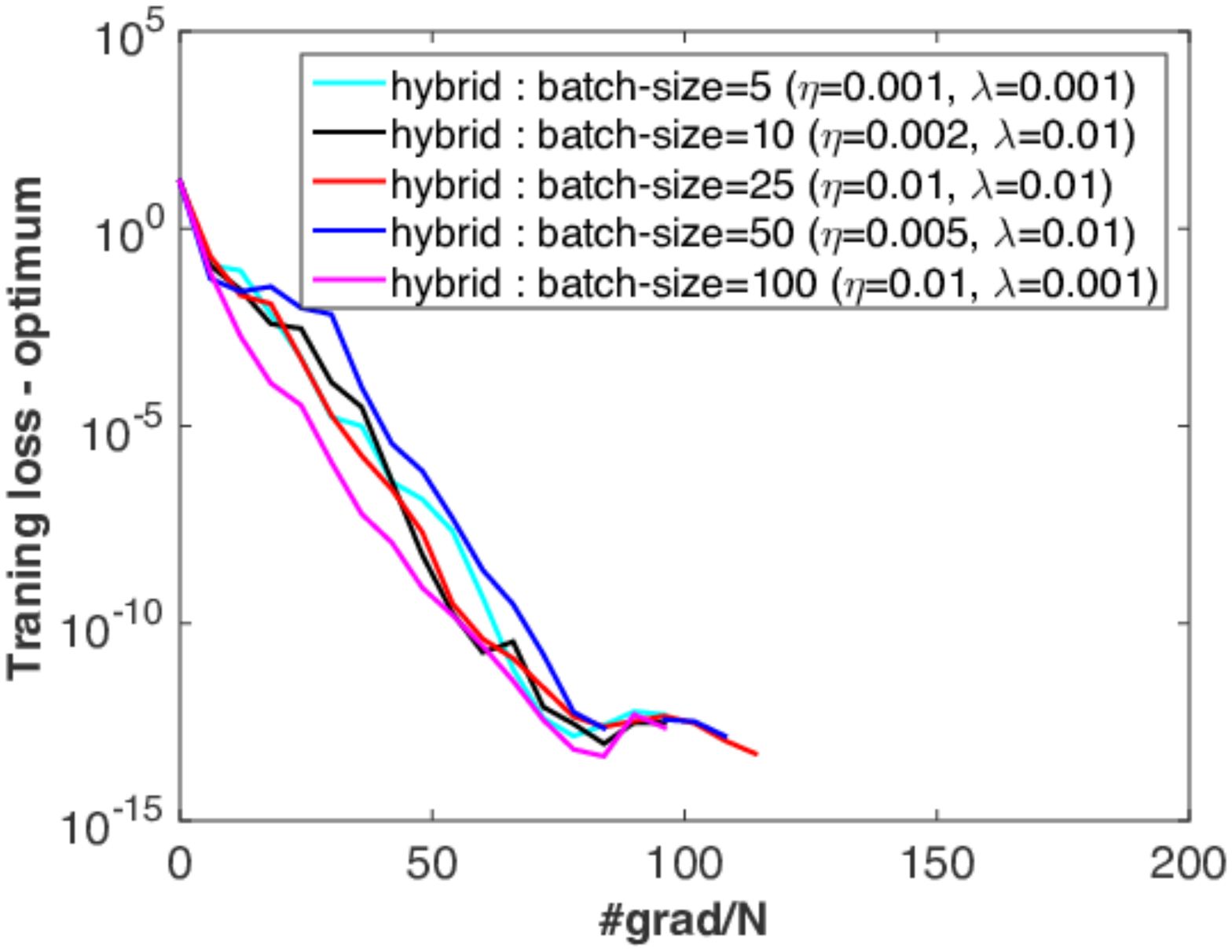}\\
		\vspace*{-1.3cm}
				
		{\small (c-2) Optimality gap. }
		
	\end{center} 
	\end{minipage}
	\hspace*{-0.1cm}
	\begin{minipage}[t]{0.32\textwidth}
	\begin{center}
		\includegraphics[width=\textwidth]{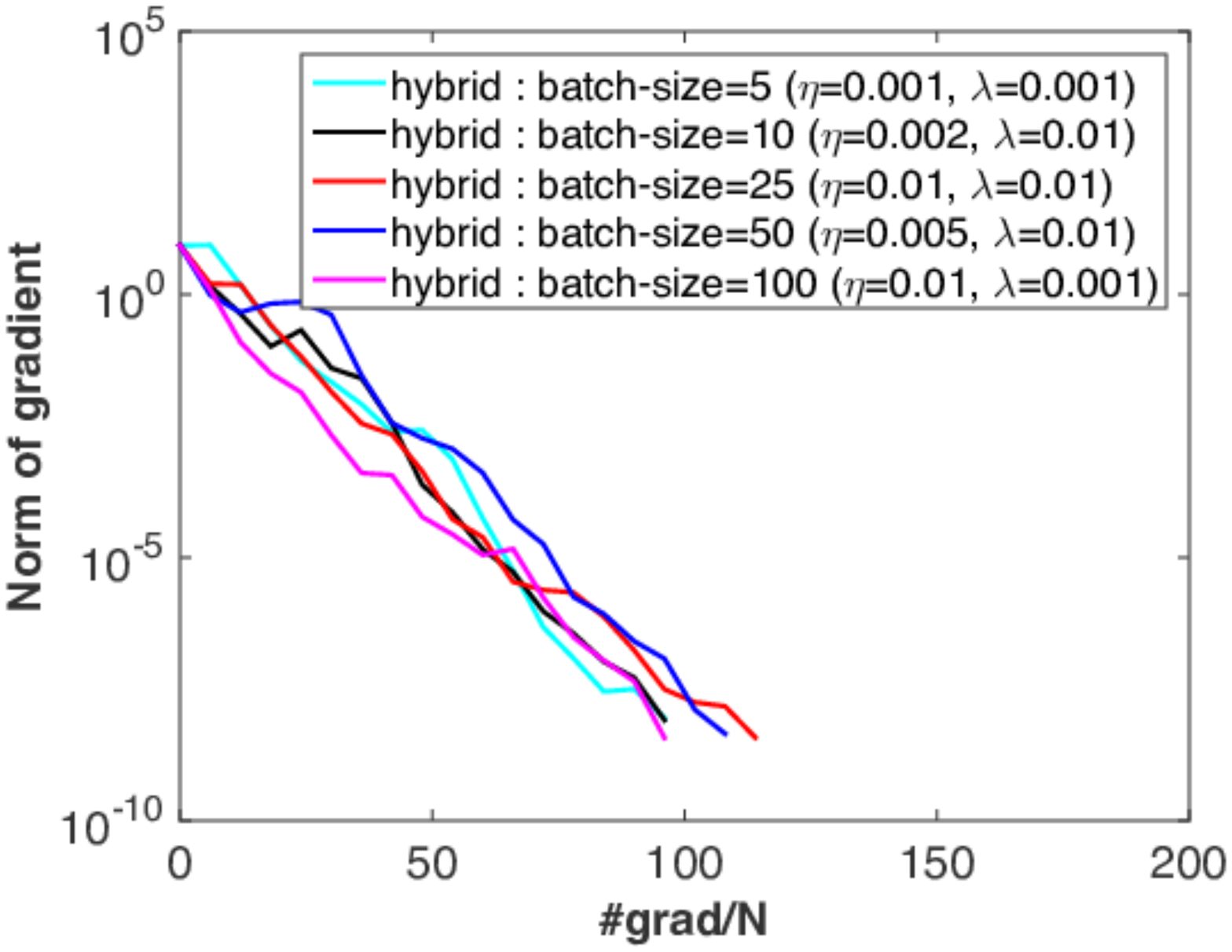}\\
		\vspace*{-1.3cm}
		
		{\small (c-3) Norm of gradient.}
		
	\end{center} 
	\end{minipage}
\vspace*{-0.5cm}
	
(c) R-SVRG with hybrid step-size.
	
\caption{Batch-size comparisons for R-SVRG (PCA problem: $N=10000, d=20, r=5$).}
\label{Appen_fig:Batchsize_Comp_PCA_results_N_10000_d_20_r_5}
\end{center}		
\end{figure}

{\bf Effect of batch-size.} Here, we show the effect of batch-size on R-SVRG. For this purpose, we consider the PCA problem of $N=10000$, $d=20$, and $r=5$. Figures \ref{Appen_fig:Batchsize_Comp_PCA_results_N_10000_d_20_r_5}(a)-(c) show the results for three step-size sequences of R-SVRG, respectively. We consider five different batch-sizes from $\{5, 10, 25, 50, 100\}$. The figures show that R-SVRG similar performance across different batch-sizes.

\clearpage

\clearpage

\clearpage

\clearpage

\clearpage

\clearpage

\end{document}